\documentclass[times,twocolumn,final]{elsarticle}

\usepackage{medima}
\usepackage{framed,multirow}
\usepackage{placeins}
\usepackage{amsmath,amssymb}
\usepackage{latexsym}
\usepackage{makecell}
\usepackage{url}
\usepackage[figuresright]{rotating}
\usepackage{pdflscape}
\usepackage[table]{xcolor}
\usepackage[dvipsnames]{xcolor}
\usepackage{adjustbox}
\usepackage{threeparttable}
\usepackage{booktabs}
\usepackage{tabularx}
\usepackage{multirow}
\usepackage{algorithm}
\usepackage{algorithmic}
\usepackage{array}

\definecolor{airforceblue}{rgb}{0.36, 0.54, 0.66}
\definecolor{ballblue}{rgb}{0.004, 0.50, 0.69}

\usepackage{xurl}
\usepackage[colorlinks]{hyperref}
\usepackage[utf8]{inputenc}
\usepackage{graphicx}

\AtBeginDocument{%
  \hypersetup{
    citecolor=ballblue,
    linkcolor=ballblue,   
    urlcolor=ballblue}}

\def\red#1{\textcolor{Red}{\textbf{#1}}}

\def\bblue#1{\textcolor{Blue}{\textbf{#1}}}
\def\green#1{\textcolor{Green}{\textbf{#1}}}

\journal{Medical Image Analysis}

\begin{document}

\verso{Tao Yang \textit{et~al.}}

\begin{frontmatter}
\title{Unsupervised Anomaly Detection in Brain MRI via Disentangled Anatomy Learning\tnoteref{note1}}
\tnotetext[note1]{Accepted in \textit{Medical Image Analysis}. Licensed under CC BY-NC-ND 4.0.}
\author[1]{Tao \snm{Yang}}
\author[2]{Xiuying \snm{Wang}}
\author[1]{Hao \snm{Liu}}
\author[3]{Guanzhong \snm{Gong}}
\author[4]{Lian-Ming \snm{Wu}\corref{cor1}}
\author[5]{Yu-Ping \snm{Wang}}
\author[1]{Lisheng \snm{Wang}\corref{cor1}}
\cortext[cor1]{Corresponding author.}\ead{lswang@sjtu.edu.cn}
\address[1]{School of Automation and Intelligent Sensing, Shanghai Jiao Tong University, Shanghai, China.}
\address[2]{School of Computer Science, The University of Sydney, Sydney, Australia.}
\address[3]{Shandong Cancer Hospital and Institute, Shandong First Medical University and Shandong Academy of Medical Sciences, Jinan, China.}
\address[4]{Department of Radiology, Renji Hospital, School of Medicine, Shanghai Jiao Tong University, Shanghai, China.}
\address[5]{Biomedical Engineering Department, Tulane University, New Orleans, USA.}
\received{-}
\finalform{-}
\accepted{-}
\availableonline{-}
\communicated{-}
\begin{abstract}
Detection of various lesions in brain MRI is clinically critical, but challenging due to the diversity of lesions and variability in imaging conditions. Current unsupervised learning methods detect anomalies mainly through reconstructing abnormal images into pseudo-healthy images (PHIs) by normal samples learning and then analyzing differences between images. However, these unsupervised models face two significant limitations: restricted generalizability to multi-modality and multi-center MRIs due to their reliance on the specific imaging information in normal training data, and constrained performance due to abnormal residuals propagated from input images to reconstructed PHIs. To address these limitations, two novel modules are proposed, forming a new PHI reconstruction framework. Firstly, the disentangled representation module is proposed to improve generalizability by decoupling brain MRI into imaging information and essential imaging-invariant anatomical images, ensuring that the reconstruction focuses on the anatomy. Specifically, brain anatomical priors and a differentiable one-hot encoding operator are introduced to constrain the disentanglement results and enhance the disentanglement stability. Secondly, the edge-to-image restoration module is designed to reconstruct high-quality PHIs by restoring the anatomical representation from the high-frequency edge information of anatomical images, and then recoupling the disentangled imaging information. This module not only suppresses abnormal residuals in PHI by reducing abnormal pixels input through edge-only input, but also effectively reconstructs normal regions using the preserved structural details in the edges. Evaluated on nine public datasets (4,443 patients’ MRIs from multiple centers), our method outperforms 17 state-of-the-art methods, achieving absolute improvements of +18.32\% in average precision and +13.64\% in Dice similarity coefficient.
\end{abstract}
\end{frontmatter}

\section{Introduction}
    Timely diagnosis and treatment of various types of brain anomalies are crucial for reducing morbidity and mortality and improving functional outcomes \citep{ref1}. Magnetic resonance imaging (MRI), due to its merits in flexible imaging, soft tissue contrast and non-ionizing acquisition, has been widely used in clinical diagnosis of brain anomalies \citep{sahraian2010role,villanueva2017current}. MRI contains both anatomical information that is independent of imaging conditions and imaging information determined by the associated imaging conditions \citep{ref4}. Imaging information in different MRIs may vary significantly with varying imaging conditions, such as different imaging sequences (T1-weighted, T2-weighted), scanners, and acquisition parameters \citep{ref3}. Additionally, different types of brain anomalies show diverse appearances on MRI. The diversity of brain lesions and variability of imaging conditions make diagnosing a wide range of brain anomalies in MRIs, from common to rare, highly dependent on the radiologists' professional knowledge, expertise, and experience, apart from its time-consuming and labor-intensive assessment process. In addition, ever-increasing volumes of 3D brain MRIs overburden radiologists with timely diagnosis and medical decision making.
	
	The rapid rise and advancement of artificial intelligence (AI), in particular deep learning, provides strong impetus to improving the diagnostic efficiency and reducing clinical workload \citep{raya2021ai,ahn2022association}. While supervised learning methods are recognized for their accuracy when detecting the pre-defined specific anomalies found in the training set, their capacity to detect new or unseen types of anomalies is often severely limited and relies on labor-intensive annotation \citep{dong2023swssl,li2023self}. More recently, unsupervised learning methods have gained increasing attention because they can detect various anomalies through learning from normal samples without requiring expert annotations \citep{cai2025medianomaly,lagogiannis2023unsupervised,ref15,ref13,xu2024facing,ref17,ref12}. Among these unsupervised models, the mainstream reconstruction-based methods highlight various anomalies by reconstructing abnormal images into pseudo-healthy images (PHIs) and calculating the differences between images. 
    
    However, existing reconstruction-based methods show significantly limited generalizability and performance for anomaly detection in brain MRIs: (1) they tend to overfit specific imaging-related information in the normal training data, thus are difficult to extend to multi-modality MRI and often have limited generalizability to MRIs from different centers \citep{ref14,ref12}; (2) they typically directly reconstruct the original abnormal image into a pseudo-healthy image \citep{ref13,ref16,ref17}. This image-to-image reconstruction approach risks propagating abnormal pixels from the input to the reconstructed output, potentially causing residual anomalies in the pseudo-healthy image and reducing detection sensitivity. 

    In this paper, two novel and effective modules are proposed to address these two limitations, forming a new reconstruction framework. (1) A disentangled representation module is proposed to improve detection generalizability. It explicitly decouples the brain MRI into imaging representation and imaging-invariant anatomical representation (image), enabling the reconstruction to focus on the anatomical image and enhance generalizability across multi-modality and multi-center MRIs. This module also exploits brain anatomical priors to constrain the disentanglement results, thereby generating semantically appropriate anatomical images. Further, a differentiable one-hot encoding operator is proposed to enhance disentanglement stability by deterministically activating the maximum-probability channels during anatomical representation binarization and eliminating activation failures and non-maximum activation noise. (2) An edge-to-image restoration module is designed to reconstruct high-quality PHI in two stages. First, it restores the abstract anatomical representation (anatomical code, which is necessary representation for recoupling the imaging representation) from the high-frequency edge information of abnormal anatomical images, and then reconstructs the imaging-aligned PHI by recoupling the disentangled imaging representation. This module suppresses abnormal residuals in the PHI by using only edge information to limit the abnormal pixels input, and effectively reconstructs the normal regions by utilizing the personalized structural information preserved in the edges. To achieve this, an anatomical code consistency loss is designed to improve the restored image quality by enhancing the model’s learning of personalized edge information and healthy anatomical distribution. In summary, our main contributions are as follows: 
	\begin{itemize}
		\item We propose an unsupervised anomaly detection framework harnessing two innovative modules to overcome the generalizability and performance limitations faced by existing reconstruction-based methods.
		\item The proposed disentangled representation module separates disturbing imaging information and enables the reconstruction to focus on imaging-invariant anatomical information, thus enhancing generalizability to different imaging sequences and multi-center MRIs.
		\item The proposed edge-to-image restoration module effectively utilizes anatomical edge information to restore pseudo-healthy images, thereby significantly reducing residual anomalies in pseudo-healthy images, while simultaneously leveraging personalized information in the edges to enhance normal region details.
	\end{itemize}
    
	After being trained on paired (T1 and T2) scans, the proposed model can perform anomaly detection using only a single (either T1 or T2) scan as input during inference. We have extensively evaluated our method and compared it with 17 state-of-the-art (SOTA) methods on nine multi-modality datasets consisting of 4,443 patients' 3D brain MRIs collected from multiple centers, covering five pathologies: glioma, meningioma, metastases, multiple sclerosis (MS), and stroke. The experimental results demonstrate that the proposed method has 18.32\% (36.62\% vs. 54.94\%) and 13.64\% (31.53\% vs. 45.17\%) considerable absolute improvements compared with the existing SOTA methods in average precision (AP) and Dice similarity coefficient (DSC) metrics, respectively.

\section{Related works}
	According to the different anomaly discrimination spaces, unsupervised anomaly detection methods can be mainly divided into three categories: (1) Self-supervised learning-based (SSL-based) methods, which directly predict the anomaly probability in the output space, (2) Deep feature embedding-based methods, which discriminate anomalies in the embedding space, and (3) Reconstruction-based methods, which identify anomalies in the image space.
	\subsection{SSL-based Methods}
	SSL-based methods use synthetic anomalies to train discriminative models in a supervised manner \citep{ref8,ref9}. For example, DRAEM \citep{ref8} combines the anomaly shape generated by Perlin noise and the anomaly texture to simulate anomalies. FPI \citep{ref9}, which synthesizes artificial anomalies on healthy images by interpolating between normal patches, has been shown to be effective at detecting different artificial anomalies in brain MRI and real lesions in the DeepLesion computed tomography (CT) dataset \citep{yan2018deeplesion}. These methods can train discriminative detection models without real anomaly samples and annotations, and can work well to detect specific anomalies similar to synthetic anomalies. However, they often struggle to detect various real brain anomalies that are unlike synthetic anomalies.
	\subsection{Deep Feature Embedding-based Methods}
	Deep feature embedding-based methods infer anomalies by analyzing abstract representations in the embedding space \citep{ref10,ref11}. These methods utilize feature-level discriminative information to improve anomaly detection, for instance, by comparing features of the test image with features of the normal set, or by calculating the differences between features extracted from teacher and student networks. Specifically, PatchCore \citep{ref10} detects anomalies by measuring the distance between the features of each patch and a memory bank of nominal patch-features. EfficientAD \citep{ref11} measures the feature differences between the teacher network and the student network with lightweight feature extractor. Other methods adopt different strategies. AMCons \citep{ref28}, an attention map-based method with inequality constraints, identifies anomalies through high activation signals and has been proven effective for detecting high-contrast tumors in brain MRI and high-signal intracranial hemorrhages in brain CT. UTRAD \citep{ref27} utilizes a U-transformer to identify anomalies by computing the difference between original and reconstructed features, and its effectiveness has been validated on different medical images, including brain MRI and head CT. However, these methods usually struggle to distinguish features of abnormal regions and personalized normal regions, and face severe limitations in highly personalized brain MRIs. 
	\subsection{Reconstruction-based Methods}
	Reconstruction-based methods reconstruct abnormal images into PHIs without anomalies by learning the normative distribution of normal samples, and highlight pixel-level anomaly detection results through reconstruction errors \citep{ref13,ref12}. Logically, reconstruction-based methods can accurately detect various anomalies if the PHIs can be reconstructed with high quality. This has motivated many methods to focus on how to better reconstruct PHIs. As a representative method, ProxyAno \citep{ref12} uses the superpixel-image as an intermediate proxy to bridge the input and the reconstructed images, which avoids identity mapping and improves detection performance. However, its reconstructed images are prone to lose personalized brain details, leading to false positives. In contrast, DAE \citep{ref13} trains a denoising model using coarse Gaussian noise to enhance its ability to suppress brain anomalies and preserve personalized details. However, it often has difficulty suppressing anomalies that are dissimilar to the coarse noise, limiting its detection performance.
    
    Additionally, many reconstruction-based methods have employed generative models to synthesize pseudo-healthy images (PHIs), such as the variational autoencoder (VAE) \citep{kingma2013auto}, generative adversarial network (GAN) \citep{goodfellow2014generative}, and denoising diffusion probabilistic model (DDPM) \citep{ho2020denoising}. Specifically, one early VAE-based method \citep{ref14} directly utilizes a VAE model to generate PHIs. The more recent PHANES \citep{bercea2023reversing} adopts a two-stage approach, first uses a VAE model to generate a coarse anomaly mask and then employs AOT-GAN \citep{zeng2022aggregated} to inpaint the masked region. f-AnoGAN \citep{schlegl2019f} trains a GAN to learn the manifold of normal images and detects anomalies by combining image reconstruction error and discriminator feature residual error. To enhance the quality of the PHIs generated by DDPM, AnoDDPM \citep{wyatt2022anoddpm} replaces standard Gaussian noise with multi-scale simplex noise and introduces a partial diffusion strategy. To control noise granularity, AutoDDPM \citep{bercea2023mask} employs a two-stage diffusion process that first generates a coarse anomaly mask with high-level noise, followed by iterative stitching and low-level noise re-sampling. While these generative model-based methods typically suppress brain anomalies well, they still face challenges in accurately reconstructing the personalized brain details of normal regions.
    
    In summary, these image-to-image reconstruction methods, which directly reconstruct PHIs from original abnormal images, often face challenges in simultaneously suppressing anomalies while reconstructing normal details. In addition, they tend to overfit the imaging-related information in the normal training MRIs, resulting in limited generalizability to different imaging sequences and multi-center MRIs. Although the recent FedDis \citep{ref5} combines disentangled representation with anomaly detection to alleviate statistical heterogeneity between different centers, it operates on single-modality images and still relies on an image-to-image reconstruction approach. 
    
    To address these challenges, we propose a novel edge-to-image restoration module that effectively suppress anomalies while leveraging personalized information in edges to enhance details in normal regions. Furthermore, the proposed disentangled representation module separates imaging information and enables the reconstruction to focus on imaging-invariant anatomical information, thus enhancing generalizability to different imaging sequences and multi-center MRIs.
	\section{Methods}
	\begin{figure*}[ht]
		\centerline{\includegraphics[width=7.0in]{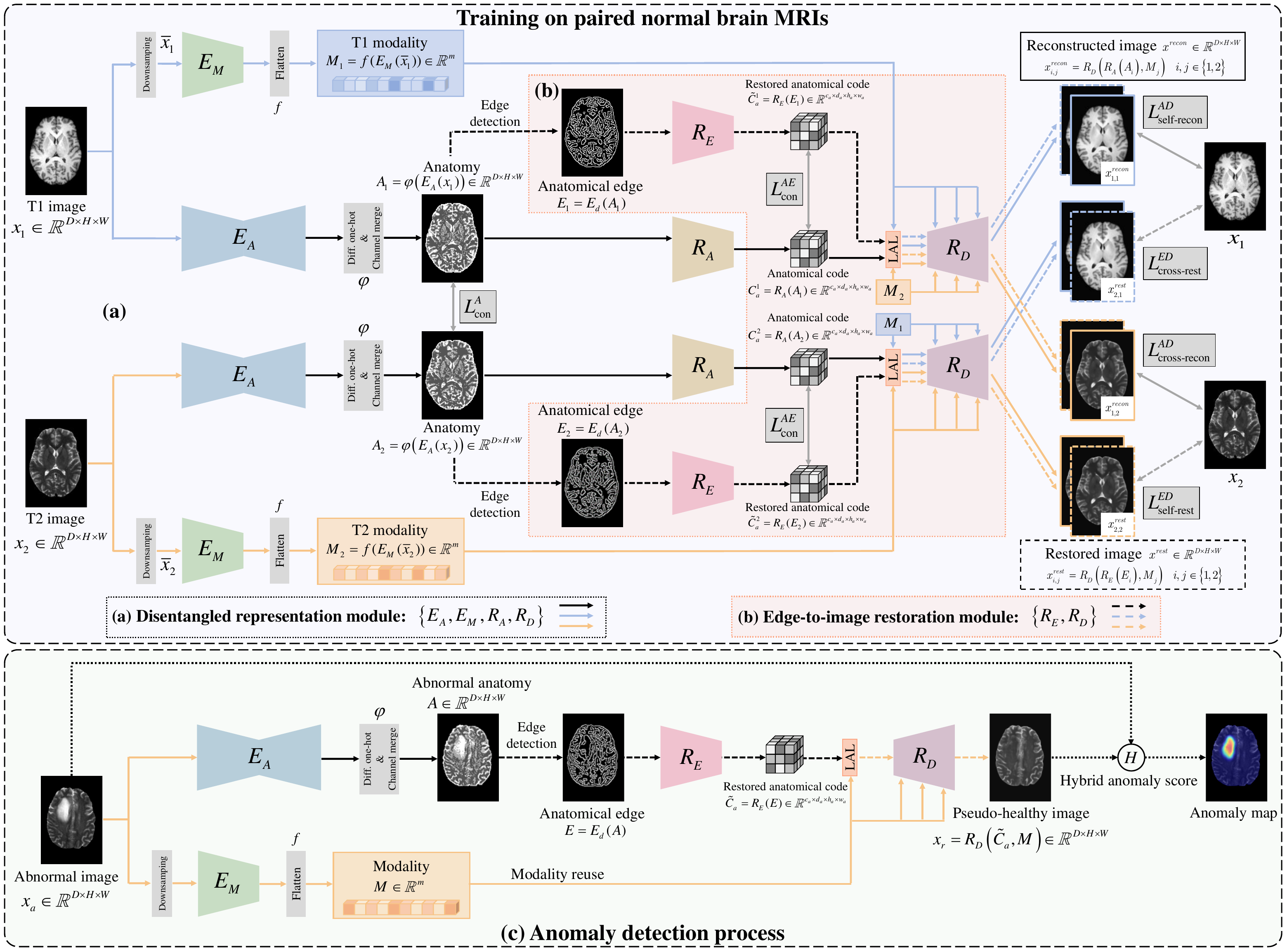}}
		\caption{\textcolor{black}{The architecture diagram of the training and detection process of the proposed two modules: (a) disentangled representation module and (b) edge-to-image restoration module. These two modules are jointly trained on paired normal 3D brain MRIs and reconstruct abnormal images into pseudo-healthy images. The proposed hybrid anomaly score locates abnormal regions by measuring the pixel-level reconstruction differences.}}
		\label{fig1}
	\end{figure*}
    \subsection{Overview}
     The disentangled representation module (DRM), the edge-to-image restoration module (EIRM), and the pseudo-healthy images (PHIs) reconstruction framework formed by these two modules are illustrated in Fig. \ref{fig1} (a), (b), and (c), respectively. For clarity and easy reference, we summarize all key notations, modules, and loss functions used throughout our paper in \ref{Notation summary}.
     
     In the DRM, an anatomy extractor $E_A$ and a modality extractor $E_M$ are trained to decouple (T1 and T2) MRI into a non-spatial imaging representation (vector) and an imaging-invariant spatial anatomical representation (image), respectively. In the EIRM, an anatomy restorer $R_E$ is trained to restore the edges of anatomical images into an abstract anatomical representation (anatomical code) by explicitly aligning its restored representation with the one encoded directly from anatomical images by a jointly trained anatomy encoder $R_A$. Additionally, a representation combiner $R_D$ is trained to reconstruct imaging-aligned PHI by recoupling the restored anatomical code with the disentangled imaging representation across multiple feature levels. During the detection process, abnormal images are reconstructed into PHIs using the trained $E_A$, $E_M$, $R_E$, and $R_D$, and their differences are analyzed using the proposed hybrid anomaly score to detect anomalies. 

     The DRM and EIRM modules are jointly trained on paired (T1 and T2) normal 3D brain MRIs under three common constraints, including shared disentangled anatomical images, shared representation combiner $R_D$, and output alignment between the anatomy encoder $R_A$ and the anatomy restorer $R_E$. This joint design of modules ensures that: (1) $R_E$ focuses on imaging-independent anatomical learning and restoration, relying only on anatomical edge information, as detailed in Sec.~\ref{Analysis for Anatomy Restorer}; (2) the trained model essentially relies on anatomical reconstruction differences to detect anomalies, independent of imaging information (modality); (3) $R_D$ can dynamically reconstruct corresponding modality (T1 or T2) PHI by recoupling disentangled (T1 or T2) imaging information. This enables the trained model to detect anomalies in multi-modality (T1 and T2) MRI, achieve good generalizability in multi-center MRI, and reduce residual anomalies in PHI by inputting only anatomical edge information to prevent abnormal pixels propagation.
	\subsection{Disentangled Representation Module}
    The DRM utilizes two network branches to decouple anatomical images and modality vectors from brain MRIs, respectively. To generate semantically appropriate anatomical images, the prior that brain tissues are mainly divided into three categories (cerebrospinal fluid, white matter, and gray matter) is first used to constrain semantic categories of anatomical representations. Then, the differentiable one-hot encoding operator is proposed to enhance disentanglement stability by addressing activation failure and noise during anatomical representation binarization. Finally, a weighted channel merging operation is employed to convert the multi-channel binary anatomical representation into a multi-class semantic image. To avoid the decoupled modality leaking the spatial anatomy, we extract compressed and flattened non-spatial modality vectors from the downsampled MRI. Specifically, five prior-based loss functions are constructed to ensure the correctness of disentangled representations.

	\subsubsection{Anatomy Disentanglement}
    We adapt the 3D U-Net as the anatomy extractor $E_A$ to decouple the spatial anatomical representation $A_c= {E_A}(x) $ from the 3D MRI $x \in \mathbb{R}^{D \times H \times W}$. $A_c \in \mathbb{R}^{C \times D \times H \times W}$ represents the continuous semantic distribution of disentangled anatomy, where $C=3$ is set by the brain anatomical prior, constraining the three semantic channels to correspond to the expected three brain tissues.
	
    Binarizing $A_c$ is crucial to prevent it from capturing imaging information because the binarization step decouples the potential association between its continuous probability value (distribution) and the continuous grayscale value (distribution) of the original MRI image, as detailed in \ref{Binarization operation}. However, existing methods suffer from activation failure  \citep{ref4} or non-maximum activation noise  \citep{ref3}. To enhance the disentanglement stability, a differentiable one-hot encoding operator is proposed to deterministically convert $A_c$ into a binary anatomical representation $A_{doh}$, while preserving gradient backpropagation. For each spatial position in $A_c$, the maximum semantic channel along dimension $C$ is set to 1, with others set to 0, mathematically expressed as:
	\begin{equation}
		A_{soft} = \mathop{softmax}\limits_{dim=C} \left( {{A_c}} \right),
	\end{equation}
	\begin{equation}
		A_{hard} = G_{OH}(\mathop{argmax}\limits_{dim=C} \left( A_{soft} \right)),
	\end{equation}
	\begin{equation}
		A_{doh} = A_{hard} - GC\left( A_{soft} \right) + A_{soft},
	\end{equation}
	where $A_{hard}$ denotes the hard one-hot encoded version of $A_c$, numerically equivalent to the output such that $A_{doh} = A_{hard}$. $A_{soft}$ represents the soft-coded probability distribution that enables gradient backpropagation via the chain rule: $\frac{\partial A_{doh}}{\partial A_c} = \frac{\partial A_{doh}}{\partial A_{soft}} \cdot \frac{\partial A_{soft}}{\partial A_c}$. The $softmax$ function normalizes elements along dimension $C$ into a probability distribution, while $argmax$ identifies the index $j \in \{0, 1, ..., C-1\}$ of the maximum value. The operator $G_{OH}(\cdot)$ generates a one-hot encoding where position $j$ is set to 1 and all others to 0. $GC(\cdot)$ denotes the gradient cutoff operation.

	Then, a weighted channel merging operation is utilized to convert the multi-channel binary semantic representation $A_{doh} \in \mathbb{R}^{C \times D \times H \times W}$ into a multi-class semantic image $A \in \mathbb{R}^{D \times H \times W}$ to reflect the overall brain semantic distribution:
	\begin{equation}
		A = \frac{1}{C}\sum\limits_{i = 1}^C {i \cdot A_{doh}^i} \odot {M_{brain}},
	\end{equation}
	where ${A_{doh}} = \left\{ {A_{doh}^i} \right\}_{i = 1}^C$ with $A_{doh}^i \in {\mathbb{R}^{D \times H \times W}}$. The brain mask $M_{brain}\in \mathbb{R}^{D \times H \times W}$ is obtained via a common thresholding operation, $M_{brain} = (x>0)$, where the input image $x \in \mathbb{R}^{D \times H \times W}$ has been pre-processed, including skull stripping and intensity normalization to [0, 1], where the background is set to 0. For simplicity, we express the process of decoupling anatomical images as: $A = \varphi \left( {{E_A}\left( x \right)} \right)$. 
    
    In essence, this anatomy disentanglement module can be regarded as an unsupervised segmentation module that outputs tissue maps for cerebrospinal fluid, white matter, and gray matter.
    The semantic information of the three different tissues is enforced through the following three mechanisms: (1) Semantic categories prior: The number of output channels of the anatomy extractor is set to three, which constrains the semantic categories of the anatomical representation to the three tissue types. (2) Binarization operation: An element-wise binarization is applied to the continuous anatomical probability map (anatomical representation), forcing a hard assignment for each voxel to a specific tissue class and preventing imaging information from leaking into the anatomical representation. (3) Anatomy consistency constraint: An anatomy consistency loss is introduced, as shown in Eq. \ref{eq_a_con}, which forces the consistency of anatomical representations disentangled from paired (T1 and T2) scans.
	\subsubsection{Modality Disentanglement}
    To prevent the modality representation from capturing brain structural information, we first employ an encoder as the modality extractor $E_M$ to compress the spatial dimensions of the downsampled MRI ($\bar{x} \in \mathbb{R}^{D/2 \times H/2 \times W/2}$): $M_d=E_M(\bar{x}) \in \mathbb{R}^{c \times d \times h \times w}$, where $c=256$ denotes the feature channels of the modality representation, $d\ll D$, $h \ll H$, $w \ll W$. Subsequently, $M_d$ is flattened into the vector $M=f(M_d)=f(E_M(\bar{x})) \in \mathbb{R}^{m}$ to remove the spatial information, where $m = c \times d \times h \times w$.
	\subsubsection{Loss Functions for Disentanglement}
    We use multiple prior-based constraints to ensure the rationality of the disentangled representations, as follows:
    \begin{equation}
		\label{eq_a_con}
		L_{\text{con}}^A = \left\| \cos(A_1, A_2) - 1 \right\|_2^2,
	\end{equation}
	\begin{equation}
		\label{eq_m_con}
		L_{\text{con}}^M = \sum\limits_{i}{ \left\| \cos(M^p_i, M^q_i) - 1 \right\|_2^2},
	\end{equation}
    \begin{equation}
		\label{eq_a_sim}
		L^{A}_{\text{sim}}=\sum\limits_{i,j}{\max(0, \cos(A^p_i, A^q_i)-\cos(A^p_i, A^p_j) + \alpha_A)},
	\end{equation}
	\begin{equation}
		\label{eq_m_sim}
		L^{M}_{\text{sim}} = \sum\limits_{i,j}{\max(0, \cos(M^p_i, M^p_j) - \cos(M^p_i, M^q_i) + \alpha_M)},
	\end{equation}
    	\begin{equation}
		\label{eq_recon}
			L_{\text{recon}}^{AD} = \sum_{i,j} \underbrace{\|x_{i,i}^{\text{recon}} - x_i\|_1}_{L_{\text{self-recon}}^{AD}} + \underbrace{\|x_{i,j}^{\text{recon}} - x_i\|_1}_{L_{\text{cross-recon}}^{AD}},
	\end{equation}
    where indices $i,j \in \{1,2\}$, $i$ and $j$ ($i \neq j$) represent images of different sequences, $p$ and $q$ ($p \neq q$) represent different subjects, $\cos(\cdot)$ denotes cosine similarity, $\|\cdot\|_1$ denotes the $L_1$-norm, $\|\cdot\|_2$ denotes the $L_2$-norm.
    
     Specifically, the anatomy consistency loss $L_{\text{con}}^A$ (Eq. \ref{eq_a_con}) enforces identical anatomy for paired images, while the modality consistency loss $L_{\text{con}}^M$ (Eq. \ref{eq_m_con}) ensures consistent modality representation within the same modality images. 
     
     The anatomy similarity loss $L_{\text{sim}}^A$ (Eq. \ref{eq_a_sim}) ensures anatomical representations of paired images are more similar than those of same-modality cross-subject pairs, while modality similarity loss $L_{\text{sim}}^M$ (Eq. \ref{eq_m_sim}) enforces the modality representations of same-modality cross-subject pairs are more similar than those of paired images. The margins $\alpha_A = 0.2$ and $\alpha_M = 0.5$, determined through hyperparameter search, prevent representation collapse by enforcing similarity differences. Specifically, the hyperparameter search was performed from the set \{0, 0.1, 0.2, \dots, 1.0\}. The optimal margin values were selected to maximize the average AP across the T1-weighted and T2-weighted sequences of the abnormal BraTS-GLI (N=1,251) dataset \citep{ref42,ref43}, as detailed in \ref{Margin hyperparameter}.

    The reconstruction loss $L_{\text{recon}}^{AD}$ (Eq. \ref{eq_recon}) consists of a self-reconstruction loss $L_{\text{self-recon}}^{AD}$ and a cross-reconstruction loss $L_{\text{cross-recon}}^{AD}$. $L_{\text{self-recon}}^{AD}$ enforces similarity between images reconstructed from the original disentangled representations and the input images, while $L_{\text{cross-recon}}^{AD}$ constrains the similarity between the image reconstructed from the swapped modality and the paired image to enhance anatomical consistency. The reconstructed image $x^{\text{recon}}_{i,j}$ is synthesized by combining disentangled anatomical representation $A_i$ and modality representation $M_j$ through a reconstruction network $R_{AD}$ composed of $R_A$ and $R_D$: ${x^{\text{recon}}_{i,j}} = R_{AD}(A_i,M_j) = R_{D}(R_A(A_i),M_j)$, where the indices $i,j \in \{1,2\}$ correspond to the anatomical and modality representations of the paired images, respectively. 

	\subsection{Edge-to-Image Restoration Module}
    In EIRM, the anatomical restorer $R_E$ first restores anatomical edges into abstract anatomical representations (anatomical codes) rather than anatomical images, so that the subsequent representation combiner $R_D$ can perform feature-level coupling between the anatomical codes and imaging information to synthesize the PHIs. To achieve this, an anatomical code consistency loss is first proposed to align the anatomical codes restored by $R_E$ with those encoded from the anatomical image by $R_A$. Then, the restored anatomical codes are constrained to reconstruct the (T1 and T2) PHIs well by recoupling with the disentangled imaging information (T1 and T2 modality representations) through $R_D$, and a restoration loss function is proposed for this purpose. Specifically, to generate high-quality, imaging-aligned PHIs, $R_D$ couples the restored anatomical codes with modality representations across multiple feature levels through learnable adaptation layers.

    \subsubsection{Edge-Guided Anatomy Learning and Restoration}
    To enhance anatomical learning of $R_E$, the anatomical code consistency loss $L_{\text{con}}^{AE}$ is proposed to align the anatomical code restored by $R_E$ with the anatomical code encoded by $R_A$. Specifically, $R_E$ restores the anatomical edges $E = E_d(A) \in \mathbb{R}^{D \times H \times W}$, which lack anatomical semantic information, into the anatomical code $\widetilde{C}_a = R_E(E) \in \mathbb{R}^{c_a \times d_a \times h_a \times w_a}$, where $E_d(\cdot)$ denotes Canny edge detection. During training, $R_A$ encodes the healthy anatomy $A$ into an anatomical code $C_a$ containing complete personalized anatomical information. Therefore, $L_{\text{con}}^{AE}$ is proposed to supervise $R_E$ directly and accurately model the normative distributions of healthy brain anatomy and learn personalized information in the edges by aligning $\widetilde{C}_a$ with $C_a$ in feature space:
	\begin{equation}
		\label{eq_accl}
			L_{\text{con}}^{AE} = \left\| \widetilde{C}_a - C_a \right\|_{2}^{2} = \left\| R_E(E) - R_A(A) \right\|_{2}^{2}.
	\end{equation}
    
   \subsubsection{Multi-level Synthesis of Pseudo-Healthy Image}
   To enhance imaging coupling of $R_D$ and anatomical learning of $R_E$, the restoration loss $L^{ED}_{\text{rest}}$, including self-restoration loss $L_{\text{self-rest}}^{ED}$ and cross-restoration loss $L_{\text{cross-rest}}^{ED}$, is proposed to align the restored image (i.e. pseudo-healthy image) with the original and paired image, respectively:
	\begin{equation}
		\label{eq_rest}
		L^{ED}_{\text{rest}} = \sum\limits_{i,j}{\underbrace{\|x_{i,i}^{\text{rest}} - x_i\|_1}_{L_{\text{self-rest}}^{ED}} + \underbrace{\|x_{i,j}^{\text{rest}} - x_i\|_1}_{L_{\text{cross-rest}}^{ED}}},
	\end{equation}
	 where indices $i,j \in \{1,2\}$, $i \neq j$, the restored image $x^{\text{rest}}_{i,j} = R_D(\widetilde{C}^i_a, M_j)$ is synthesized by coupling restored anatomical code $\widetilde{C}^i_a$ and modality representation $M_j$. Inspired by adaptation instance normalization (AdaIN)  \citep{ref35}, we introduce a learnable adaptation layer (LAL) after $\widetilde{C}_a$ and each layer feature of $R_D$ to more effectively combine anatomical features and modality representation across multiple feature levels. In particular, LAL ensures that modality representation influences only the intensity of the restored image, without altering its anatomy. The $k$-th layer $LAL_k$ combines anatomical features $F_k \in \mathbb{R}^{c_k \times d_k \times h_k \times w_k}$ with modality $M \in \mathbb{R}^m$ through:
	\begin{equation}
		{F_k} = (1 + FC_k^{\gamma}(M))(\frac{{{F_k} - \mu ({F_k})}}{\sqrt{{\sigma^2 ({F_k})}}}) + FC_k^{\beta}(M),
	\end{equation}
	where $\mu(F_k),\;\sigma^2(F_k) \in \mathbb{R}^{c_k}$ denote channel-wise statistics computed across spatial dimensions $\{d_k, h_k, w_k\}$. The scaling and shifting parameters $FC_k^{\gamma}(M_i),\;FC_k^{\beta}(M_i) \in {\mathbb{R}^{c_k}}$ are learned via fully connected layer $FC_k$ mapping $\mathbb{R}^m \to \mathbb{R}^{c_k}$.

    \subsection{Overall Loss Function for Joint Training}
	We use the total loss $\mathcal{L}$ to jointly train all the above sub-networks, including $E_A$, $E_M$, $R_A$, $R_E$, and $R_D$:
	\begin{equation}
		\label{eq13}
		\begin{split}
			\mathcal{L} &= \lambda_{rec} (L_{\text{recon}}^{AD} + L_{\text{rest}}^{ED}) + \lambda_{a} (L_{\text{con}}^{A} + L_{\text{sim}}^{A}) \\
			&\quad + \lambda_{m} (L_{\text{con}}^{M} + L_{\text{sim}}^{M}) + \lambda_{ae} L_{\text{con}}^{AE},
		\end{split}
	\end{equation}
	where $L_{\text{con}}^{A}$ and $L_{\text{sim}}^{A}$ (Eq. \ref{eq_a_con} and \ref{eq_a_sim}) are used to train $E_A$. $L_{\text{con}}^{M}$ and $L_{\text{sim}}^{M}$ (Eq. \ref{eq_m_con} and \ref{eq_m_sim}) are used to train $E_M$. $L_{\text{recon}}^{AD}$ (Eq. \ref{eq_recon}) is mainly used to train $R_A$ and $R_D$. $L_{\text{con}}^{AE}$ and $L_{\text{rest}}^{ED}$ (Eq. \ref{eq_accl} and \ref{eq_rest}) are mainly used to train $R_E$ and $R_D$. In our experiments, the weight settings, $\lambda_{rec} = \lambda_m = \lambda_{ae} = 1.0$, and $\lambda_a = 0.1$, were determined through a hyperparameter search. Specifically, the weight $\lambda_{rec}$ was first fixed to 1.0 to serve as a weight baseline and reduce the search space. The remaining weights were then searched from the set \{0.01, 0.1, 1.0, 10.0\}, determined by producing the best average AP across the T1-weighted and T2-weighted sequences of the abnormal BraTS-GLI (N=1,251) dataset \citep{ref42,ref43}.
	\subsection{Anomaly Detection Process}
	\subsubsection{Pseudo-Healthy Image Reconstruction}
	As shown in Fig. \ref{fig1} (c), the trained model reconstructs abnormal (T1 or T2) 3D brain MRI $x_a \in \mathbb{R}^{D \times H \times W}$ into imaging-aligned pseudo-healthy images $x_r \in \mathbb{R}^{D \times H \times W}$ with personalized details: 
	\begin{equation}
		x_r = R_{D}(R_E(E_d(\varphi (E_A(x_a)))),f(E_M(\bar{x}_a))),
	\end{equation}
    where $\bar{x}_a$ is the downsampled version of $x_a$, which is used to extract the modality representation.
    
	Specifically, $x_a$ is first disentangled into a anatomy image and a modality vector by the anatomy extractor $E_A$ and modality extractor $E_M$, respectively. Then, the anatomy restorer $R_E$ restores the personalized pseudo-healthy anatomical code based on edges extracted from the anatomy image without being affected by the imaging information. Finally, the representation combiner $R_D$ reconstructs the pseudo-healthy image with personalized details and aligned imaging information using the restored anatomical code and reused modality representation.
	\subsubsection{Hybrid Anomaly Score}
	 We propose the hybrid anomaly score that incorporates structural similarity (SSIM) \citep{ref38} to enhance structure-based difference measurement between $x_a$ and $x_r$ from different aspects of luminance, contrast, and structural information:
	\begin{equation}
		\label{eq15}
		\Delta_H(x_a, x_r) = |x_a - x_r| \cdot (1 - SSIM(x_a, x_r)).
	\end{equation}
     
     The original abnormal MRI $x_a$ often contains grainy noise, while its abnormal regions exhibit non-uniform grayscale values. The pseudo-healthy image $x_r$ generated by our edge-to-image restoration model tends to smooth this noise, which introduces small false positives in the resulting anomaly map $\Delta_H(x_a, x_r)$. Therefore, our post-processing pipeline first applies a minimum filter to remove these small false positives, followed by a mean filter to smooth the anomaly map for more uniform scores within true abnormal regions. The kernel sizes for the minimum and mean filters are set to \(3 \times 3 \times 3\) and \(9 \times 9 \times 9\), respectively. These kernel sizes were determined via hyperparameter search within \{0, 3, 5, 7, 9\} for minimum filter and \{0, 3, 5, 7, \dots, 15\} for mean filter, where 0 indicates the filter is not applied, aiming to maximize the average AP on both T1-weighted and T2-weighted sequences across both large-lesion (BraTS-GLI \citep{ref42,ref43}, adult glioma) and small-lesion (MSLUB \citep{ref53}, multiple sclerosis) datasets, as detailed in \ref{Filter size}. Then, the anomaly map is masked by the brain mask eroded with a \(3 \times 3 \times 3\) kernel to suppress brain boundary disturbance. To binarize the final anomaly map, we follow the common practice \citep{cai2025medianomaly,ref15} in unsupervised anomaly detection and choose an optimal threshold that maximizes the DSC for each dataset.
	\section{Experiments}
	\subsection{Experimental Datasets}
	\subsubsection{Dataset Description}
    Ten publicly available multi-modality brain MRI datasets were used in our experiments, as summarized in Table~\ref{tab1}. Among them, the IXI dataset, which contains healthy adult brain images, was used for model training. Since IXI provides only T1 and T2 images, we limited the evaluation on abnormal datasets to these two sequences for consistency. The remaining nine datasets, collected from multiple institutions and centers, include a total of 4,443 subjects with various types of brain anomalies and corresponding pixel-level annotations (i.e., adult glioma, low-quality imaging adult glioma, pediatric glioma, meningioma, metastases, multiple sclerosis, and stroke), comprising 4,443 T1 and 3,788 T2 images. A brief description of each dataset is provided below:
    \begin{table*}[htbp]
    \caption{\textcolor{black}{Summary of Datasets Used for Training and Evaluation. The healthy IXI dataset was used for model training, while the other datasets were used for model evaluation. Sex is abbreviated as M (Male), F (Female), and U (Unknown). Age is reported as mean $\pm$ standard deviation (range). Dashes ('--') indicate unreported information.}}
    \label{tab1}
    \centering
    \setlength{\tabcolsep}{1.5mm}
    \resizebox{\linewidth}{!}{
    \begin{tabular}{@{} l l l l c c l c @{}}
        \toprule
        \textbf{Sequences} & \textbf{Pathology} & \textbf{Dataset} & \textbf{Subjects} & \textbf{\makecell{Sex \\ (M/F/U)}} & \textbf{\makecell{Age}} & \textbf{Scanner Model} & \textbf{\makecell{Field Strength \\ (T)}} \\
        \midrule
        \multirow{19}{*}{T1, T2} & Healthy & IXI\textsuperscript{\ref{fn:ixi_dataset}} & 577 & 249/313/15 & \makecell{48.7$\pm$16.5 \\ (20.0--86.3)} & \makecell[l]{Philips Intera \\ Philips Gyroscan Intera \\ GE} & \makecell{1.5 \\ 3.0 \\ 1.5} \\
        \cmidrule(lr){2-8}
        & \multirow{6}{*}{Adult glioma} & BraTS-GLI \citep{ref42,ref43} & 1,251 & -- & -- & -- & -- \\
        \cmidrule(lr){3-8}
        & & UPenn-GBM \citep{ref45,ref46} & 611 & 367/244/0 & \makecell{63.0$\pm$12.4 \\ (18.7--88.5)} & \makecell[l]{Siemens TrioTim \\ Siemens Verio \\ Siemens Espree} & \makecell{3.0 \\ 3.0 \\ 1.5} \\
        \cmidrule(lr){3-8}
        & & UCSF-PDGM  \citep{ref47,ref48} & 500 & 299/201/0 & \makecell{56.9$\pm$15.0 \\ (17--94)} & GE Discovery MR750 & 3.0 \\
        \cmidrule(lr){2-8}
        & \makecell[l]{Adult glioma \\ (low-quality)} & BraTS-SSA  \citep{ref49} & 60 & -- & -- & \makecell[l]{Siemens Magnetom Essenza \\ Toshiba Vantage Elan \\ Philips Achieva \\ GE SIGNA Explorer \\ GE SIGNA Creator} & \makecell{1.5 \\ 1.5 \\ 1.5 \\ 1.5 \\ 1.5} \\
        \cmidrule(lr){2-8}
        & Pediatric glioma & BraTS-PED \citep{ref50} & 99 & -- & -- & -- & -- \\
        \cmidrule(lr){2-8}
        & Meningioma & BraTS-MEN \citep{ref51,labella2024multi} & 1,000 & -- & -- & -- & -- \\
        \cmidrule(lr){2-8}
        & Metastases & BraTS-MET \citep{ref52} & 237 & -- & -- & -- & -- \\
        \cmidrule(lr){2-8}
        & Multiple sclerosis & MSLUB \citep{ref53} & 30 & 7/23/0 & \makecell{40.0$\pm$11.2 \\ (25--64)} & Siemens Magnetom Trio & 3.0 \\
        \midrule
        T1 & Stroke & ATLAS \citep{ref54} & 655 & -- & -- & \makecell[l]{GE Discovery MR750  \\ Siemens Trio \\ Philips Achieva  \\ Siemens Prisma \\ Siemens Allegra \\ Siemens Magnetom Vision} & \makecell{3.0 \\3.0 \\3.0  \\3.0 \\3.0 \\ 1.5} \\
        \midrule
        \multicolumn{8}{c}{\textbf{Total}: Normal: 577 (T1: 577; T2: 577); Abnormal: 4,443 (T1: 4,443; T2: 3,788)} \\
        \bottomrule
    \end{tabular}
    }
\end{table*}
     \begin{itemize}
    \item \textbf{IXI}\footnote{\url{https://brain-development.org/ixi-dataset/}\label{fn:ixi_dataset}}: This dataset contains T1, T2, PD-weighted, MRA, and diffusion-weighted scans from approximately 600 healthy subjects collected at three hospitals in London. After excluding subjects missing either T1 or T2 scans, we used paired T1 and T2 images from 577 subjects for model training.
    \item \textbf{BraTS-GLI} \citep{ref42,ref43}: This dataset includes multi-institutional scans from 1,470 patients diagnosed with brain diffuse glioma, with available sequences including T1, T2, post-contrast T1-weighted (T1Gd), and T2 fluid-attenuated inversion recovery (T2-FLAIR). We used the T1 and T2 scans from 1,251 annotated subjects in the training split, targeting the detection of the entire tumor, including Gd-enhancing tumor, peritumoral edematous/invaded tissue, and necrotic tumor core.
    \item \textbf{UPenn-GBM} \citep{ref45,ref46}: This dataset consists of scans from 630 subjects diagnosed with de novo glioblastoma (GBM), collected at the University of Pennsylvania Health System. Each subject has T1, T2, T1Gd, and T2-FLAIR scans. We used the paired T1 and T2 images from 611 annotated cases for model evaluation.
    \item \textbf{UCSF-PDGM} \citep{ref47,ref48}: This dataset includes 501 adult subjects with histologically confirmed grade II–IV diffuse glioma. A variety of sequences are available, including T1, T2, T1Gd, T2-FLAIR, diffusion-weighted imaging (DWI), susceptibility-weighted imaging (SWI), 3D arterial spin labeling (ASL), and 2D 55-direction high angular resolution diffusion imaging (HARDI). We used T1 and T2 scans from 500 annotated subjects, excluding one subject due to abnormal labeling.
    \item \textbf{BraTS-SSA} \citep{ref49}: This dataset includes a retrospective cohort of adult pre-operative glioma cases from Sub-Saharan Africa, with T1, T2, T1Gd, and T2-FLAIR scans. Notably, the dataset reflects the use of lower-quality MRI technology, which produces poor image contrast and resolution. We used 60 annotated cases for evaluation.
    \item \textbf{BraTS-PED} \citep{ref50}: This dataset consists of a retrospective cohort of 228 children with high-grade glioma, including astrocytoma and diffuse midline glioma (DMG)/diffuse intrinsic pontine glioma (DIPG). All cases include T1, T2, T1Gd, and T2-FLAIR sequences. We used 99 annotated cases for evaluation.
    \item \textbf{BraTS-MEN} \citep{ref51,labella2024multi}: This dataset contains pre-operative MRI scans from 1,650 meningioma patients collected from major academic medical centers across the United States. We used 1,000 annotated cases for evaluation.
    \item \textbf{BraTS-MET} \citep{ref52}: This dataset comprises 328 treatment-naive brain metastases patients obtained under standard clinical conditions from various institutions, including T1, T2, T1Gd, and T2-FLAIR scans. After excluding one subject with an abnormal T2 scan, we used 237 annotated cases for evaluation.
    \item \textbf{MSLUB} \citep{ref53}: This dataset includes scans and demographic information from 30 patients with multiple sclerosis (MS) collected at the University Medical Centre Ljubljana (UMCL), including T1, T2, contrast-enhanced T1-weighted (T1WKS), and FLAIR images. All 30 annotated cases were used in the experiments.
    \item \textbf{ATLAS} \citep{ref54}: This dataset contains T1-weighted MRI scans from 1,271 stroke patients collected from 44 different research cohorts across 11 countries. We used 655 annotated cases for model evaluation.
    \end{itemize}
	\subsubsection{Data Preprocessing}
    All multi-modality images were first reoriented to the RAI orientation and underwent N4 bias field correction \citep{ref55}. Subsequently, rigid registration to the SRI24 atlas template space \citep{ref58} was performed using the Greedy registration technique\footnote{\url{https://github.com/pyushkevich/greedy}}, where all images were resampled to an isotropic resolution of $1 \times 1 \times 1$ mm$^3$. Skull stripping \citep{thakur2020brain} was then applied followed by center cropping and padding from the original size of $240 \times 240 \times 155$ to $160 \times 224 \times 160$. The N4 bias field correction, rigid registration and skull stripping steps were performed using the Cancer Imaging Phenomics Toolkit (CaPTk) \citep{davatzikos2018cancer}. Next, to mitigate inter-center imaging differences, the intensity distributions of all abnormal datasets were aligned with those of the normal dataset using a piecewise linear histogram matching method \citep{ref59,ref60}. Finally, the image intensities were scaled according to fixed values, followed by clipping the intensity values to the range \([0,1]\).
    \begin{table*}[]
		\caption{\textcolor{black}{Summary of Comparison Methods.}}
		\label{tab2}
		\centering
		\setlength{\tabcolsep}{1.7mm}{
        \resizebox{\linewidth}{!}{
			\begin{tabular}{lll}
				\toprule[1.pt]
				Method & Venue/Year & Description \\ \midrule \midrule
                
                \multicolumn{3}{l}{\textbf{Self-supervised learning-based (SSL-based)}} \\ \midrule 
				 DRAEM \citep{ref8}           & ICCV 2021            & A discriminatively trained model with anomaly simulations      \\
				 FPI \citep{ref9}             & MELBA 2022           & It synthesizes anomalies by interpolating between both normal patches                               \\ \midrule \midrule
                 
				 \multicolumn{3}{l}{\textbf{Deep feature embedding-based}} \\      \midrule   
				 MKD \citep{ref25}             & CVPR 2021            & It distills multi-layer features of an expert network into a simpler cloner network                  \\
				 RD4AD \citep{ref26}           & CVPR 2022            & A novel teacher-student model with reverse distillation                                             \\
				 UTRAD \citep{ref27}           & NN 2022 & It detects anomalies by U-transformer based feature-level reconstruction                                                    \\
				 AMCons \citep{ref28}          & MIA 2022             & An attention maps based method with inequality constraints                                             \\
                 PatchCore \citep{ref10}       & CVPR 2022            & It uses a maximally representative memory bank of nominal patch-features                            \\

                 EfficientAD \citep{ref11}     & WACV 2024            & A student-teacher approach with lightweight feature extractor                                       \\ \midrule \midrule
                 
				\multicolumn{3}{l}{\textbf{Reconstruction-based}} \\ \midrule 
                AE \citep{ref16}              & Brainlesion 2019             & The initial reconstruction-based method using an autoencoder                                               \\
				 VAE \citep{ref14}             & MICCAI 2019          & Reconstruction via variational autoencoder                                                          \\
				 MemAE \citep{ref17}           & ICCV 2019            & It augments the autoencoder with a memory module                                                     \\
				 RIAD \citep{ref18}            & PR 2021              & It removes partial image regions and reconstructs the image from partial inpaintings                \\
                 ProxyAno \citep{ref12}        & TMI 2022             & It uses the intermediate proxy to bridge the input and reconstructed images                            \\
                 AnoDDPM \citep{wyatt2022anoddpm}        & CVPR 2022             & It improves image quality using simplex noise and partial diffusion strategy                           \\
                 PHANES \citep{bercea2023reversing}        & MICCAI 2023             &  It uses a VAE to coarsely mask anomalies and an AOT-GAN to inpaint                           \\
				 DAE \citep{ref13}             & MIA 2023             & It trains denoising model with coarse noise \\
                 AutoDDPM \citep{bercea2023mask}        & IMLH 2023             & It uses a two-stage diffusion process with masking, stitching and re-sampling                            \\
                 \midrule
				 \bottomrule[1.pt]
			\end{tabular}
            }
		}
	\end{table*}
    \subsection{Experiment Setup}
	\subsubsection{Implementation Details} 
    During training, we downsampled the image size from $160 \times 224 \times 160$ to $112 \times 160 \times 112$ through trilinear interpolation to reduce computational costs. We trained the model for 150 epochs with a batch size of 2, consisting of 4 images, T1 and T2 images from 2 different subjects. We used the Adam optimizer with a learning rate of 0.0001. Following previous work \citep{ref40}, we randomly selected 64 subjects from the BraTS-GLI dataset as the validation dataset to select the optimal model with the highest AP for both our method and the comparison methods. We trained our model with automatic mixed precision (AMP) on two 24GB NVIDIA RTX 3090 GPUs.
	\subsubsection{Evaluation Metrics}
    We used two voxel-level metrics to evaluate the detection performance of our method and all comparison models: average precision (AP) and Dice similarity coefficient (DSC). AP is the area under the precision (TP/(TP+FP)-recall (TP/(TP+FN)) curve (AUPRC), where TP, FP, and FN represent voxel-wise true positives, false positives, and false negatives, respectively. The dataset's AP score is the AUPRC of a single curve generated by pooling all voxel-level predictions and labels across all subjects in the dataset. Additionally, we calculated the average DSC for each abnormal dataset, calculated as 2TP/(2TP+FP+FN). Following previous works \citep{cai2025medianomaly,ref15}, we performed a greedy search over the entire abnormal dataset to determine the optimal binarization threshold for the best DSC.
    \subsection{Comparison Methods}
    Using identical experimental settings and trained solely on the normal dataset (IXI\textsuperscript{\ref{fn:ixi_dataset}}), we compared our method with 17 unsupervised pixel-level anomaly detection SOTA methods, as shown in Table \ref{tab2}, including (1) two SSL-based methods, (2) six deep feature embedding-based methods, and (3) nine reconstruction-based methods. These comparison methods were carefully selected based on their strong relevance and performance: they either exhibit the top-ranked performance as documented in recent review articles \citep{cai2025medianomaly} (e.g., DAE \citep{ref13}, AutoDDPM \citep{bercea2023mask} and AnoDDPM \citep{wyatt2022anoddpm}) or represent the Top-1 implementations in relevant competitions \citep{zimmerer2022mood}, such as FPI \citep{ref9}. We have incorporated several approaches that have been published in top-tier journals (or conferences) and the latest SOTA methods, such as PatchCore \citep{ref10} and EfficientAD \citep{ref11}. 
    
    For the implementation of these comparison methods, we utilized the original source code (e.g., FPI \citep{ref9}, AMCons \citep{ref28}, MemAE \citep{ref17}, RIAD \citep{ref18}, AnoDDPM \citep{wyatt2022anoddpm}, PHANES \citep{bercea2023reversing}, DAE \citep{ref13}, AutoDDPM \citep{bercea2023mask}), open-source libraries \citep{ref40,akcay2022anomalib} (e.g., DRAEM \citep{ref8}, MKD \citep{ref25}, RD4AD \citep{ref26}, UTRAD \citep{ref27}, PatchCore \citep{ref10}, EfficientAD \citep{ref11}), and the repository \citep{ref15} (e.g., AE \citep{ref16}, VAE \citep{ref14}). Since there is no open-source code for ProxyAno \citep{ref12}, we reimplemented ProxyAno based on its original paper. We followed the post-processing steps used in their original papers or official implementations.
            \begin{sidewaystable*}[]
\centering
\caption{Experimental Results (AP $\uparrow$ (\%) and DSC $\uparrow$ (\%)). The \red{best}, \bblue{second-best} and \green{third-best} results are highlighted in red, blue and green, respectively. According to relevant literature \citep{cai2025medianomaly,varoquaux2023evaluating}, each experiment was conducted three times with different random seeds, and we report the mean and standard deviation ($\mu\pm\sigma$) for all metrics.}
\label{tab3}
\resizebox{\linewidth}{!}{
\renewcommand{\arraystretch}{1.2} 
\setlength{\tabcolsep}{4pt}
\begin{tabular}{llcccccccccccccccccccc}
\toprule[1.pt]
\multicolumn{2}{l}{\multirow{4}{*}{Method}} & \multicolumn{6}{c}{Adult glioma} & \multicolumn{2}{c}{\makecell{Adult glioma \\ (low-quality)}} & \multicolumn{2}{c}{Pediatric glioma} & \multicolumn{2}{c}{Meningioma} & \multicolumn{2}{c}{Metastases} & \multicolumn{2}{c}{MS} & \multicolumn{1}{c}{Stroke} &  &  &  \\
\cmidrule(lr){3-8}\cmidrule(lr){9-10} \cmidrule(lr){11-12} \cmidrule(lr){13-14} \cmidrule(lr){15-16} \cmidrule(lr){17-18} \cmidrule(lr){19-19}
\multicolumn{2}{c}{} & \multicolumn{2}{c}{BraTS-GLI} & \multicolumn{2}{c}{UPenn-GBM} & \multicolumn{2}{c}{UCSF-PDGM} & \multicolumn{2}{c}{BraTS-SSA} & \multicolumn{2}{c}{BraTS-PED} & \multicolumn{2}{c}{BraTS-MEN} & \multicolumn{2}{c}{BraTS-MET} & \multicolumn{2}{c}{MSLUB} & \multicolumn{1}{c}{ATLAS} & \multicolumn{3}{c}{Average} \\
\multicolumn{2}{c}{} & \multicolumn{2}{c}{(N=1,251)} & \multicolumn{2}{c}{(N=611)} & \multicolumn{2}{c}{(N=500)} & \multicolumn{2}{c}{(N=60)} & \multicolumn{2}{c}{(N=99)} & \multicolumn{2}{c}{(N=1,000)} & \multicolumn{2}{c}{(N=237)} & \multicolumn{2}{c}{(N=30)} & (N=655) & (4,443) & (3,788) & (8,231) \\
\multicolumn{2}{c}{} & T1 & T2 & T1 & T2 & T1 & T2 & T1 & T2 & T1 & T2 & T1 & T2 & T1 & T2 & T1 & T2 & T1 & T1 & T2 & T1\&T2 \\

\midrule \midrule
\multicolumn{22}{l}{\textbf{Self-supervised learning-based (SSL-based)}} \\ \midrule 
 \multirow{2}{*}{DRAEM} & AP & 14.23{\scriptsize$\pm$2.30} & 56.59{\scriptsize$\pm$3.52} & 13.19{\scriptsize$\pm$3.46} & 59.29{\scriptsize$\pm$4.08} & 13.29{\scriptsize$\pm$2.11} & \green{67.06}{\scriptsize$\pm$5.13} & 26.98{\scriptsize$\pm$5.79} & 54.90{\scriptsize$\pm$2.86} & 6.25{\scriptsize$\pm$0.58} & 23.30{\scriptsize$\pm$6.43} & 7.65{\scriptsize$\pm$1.82} & 37.49{\scriptsize$\pm$2.61} & 10.64{\scriptsize$\pm$1.98} & 42.84{\scriptsize$\pm$3.43} & 2.04{\scriptsize$\pm$0.02} & \bblue{6.06}{\scriptsize$\pm$1.26} & 4.17{\scriptsize$\pm$0.81} & 10.94 & 43.44 & 26.23 \\
 & \cellcolor{gray!12}DSC & \cellcolor{gray!12}19.47{\scriptsize$\pm$3.27} & \cellcolor{gray!12}50.12{\scriptsize$\pm$2.92} & \cellcolor{gray!12}19.03{\scriptsize$\pm$4.28} & \cellcolor{gray!12}49.76{\scriptsize$\pm$3.10} & \cellcolor{gray!12}18.83{\scriptsize$\pm$3.67} & \cellcolor{gray!12}\bblue{58.85}{\scriptsize$\pm$5.09} & \cellcolor{gray!12}28.07{\scriptsize$\pm$3.90} & \cellcolor{gray!12}52.07{\scriptsize$\pm$4.13} & \cellcolor{gray!12}11.43{\scriptsize$\pm$1.71} & \cellcolor{gray!12}23.36{\scriptsize$\pm$7.06} & \cellcolor{gray!12}7.64{\scriptsize$\pm$1.52} & \cellcolor{gray!12}19.21{\scriptsize$\pm$1.25} & \cellcolor{gray!12}10.71{\scriptsize$\pm$1.26} & \cellcolor{gray!12}27.05{\scriptsize$\pm$1.94} & \cellcolor{gray!12}4.71{\scriptsize$\pm$0.04} & \cellcolor{gray!12}\bblue{9.72}{\scriptsize$\pm$1.68} & \cellcolor{gray!12}2.12{\scriptsize$\pm$0.73} & \cellcolor{gray!12}13.56 & \cellcolor{gray!12}36.27 & \cellcolor{gray!12}24.24 \\
\multirow{2}{*}{FPI} & AP & \bblue{38.76}{\scriptsize$\pm$1.97} & 54.59{\scriptsize$\pm$4.86} & \green{46.02}{\scriptsize$\pm$3.53} & 54.02{\scriptsize$\pm$3.75} & \bblue{37.52}{\scriptsize$\pm$2.71} & 54.96{\scriptsize$\pm$1.23} & 29.37{\scriptsize$\pm$2.54} & 44.83{\scriptsize$\pm$4.94} & \bblue{25.74}{\scriptsize$\pm$3.83} & 45.48{\scriptsize$\pm$5.72} & 19.77{\scriptsize$\pm$1.78} & 37.21{\scriptsize$\pm$2.56} & 16.14{\scriptsize$\pm$3.06} & 32.96{\scriptsize$\pm$6.49} & 1.68{\scriptsize$\pm$0.18} & 2.58{\scriptsize$\pm$0.59} & 2.15{\scriptsize$\pm$0.21} & \green{24.13} & 40.83 & 31.99 \\
 & \cellcolor{gray!12}DSC & \cellcolor{gray!12}\bblue{35.03}{\scriptsize$\pm$0.81} & \cellcolor{gray!12}51.85{\scriptsize$\pm$4.02} & \cellcolor{gray!12}\green{39.51}{\scriptsize$\pm$1.41} & \cellcolor{gray!12}52.28{\scriptsize$\pm$2.77} & \cellcolor{gray!12}\bblue{32.86}{\scriptsize$\pm$1.26} & \cellcolor{gray!12}51.26{\scriptsize$\pm$1.20} & \cellcolor{gray!12}31.75{\scriptsize$\pm$1.04} & \cellcolor{gray!12}43.97{\scriptsize$\pm$2.44} & \cellcolor{gray!12}22.72{\scriptsize$\pm$0.27} & \cellcolor{gray!12}36.44{\scriptsize$\pm$3.72} & \cellcolor{gray!12}12.77{\scriptsize$\pm$0.37} & \cellcolor{gray!12}23.18{\scriptsize$\pm$1.37} & \cellcolor{gray!12}14.40{\scriptsize$\pm$0.76} & \cellcolor{gray!12}23.91{\scriptsize$\pm$2.83} & \cellcolor{gray!12}2.94{\scriptsize$\pm$0.32} & \cellcolor{gray!12}3.45{\scriptsize$\pm$0.79} & \cellcolor{gray!12}3.16{\scriptsize$\pm$0.37} & \cellcolor{gray!12}21.68 & \cellcolor{gray!12}35.80 & \cellcolor{gray!12}28.32 \\

 \midrule \midrule
\multicolumn{22}{l}{\textbf{Deep feature embedding-based}} \\ \midrule 
\multirow{2}{*}{MKD} & AP & 7.61{\scriptsize$\pm$0.18} & 10.61{\scriptsize$\pm$0.07} & 6.46{\scriptsize$\pm$0.21} & 9.71{\scriptsize$\pm$0.07} & 7.47{\scriptsize$\pm$0.21} & 11.38{\scriptsize$\pm$0.10} & 14.46{\scriptsize$\pm$0.12} & 17.99{\scriptsize$\pm$0.81} & 10.51{\scriptsize$\pm$0.06} & 12.17{\scriptsize$\pm$0.57} & 3.82{\scriptsize$\pm$0.13} & 5.35{\scriptsize$\pm$0.13} & 5.17{\scriptsize$\pm$0.27} & 7.07{\scriptsize$\pm$0.20} & 1.93{\scriptsize$\pm$0.02} & 1.84{\scriptsize$\pm$0.05} & 1.95{\scriptsize$\pm$0.13} & 6.60 & 9.51 & 7.97 \\
 & \cellcolor{gray!12}DSC & \cellcolor{gray!12}13.78{\scriptsize$\pm$0.08} & \cellcolor{gray!12}17.09{\scriptsize$\pm$0.08} & \cellcolor{gray!12}12.11{\scriptsize$\pm$0.04} & \cellcolor{gray!12}15.74{\scriptsize$\pm$0.14} & \cellcolor{gray!12}12.91{\scriptsize$\pm$0.09} & \cellcolor{gray!12}17.33{\scriptsize$\pm$0.19} & \cellcolor{gray!12}21.99{\scriptsize$\pm$0.13} & \cellcolor{gray!12}24.93{\scriptsize$\pm$0.80} & \cellcolor{gray!12}15.28{\scriptsize$\pm$0.35} & \cellcolor{gray!12}19.11{\scriptsize$\pm$0.83} & \cellcolor{gray!12}5.95{\scriptsize$\pm$0.36} & \cellcolor{gray!12}8.15{\scriptsize$\pm$0.15} & \cellcolor{gray!12}8.12{\scriptsize$\pm$0.04} & \cellcolor{gray!12}10.62{\scriptsize$\pm$0.21} & \cellcolor{gray!12}3.91{\scriptsize$\pm$0.06} & \cellcolor{gray!12}3.89{\scriptsize$\pm$0.06} & \cellcolor{gray!12}3.69{\scriptsize$\pm$0.10} & \cellcolor{gray!12}10.86 & \cellcolor{gray!12}14.61 & \cellcolor{gray!12}12.62 \\
\multirow{2}{*}{AMCons} & AP & 5.86{\scriptsize$\pm$0.11} & 45.60{\scriptsize$\pm$0.58} & 4.74{\scriptsize$\pm$0.11} & 42.54{\scriptsize$\pm$0.56} & 5.25{\scriptsize$\pm$0.11} & 38.24{\scriptsize$\pm$0.66} & 9.96{\scriptsize$\pm$0.24} & 59.31{\scriptsize$\pm$0.45} & 5.89{\scriptsize$\pm$0.07} & 35.75{\scriptsize$\pm$0.51} & 2.60{\scriptsize$\pm$0.06} & 19.86{\scriptsize$\pm$0.48} & 3.92{\scriptsize$\pm$0.12} & 26.72{\scriptsize$\pm$0.46} & 2.56{\scriptsize$\pm$0.08} & 3.02{\scriptsize$\pm$0.06} & 1.16{\scriptsize$\pm$0.00} & 4.66 & 33.88 & 18.41 \\
 & \cellcolor{gray!12}DSC & \cellcolor{gray!12}13.20{\scriptsize$\pm$0.08} & \cellcolor{gray!12}42.01{\scriptsize$\pm$0.19} & \cellcolor{gray!12}11.64{\scriptsize$\pm$0.07} & \cellcolor{gray!12}40.63{\scriptsize$\pm$0.15} & \cellcolor{gray!12}11.98{\scriptsize$\pm$0.07} & \cellcolor{gray!12}38.28{\scriptsize$\pm$0.16} & \cellcolor{gray!12}20.40{\scriptsize$\pm$0.04} & \cellcolor{gray!12}55.32{\scriptsize$\pm$0.14} & \cellcolor{gray!12}8.12{\scriptsize$\pm$0.10} & \cellcolor{gray!12}39.89{\scriptsize$\pm$0.05} & \cellcolor{gray!12}5.92{\scriptsize$\pm$0.02} & \cellcolor{gray!12}15.52{\scriptsize$\pm$0.10} & \cellcolor{gray!12}8.10{\scriptsize$\pm$0.04} & \cellcolor{gray!12}20.42{\scriptsize$\pm$0.13} & \cellcolor{gray!12}5.47{\scriptsize$\pm$0.27} & \cellcolor{gray!12}5.87{\scriptsize$\pm$0.06} & \cellcolor{gray!12}3.35{\scriptsize$\pm$0.00} & \cellcolor{gray!12}9.80 & \cellcolor{gray!12}32.24 & \cellcolor{gray!12}20.36 \\
\multirow{2}{*}{RD4AD} & AP & 24.89{\scriptsize$\pm$1.87} & 36.13{\scriptsize$\pm$2.21} & 31.34{\scriptsize$\pm$3.88} & 37.59{\scriptsize$\pm$3.24} & 25.28{\scriptsize$\pm$1.88} & 33.78{\scriptsize$\pm$2.42} & 27.43{\scriptsize$\pm$0.76} & 49.67{\scriptsize$\pm$0.97} & 14.83{\scriptsize$\pm$0.71} & 28.78{\scriptsize$\pm$0.47} & 11.98{\scriptsize$\pm$1.10} & 20.44{\scriptsize$\pm$2.08} & 13.64{\scriptsize$\pm$1.63} & 28.83{\scriptsize$\pm$4.52} & 4.58{\scriptsize$\pm$0.49} & 3.26{\scriptsize$\pm$0.10} & \green{16.27}{\scriptsize$\pm$2.85} & 18.92 & 29.81 & 24.04 \\
 & \cellcolor{gray!12}DSC & \cellcolor{gray!12}32.41{\scriptsize$\pm$1.03} & \cellcolor{gray!12}37.89{\scriptsize$\pm$0.45} & \cellcolor{gray!12}36.60{\scriptsize$\pm$1.80} & \cellcolor{gray!12}38.64{\scriptsize$\pm$0.39} & \cellcolor{gray!12}\green{31.03}{\scriptsize$\pm$0.85} & \cellcolor{gray!12}35.75{\scriptsize$\pm$0.86} & \cellcolor{gray!12}34.27{\scriptsize$\pm$0.69} & \cellcolor{gray!12}48.53{\scriptsize$\pm$0.22} & \cellcolor{gray!12}\green{28.98}{\scriptsize$\pm$1.70} & \cellcolor{gray!12}34.43{\scriptsize$\pm$0.58} & \cellcolor{gray!12}10.68{\scriptsize$\pm$0.23} & \cellcolor{gray!12}14.72{\scriptsize$\pm$0.36} & \cellcolor{gray!12}13.22{\scriptsize$\pm$0.39} & \cellcolor{gray!12}19.54{\scriptsize$\pm$0.38} & \cellcolor{gray!12}6.31{\scriptsize$\pm$0.89} & \cellcolor{gray!12}4.26{\scriptsize$\pm$0.34} & \cellcolor{gray!12}\green{13.44}{\scriptsize$\pm$0.50} & \cellcolor{gray!12}22.99 & \cellcolor{gray!12}29.22 & \cellcolor{gray!12}25.92 \\
\multirow{2}{*}{UTRAD} & AP & 26.42{\scriptsize$\pm$2.82} & 34.75{\scriptsize$\pm$1.10} & 34.53{\scriptsize$\pm$3.26} & 39.63{\scriptsize$\pm$2.12} & 29.88{\scriptsize$\pm$3.23} & 35.40{\scriptsize$\pm$1.68} & \green{35.64}{\scriptsize$\pm$0.99} & 40.89{\scriptsize$\pm$0.65} & 9.60{\scriptsize$\pm$1.23} & 18.82{\scriptsize$\pm$1.54} & \bblue{23.08}{\scriptsize$\pm$2.45} & 28.11{\scriptsize$\pm$2.72} & 21.41{\scriptsize$\pm$2.43} & 32.85{\scriptsize$\pm$1.35} & 4.45{\scriptsize$\pm$0.40} & 4.27{\scriptsize$\pm$0.31} & 10.32{\scriptsize$\pm$0.64} & 21.70 & 29.34 & 25.30 \\
 & \cellcolor{gray!12}DSC & \cellcolor{gray!12}27.38{\scriptsize$\pm$1.84} & \cellcolor{gray!12}32.56{\scriptsize$\pm$2.04} & \cellcolor{gray!12}30.56{\scriptsize$\pm$2.03} & \cellcolor{gray!12}34.21{\scriptsize$\pm$2.06} & \cellcolor{gray!12}24.20{\scriptsize$\pm$3.35} & \cellcolor{gray!12}28.91{\scriptsize$\pm$2.07} & \cellcolor{gray!12}\green{34.42}{\scriptsize$\pm$0.63} & \cellcolor{gray!12}39.82{\scriptsize$\pm$0.27} & \cellcolor{gray!12}15.57{\scriptsize$\pm$1.82} & \cellcolor{gray!12}26.94{\scriptsize$\pm$2.04} & \cellcolor{gray!12}10.27{\scriptsize$\pm$0.85} & \cellcolor{gray!12}13.06{\scriptsize$\pm$0.27} & \cellcolor{gray!12}15.12{\scriptsize$\pm$2.10} & \cellcolor{gray!12}18.33{\scriptsize$\pm$0.94} & \cellcolor{gray!12}5.32{\scriptsize$\pm$0.48} & \cellcolor{gray!12}5.28{\scriptsize$\pm$0.18} & \cellcolor{gray!12}6.40{\scriptsize$\pm$0.70} & \cellcolor{gray!12}18.80 & \cellcolor{gray!12}24.89 & \cellcolor{gray!12}21.67 \\
\multirow{2}{*}{PatchCore} & AP & \green{33.57}{\scriptsize$\pm$0.09} & 54.19{\scriptsize$\pm$0.06} & \bblue{46.73}{\scriptsize$\pm$0.08} & 59.42{\scriptsize$\pm$0.07} & \green{33.08}{\scriptsize$\pm$0.12} & 53.32{\scriptsize$\pm$0.09} & 32.69{\scriptsize$\pm$0.05} & 54.70{\scriptsize$\pm$0.14} & 15.30{\scriptsize$\pm$0.02} & 39.95{\scriptsize$\pm$0.06} & \green{22.99}{\scriptsize$\pm$0.13} & 43.95{\scriptsize$\pm$0.06} & \green{21.62}{\scriptsize$\pm$0.10} & \green{50.24}{\scriptsize$\pm$0.06} & 4.54{\scriptsize$\pm$0.03} & 4.45{\scriptsize$\pm$0.04} & 15.94{\scriptsize$\pm$0.05} & \bblue{25.16} & 45.03 & \green{34.51} \\
 & \cellcolor{gray!12}DSC & \cellcolor{gray!12}\green{33.99}{\scriptsize$\pm$0.06} & \cellcolor{gray!12}46.37{\scriptsize$\pm$0.06} & \cellcolor{gray!12}\bblue{40.12}{\scriptsize$\pm$0.01} & \cellcolor{gray!12}47.84{\scriptsize$\pm$0.06} & \cellcolor{gray!12}30.62{\scriptsize$\pm$0.13} & \cellcolor{gray!12}43.28{\scriptsize$\pm$0.03} & \cellcolor{gray!12}33.39{\scriptsize$\pm$0.12} & \cellcolor{gray!12}51.96{\scriptsize$\pm$0.06} & \cellcolor{gray!12}27.57{\scriptsize$\pm$0.16} & \cellcolor{gray!12}43.86{\scriptsize$\pm$0.09} & \cellcolor{gray!12}13.22{\scriptsize$\pm$0.04} & \cellcolor{gray!12}21.01{\scriptsize$\pm$0.05} & \cellcolor{gray!12}\green{16.31}{\scriptsize$\pm$0.07} & \cellcolor{gray!12}27.28{\scriptsize$\pm$0.06} & \cellcolor{gray!12}5.40{\scriptsize$\pm$0.05} & \cellcolor{gray!12}3.52{\scriptsize$\pm$0.05} & \cellcolor{gray!12}9.65{\scriptsize$\pm$0.04} & \cellcolor{gray!12}\green{23.37} & \cellcolor{gray!12}35.64 & \cellcolor{gray!12}29.14 \\
\multirow{2}{*}{EfficientAD} & AP & 30.11{\scriptsize$\pm$0.23} & \green{64.05}{\scriptsize$\pm$1.46} & 40.05{\scriptsize$\pm$1.00} & \green{66.26}{\scriptsize$\pm$2.01} & 25.64{\scriptsize$\pm$0.27} & \bblue{67.45}{\scriptsize$\pm$1.91} & \bblue{39.84}{\scriptsize$\pm$1.05} & \green{64.08}{\scriptsize$\pm$0.83} & 8.75{\scriptsize$\pm$0.23} & \green{48.52}{\scriptsize$\pm$1.19} & 21.43{\scriptsize$\pm$0.34} & \bblue{48.90}{\scriptsize$\pm$3.00} & \bblue{25.89}{\scriptsize$\pm$1.25} & \bblue{52.11}{\scriptsize$\pm$2.32} & 3.15{\scriptsize$\pm$0.33} & 2.47{\scriptsize$\pm$0.42} & 13.80{\scriptsize$\pm$0.79} & 23.18 & \bblue{51.73} & \bblue{36.62} \\
 & \cellcolor{gray!12}DSC & \cellcolor{gray!12}32.25{\scriptsize$\pm$0.92} & \cellcolor{gray!12}\green{53.57}{\scriptsize$\pm$1.42} & \cellcolor{gray!12}37.12{\scriptsize$\pm$0.84} & \cellcolor{gray!12}\green{53.50}{\scriptsize$\pm$1.94} & \cellcolor{gray!12}27.02{\scriptsize$\pm$0.71} & \cellcolor{gray!12}\green{56.19}{\scriptsize$\pm$1.91} & \cellcolor{gray!12}\bblue{40.29}{\scriptsize$\pm$1.03} & \cellcolor{gray!12}\bblue{61.64}{\scriptsize$\pm$1.07} & \cellcolor{gray!12}24.02{\scriptsize$\pm$0.88} & \cellcolor{gray!12}\green{46.83}{\scriptsize$\pm$1.11} & \cellcolor{gray!12}\bblue{15.15}{\scriptsize$\pm$0.50} & \cellcolor{gray!12}\green{23.21}{\scriptsize$\pm$1.38} & \cellcolor{gray!12}\bblue{17.44}{\scriptsize$\pm$0.49} & \cellcolor{gray!12}\bblue{29.09}{\scriptsize$\pm$1.13} & \cellcolor{gray!12}4.85{\scriptsize$\pm$0.53} & \cellcolor{gray!12}3.28{\scriptsize$\pm$0.70} & \cellcolor{gray!12}10.64{\scriptsize$\pm$0.31} & \cellcolor{gray!12}23.20 & \cellcolor{gray!12}\bblue{40.91} & \cellcolor{gray!12}\bblue{31.53} \\

 \midrule \midrule
\multicolumn{22}{l}{\textbf{Reconstruction-based}} \\ \midrule
\multirow{2}{*}{AE} & AP & 5.84{\scriptsize$\pm$0.19} & 10.15{\scriptsize$\pm$0.35} & 4.93{\scriptsize$\pm$0.21} & 9.53{\scriptsize$\pm$0.51} & 5.34{\scriptsize$\pm$0.17} & 7.05{\scriptsize$\pm$0.31} & 9.13{\scriptsize$\pm$0.26} & 26.27{\scriptsize$\pm$4.14} & 3.80{\scriptsize$\pm$0.11} & 9.77{\scriptsize$\pm$0.56} & 2.75{\scriptsize$\pm$0.13} & 4.63{\scriptsize$\pm$0.33} & 3.45{\scriptsize$\pm$0.02} & 6.75{\scriptsize$\pm$0.72} & 1.64{\scriptsize$\pm$0.07} & 1.58{\scriptsize$\pm$0.16} & 2.38{\scriptsize$\pm$0.15} & 4.36 & 9.47 & 6.76 \\
 & \cellcolor{gray!12}DSC & \cellcolor{gray!12}12.76{\scriptsize$\pm$0.05} & \cellcolor{gray!12}18.44{\scriptsize$\pm$0.50} & \cellcolor{gray!12}11.25{\scriptsize$\pm$0.11} & \cellcolor{gray!12}16.99{\scriptsize$\pm$0.62} & \cellcolor{gray!12}11.57{\scriptsize$\pm$0.06} & \cellcolor{gray!12}14.28{\scriptsize$\pm$0.29} & \cellcolor{gray!12}19.09{\scriptsize$\pm$0.04} & \cellcolor{gray!12}32.70{\scriptsize$\pm$2.99} & \cellcolor{gray!12}7.88{\scriptsize$\pm$0.05} & \cellcolor{gray!12}17.18{\scriptsize$\pm$0.93} & \cellcolor{gray!12}5.82{\scriptsize$\pm$0.02} & \cellcolor{gray!12}7.94{\scriptsize$\pm$0.26} & \cellcolor{gray!12}7.50{\scriptsize$\pm$0.01} & \cellcolor{gray!12}11.29{\scriptsize$\pm$0.27} & \cellcolor{gray!12}2.84{\scriptsize$\pm$0.03} & \cellcolor{gray!12}3.33{\scriptsize$\pm$0.18} & \cellcolor{gray!12}2.88{\scriptsize$\pm$0.35} & \cellcolor{gray!12}9.07 & \cellcolor{gray!12}15.27 & \cellcolor{gray!12}11.98 \\
\multirow{2}{*}{MemAE} & AP & 7.41{\scriptsize$\pm$0.19} & 19.07{\scriptsize$\pm$1.01} & 6.54{\scriptsize$\pm$0.41} & 18.33{\scriptsize$\pm$0.85} & 6.32{\scriptsize$\pm$0.31} & 12.32{\scriptsize$\pm$1.21} & 10.96{\scriptsize$\pm$0.22} & 36.23{\scriptsize$\pm$2.70} & 5.40{\scriptsize$\pm$0.19} & 17.54{\scriptsize$\pm$1.02} & 3.37{\scriptsize$\pm$0.18} & 9.04{\scriptsize$\pm$0.29} & 4.12{\scriptsize$\pm$0.52} & 15.00{\scriptsize$\pm$0.53} & 1.68{\scriptsize$\pm$0.89} & 1.83{\scriptsize$\pm$0.10} & 2.08{\scriptsize$\pm$0.08} & 5.32 & 16.17 & 10.43 \\
 & \cellcolor{gray!12}DSC & \cellcolor{gray!12}12.69{\scriptsize$\pm$0.40} & \cellcolor{gray!12}23.67{\scriptsize$\pm$0.87} & \cellcolor{gray!12}11.25{\scriptsize$\pm$0.39} & \cellcolor{gray!12}22.71{\scriptsize$\pm$1.08} & \cellcolor{gray!12}11.53{\scriptsize$\pm$0.22} & \cellcolor{gray!12}18.88{\scriptsize$\pm$1.20} & \cellcolor{gray!12}19.09{\scriptsize$\pm$0.07} & \cellcolor{gray!12}36.97{\scriptsize$\pm$1.08} & \cellcolor{gray!12}9.72{\scriptsize$\pm$0.76} & \cellcolor{gray!12}23.47{\scriptsize$\pm$1.21} & \cellcolor{gray!12}5.37{\scriptsize$\pm$0.41} & \cellcolor{gray!12}10.12{\scriptsize$\pm$0.66} & \cellcolor{gray!12}7.19{\scriptsize$\pm$0.53} & \cellcolor{gray!12}15.50{\scriptsize$\pm$0.64} & \cellcolor{gray!12}3.33{\scriptsize$\pm$0.80} & \cellcolor{gray!12}3.12{\scriptsize$\pm$0.30} & \cellcolor{gray!12}3.31{\scriptsize$\pm$0.53} & \cellcolor{gray!12}9.28 & \cellcolor{gray!12}19.30 & \cellcolor{gray!12}14.00 \\
 \multirow{2}{*}{PHANES} & AP & 11.20{\scriptsize$\pm$0.19} & 21.02{\scriptsize$\pm$3.89} & 10.40{\scriptsize$\pm$0.32} & 20.75{\scriptsize$\pm$3.62} & 9.97{\scriptsize$\pm$0.25} & 15.57{\scriptsize$\pm$1.80} & 16.08{\scriptsize$\pm$0.30} & 24.90{\scriptsize$\pm$5.45} & 8.79{\scriptsize$\pm$0.20} & 18.17{\scriptsize$\pm$2.70} & 5.42{\scriptsize$\pm$0.13} & 10.51{\scriptsize$\pm$2.11} & 6.15{\scriptsize$\pm$0.15} & 15.47{\scriptsize$\pm$3.25} & 2.04{\scriptsize$\pm$0.09} & 1.87{\scriptsize$\pm$0.11} & 4.04{\scriptsize$\pm$0.54} & 8.23 & 16.03 & 11.90 \\
 & \cellcolor{gray!12}DSC & \cellcolor{gray!12}24.29{\scriptsize$\pm$0.42} & \cellcolor{gray!12}37.27{\scriptsize$\pm$0.59} & \cellcolor{gray!12}24.28{\scriptsize$\pm$0.71} & \cellcolor{gray!12}37.66{\scriptsize$\pm$0.71} & \cellcolor{gray!12}21.82{\scriptsize$\pm$0.61} & \cellcolor{gray!12}30.95{\scriptsize$\pm$0.78} & \cellcolor{gray!12}29.15{\scriptsize$\pm$0.26} & \cellcolor{gray!12}39.85{\scriptsize$\pm$1.15} & \cellcolor{gray!12}26.33{\scriptsize$\pm$0.67} & \cellcolor{gray!12}37.25{\scriptsize$\pm$0.95} & \cellcolor{gray!12}11.25{\scriptsize$\pm$0.12} & \cellcolor{gray!12}17.21{\scriptsize$\pm$0.38} & \cellcolor{gray!12}12.59{\scriptsize$\pm$0.18} & \cellcolor{gray!12}22.03{\scriptsize$\pm$0.17} & \cellcolor{gray!12}5.42{\scriptsize$\pm$0.27} & \cellcolor{gray!12}4.80{\scriptsize$\pm$0.35} & \cellcolor{gray!12}10.57{\scriptsize$\pm$0.64} & \cellcolor{gray!12}18.41 & \cellcolor{gray!12}28.38 & \cellcolor{gray!12}23.10 \\
\multirow{2}{*}{VAE} & AP & 12.59{\scriptsize$\pm$0.14} & 37.30{\scriptsize$\pm$0.14} & 13.41{\scriptsize$\pm$0.05} & 39.04{\scriptsize$\pm$0.24} & 9.49{\scriptsize$\pm$0.16} & 25.97{\scriptsize$\pm$0.44} & 15.17{\scriptsize$\pm$0.44} & 45.56{\scriptsize$\pm$0.66} & 10.10{\scriptsize$\pm$0.19} & 29.31{\scriptsize$\pm$0.05} & 5.75{\scriptsize$\pm$0.17} & 19.86{\scriptsize$\pm$0.25} & 7.52{\scriptsize$\pm$0.07} & 30.38{\scriptsize$\pm$0.43} & 2.76{\scriptsize$\pm$0.11} & 2.84{\scriptsize$\pm$0.17} & 2.64{\scriptsize$\pm$0.25} & 8.83 & 28.78 & 18.22 \\
 & \cellcolor{gray!12}DSC & \cellcolor{gray!12}18.69{\scriptsize$\pm$0.13} & \cellcolor{gray!12}38.26{\scriptsize$\pm$0.20} & \cellcolor{gray!12}19.99{\scriptsize$\pm$0.12} & \cellcolor{gray!12}39.18{\scriptsize$\pm$0.22} & \cellcolor{gray!12}15.01{\scriptsize$\pm$0.19} & \cellcolor{gray!12}31.74{\scriptsize$\pm$0.29} & \cellcolor{gray!12}21.45{\scriptsize$\pm$0.54} & \cellcolor{gray!12}47.66{\scriptsize$\pm$0.48} & \cellcolor{gray!12}18.65{\scriptsize$\pm$0.43} & \cellcolor{gray!12}37.24{\scriptsize$\pm$0.21} & \cellcolor{gray!12}7.84{\scriptsize$\pm$0.28} & \cellcolor{gray!12}16.08{\scriptsize$\pm$0.13} & \cellcolor{gray!12}9.89{\scriptsize$\pm$0.13} & \cellcolor{gray!12}22.75{\scriptsize$\pm$0.25} & \cellcolor{gray!12}5.22{\scriptsize$\pm$0.11} & \cellcolor{gray!12}5.76{\scriptsize$\pm$0.32} & \cellcolor{gray!12}4.26{\scriptsize$\pm$0.42} & \cellcolor{gray!12}13.44 & \cellcolor{gray!12}29.83 & \cellcolor{gray!12}21.16 \\
\multirow{2}{*}{RIAD} & AP & 21.56{\scriptsize$\pm$2.03} & 23.28{\scriptsize$\pm$5.07} & 28.38{\scriptsize$\pm$2.96} & 26.72{\scriptsize$\pm$5.88} & 18.66{\scriptsize$\pm$2.02} & 18.86{\scriptsize$\pm$2.87} & 20.98{\scriptsize$\pm$1.85} & 33.11{\scriptsize$\pm$5.30} & 13.60{\scriptsize$\pm$0.60} & 22.41{\scriptsize$\pm$4.75} & 14.05{\scriptsize$\pm$2.20} & 17.47{\scriptsize$\pm$1.13} & 13.52{\scriptsize$\pm$2.20} & 21.06{\scriptsize$\pm$4.18} & \green{6.91}{\scriptsize$\pm$0.58} & 4.04{\scriptsize$\pm$0.55} & 12.92{\scriptsize$\pm$0.63} & 16.73 & 20.87 & 18.68 \\
 & \cellcolor{gray!12}DSC & \cellcolor{gray!12}26.21{\scriptsize$\pm$1.94} & \cellcolor{gray!12}28.29{\scriptsize$\pm$3.50} & \cellcolor{gray!12}29.96{\scriptsize$\pm$2.27} & \cellcolor{gray!12}29.65{\scriptsize$\pm$3.54} & \cellcolor{gray!12}23.04{\scriptsize$\pm$1.93} & \cellcolor{gray!12}23.21{\scriptsize$\pm$1.71} & \cellcolor{gray!12}25.50{\scriptsize$\pm$1.35} & \cellcolor{gray!12}36.01{\scriptsize$\pm$4.14} & \cellcolor{gray!12}23.54{\scriptsize$\pm$0.79} & \cellcolor{gray!12}27.37{\scriptsize$\pm$3.71} & \cellcolor{gray!12}12.29{\scriptsize$\pm$1.28} & \cellcolor{gray!12}15.15{\scriptsize$\pm$0.40} & \cellcolor{gray!12}13.20{\scriptsize$\pm$1.27} & \cellcolor{gray!12}17.91{\scriptsize$\pm$1.41} & \cellcolor{gray!12}\green{7.14}{\scriptsize$\pm$0.65} & \cellcolor{gray!12}3.52{\scriptsize$\pm$0.18} & \cellcolor{gray!12}10.37{\scriptsize$\pm$0.41} & \cellcolor{gray!12}19.03 & \cellcolor{gray!12}22.64 & \cellcolor{gray!12}20.73 \\
\multirow{2}{*}{ProxyAno} & AP & 11.51{\scriptsize$\pm$0.70} & 39.30{\scriptsize$\pm$3.42} & 11.15{\scriptsize$\pm$0.84} & 40.57{\scriptsize$\pm$3.55} & 8.61{\scriptsize$\pm$0.40} & 25.50{\scriptsize$\pm$2.30} & 15.02{\scriptsize$\pm$0.66} & 54.21{\scriptsize$\pm$3.42} & 8.93{\scriptsize$\pm$0.55} & 34.67{\scriptsize$\pm$2.78} & 5.01{\scriptsize$\pm$0.30} & 23.12{\scriptsize$\pm$2.77} & 6.51{\scriptsize$\pm$0.41} & 28.86{\scriptsize$\pm$2.96} & 3.32{\scriptsize$\pm$0.16} & 3.56{\scriptsize$\pm$0.22} & 6.94{\scriptsize$\pm$0.62} & 8.55 & 31.22 & 19.22 \\
 & \cellcolor{gray!12}DSC & \cellcolor{gray!12}15.44{\scriptsize$\pm$0.34} & \cellcolor{gray!12}34.57{\scriptsize$\pm$1.56} & \cellcolor{gray!12}15.04{\scriptsize$\pm$0.60} & \cellcolor{gray!12}35.13{\scriptsize$\pm$1.72} & \cellcolor{gray!12}12.97{\scriptsize$\pm$0.17} & \cellcolor{gray!12}26.29{\scriptsize$\pm$1.10} & \cellcolor{gray!12}20.84{\scriptsize$\pm$0.24} & \cellcolor{gray!12}48.58{\scriptsize$\pm$1.93} & \cellcolor{gray!12}16.05{\scriptsize$\pm$0.94} & \cellcolor{gray!12}36.22{\scriptsize$\pm$2.16} & \cellcolor{gray!12}6.53{\scriptsize$\pm$0.11} & \cellcolor{gray!12}14.23{\scriptsize$\pm$0.67} & \cellcolor{gray!12}8.02{\scriptsize$\pm$0.12} & \cellcolor{gray!12}19.89{\scriptsize$\pm$0.94} & \cellcolor{gray!12}4.77{\scriptsize$\pm$0.17} & \cellcolor{gray!12}5.60{\scriptsize$\pm$0.31} & \cellcolor{gray!12}9.40{\scriptsize$\pm$0.74} & \cellcolor{gray!12}12.12 & \cellcolor{gray!12}27.56 & \cellcolor{gray!12}19.39 \\
 \multirow{2}{*}{AnoDDPM} & AP & 17.77{\scriptsize$\pm$0.50} & 58.61{\scriptsize$\pm$1.49} & 17.99{\scriptsize$\pm$0.49} & 59.69{\scriptsize$\pm$1.51} & 15.93{\scriptsize$\pm$0.40} & 53.32{\scriptsize$\pm$1.35} & 18.10{\scriptsize$\pm$0.45} & 50.94{\scriptsize$\pm$1.29} & 14.58{\scriptsize$\pm$0.36} & 40.66{\scriptsize$\pm$1.03} & 8.07{\scriptsize$\pm$0.20} & 35.76{\scriptsize$\pm$0.90} & 6.67{\scriptsize$\pm$0.17} & 43.38{\scriptsize$\pm$1.10} & 2.61{\scriptsize$\pm$0.07} & 4.58{\scriptsize$\pm$0.12} & \bblue{19.12}{\scriptsize$\pm$0.48} & 13.43 & 43.37 & 27.52 \\
 & \cellcolor{gray!12}DSC & \cellcolor{gray!12}20.00{\scriptsize$\pm$0.43} & \cellcolor{gray!12}48.30{\scriptsize$\pm$1.09} & \cellcolor{gray!12}19.50{\scriptsize$\pm$0.43} & \cellcolor{gray!12}47.91{\scriptsize$\pm$1.08} & \cellcolor{gray!12}20.67{\scriptsize$\pm$0.46} & \cellcolor{gray!12}44.22{\scriptsize$\pm$0.99} & \cellcolor{gray!12}22.05{\scriptsize$\pm$0.52} & \cellcolor{gray!12}48.98{\scriptsize$\pm$1.10} & \cellcolor{gray!12}22.94{\scriptsize$\pm$0.51} & \cellcolor{gray!12}44.31{\scriptsize$\pm$0.99} & \cellcolor{gray!12}8.04{\scriptsize$\pm$0.18} & \cellcolor{gray!12}19.17{\scriptsize$\pm$0.43} & \cellcolor{gray!12}8.64{\scriptsize$\pm$0.20} & \cellcolor{gray!12}25.26{\scriptsize$\pm$0.56} & \cellcolor{gray!12}4.30{\scriptsize$\pm$0.10} & \cellcolor{gray!12}\green{6.75}{\scriptsize$\pm$0.15} & \cellcolor{gray!12}11.65{\scriptsize$\pm$0.26} & \cellcolor{gray!12}15.31 & \cellcolor{gray!12}35.61 & \cellcolor{gray!12}24.86 \\
\multirow{2}{*}{DAE} & AP & 13.31{\scriptsize$\pm$0.37} & \bblue{64.41}{\scriptsize$\pm$3.57} & 9.86{\scriptsize$\pm$0.67} & \bblue{69.72}{\scriptsize$\pm$1.98} & 15.68{\scriptsize$\pm$1.33} & 63.56{\scriptsize$\pm$4.98} & 24.02{\scriptsize$\pm$1.01} & \bblue{65.39}{\scriptsize$\pm$1.53} & 7.44{\scriptsize$\pm$0.45} & 44.38{\scriptsize$\pm$1.74} & 9.29{\scriptsize$\pm$0.19} & \green{48.85}{\scriptsize$\pm$3.76} & 9.39{\scriptsize$\pm$0.60} & 49.77{\scriptsize$\pm$3.45} & 1.69{\scriptsize$\pm$0.02} & 3.56{\scriptsize$\pm$0.74} & 6.05{\scriptsize$\pm$1.35} & 10.75 & \green{51.21} & 29.79 \\
 & \cellcolor{gray!12}DSC & \cellcolor{gray!12}16.84{\scriptsize$\pm$0.47} & \cellcolor{gray!12}\bblue{54.93}{\scriptsize$\pm$2.32} & \cellcolor{gray!12}12.65{\scriptsize$\pm$0.20} & \cellcolor{gray!12}\bblue{60.05}{\scriptsize$\pm$0.34} & \cellcolor{gray!12}19.65{\scriptsize$\pm$1.70} & \cellcolor{gray!12}53.56{\scriptsize$\pm$3.58} & \cellcolor{gray!12}26.94{\scriptsize$\pm$0.41} & \cellcolor{gray!12}\green{57.65}{\scriptsize$\pm$1.47} & \cellcolor{gray!12}13.93{\scriptsize$\pm$0.52} & \cellcolor{gray!12}38.03{\scriptsize$\pm$1.77} & \cellcolor{gray!12}8.35{\scriptsize$\pm$0.10} & \cellcolor{gray!12}\bblue{24.09}{\scriptsize$\pm$1.58} & \cellcolor{gray!12}10.11{\scriptsize$\pm$0.18} & \cellcolor{gray!12}28.38{\scriptsize$\pm$1.59} & \cellcolor{gray!12}2.91{\scriptsize$\pm$0.09} & \cellcolor{gray!12}4.06{\scriptsize$\pm$0.48} & \cellcolor{gray!12}5.83{\scriptsize$\pm$1.17} & \cellcolor{gray!12}13.02 & \cellcolor{gray!12}\green{40.09} & \cellcolor{gray!12}25.76 \\
\multirow{2}{*}{AutoDDPM} & AP & 30.03{\scriptsize$\pm$2.03} & 55.83{\scriptsize$\pm$1.07} & 34.43{\scriptsize$\pm$2.25} & 61.50{\scriptsize$\pm$0.58} & 26.74{\scriptsize$\pm$2.41} & 49.51{\scriptsize$\pm$1.22} & 26.80{\scriptsize$\pm$2.52} & 52.30{\scriptsize$\pm$0.90} & \green{25.70}{\scriptsize$\pm$0.93} & \bblue{55.03}{\scriptsize$\pm$1.61} & 19.52{\scriptsize$\pm$2.18} & 41.26{\scriptsize$\pm$0.84} & 21.54{\scriptsize$\pm$1.91} & 47.71{\scriptsize$\pm$0.97} & \bblue{8.84}{\scriptsize$\pm$1.86} & \green{4.92}{\scriptsize$\pm$0.75} & 16.03{\scriptsize$\pm$0.26} & 23.29 & 46.01 & 33.98 \\
 & \cellcolor{gray!12}DSC & \cellcolor{gray!12}30.26{\scriptsize$\pm$1.61} & \cellcolor{gray!12}47.76{\scriptsize$\pm$0.54} & \cellcolor{gray!12}32.47{\scriptsize$\pm$1.58} & \cellcolor{gray!12}50.67{\scriptsize$\pm$0.45} & \cellcolor{gray!12}28.68{\scriptsize$\pm$2.02} & \cellcolor{gray!12}41.85{\scriptsize$\pm$1.08} & \cellcolor{gray!12}29.71{\scriptsize$\pm$1.46} & \cellcolor{gray!12}49.62{\scriptsize$\pm$1.27} & \cellcolor{gray!12}\bblue{35.52}{\scriptsize$\pm$0.72} & \cellcolor{gray!12}\bblue{51.75}{\scriptsize$\pm$1.68} & \cellcolor{gray!12}\green{14.89}{\scriptsize$\pm$1.17} & \cellcolor{gray!12}22.25{\scriptsize$\pm$0.68} & \cellcolor{gray!12}15.64{\scriptsize$\pm$1.08} & \cellcolor{gray!12}\green{28.48}{\scriptsize$\pm$0.49} & \cellcolor{gray!12}\bblue{9.97}{\scriptsize$\pm$1.41} & \cellcolor{gray!12}6.10{\scriptsize$\pm$0.83} & \cellcolor{gray!12}\bblue{15.50}{\scriptsize$\pm$0.19} & \cellcolor{gray!12}\bblue{23.63} & \cellcolor{gray!12}37.31 & \cellcolor{gray!12}\green{30.07} \\
\multirow{2}{*}{Ours} & AP & \red{54.30}{\scriptsize$\pm$1.85} & \red{83.45}{\scriptsize$\pm$1.47} & \red{64.60}{\scriptsize$\pm$2.30} & \red{87.34}{\scriptsize$\pm$1.40} & \red{51.11}{\scriptsize$\pm$2.08} & \red{85.79}{\scriptsize$\pm$0.36} & \red{39.89}{\scriptsize$\pm$3.23} & \red{71.24}{\scriptsize$\pm$1.59} & \red{39.13}{\scriptsize$\pm$2.30} & \red{66.28}{\scriptsize$\pm$1.86} & \red{31.35}{\scriptsize$\pm$1.92} & \red{69.29}{\scriptsize$\pm$2.06} & \red{35.38}{\scriptsize$\pm$3.44} & \red{74.55}{\scriptsize$\pm$2.34} & \red{18.90}{\scriptsize$\pm$1.46} & \red{18.18}{\scriptsize$\pm$2.21} & \red{43.25}{\scriptsize$\pm$1.05} & \red{41.99} & \red{69.51} & \red{54.94} \\
 & \cellcolor{gray!12}DSC & \cellcolor{gray!12}\red{48.79}{\scriptsize$\pm$1.23} & \cellcolor{gray!12}\red{72.52}{\scriptsize$\pm$1.03} & \cellcolor{gray!12}\red{53.73}{\scriptsize$\pm$1.94} & \cellcolor{gray!12}\red{75.40}{\scriptsize$\pm$0.79} & \cellcolor{gray!12}\red{48.01}{\scriptsize$\pm$1.56} & \cellcolor{gray!12}\red{73.03}{\scriptsize$\pm$0.17} & \cellcolor{gray!12}\red{44.38}{\scriptsize$\pm$1.30} & \cellcolor{gray!12}\red{65.31}{\scriptsize$\pm$1.82} & \cellcolor{gray!12}\red{50.66}{\scriptsize$\pm$0.99} & \cellcolor{gray!12}\red{62.76}{\scriptsize$\pm$0.92} & \cellcolor{gray!12}\red{19.21}{\scriptsize$\pm$0.56} & \cellcolor{gray!12}\red{34.38}{\scriptsize$\pm$0.50} & \cellcolor{gray!12}\red{24.98}{\scriptsize$\pm$1.17} & \cellcolor{gray!12}\red{45.68}{\scriptsize$\pm$1.46} & \cellcolor{gray!12}\red{12.74}{\scriptsize$\pm$0.74} & \cellcolor{gray!12}\red{13.28}{\scriptsize$\pm$0.94} & \cellcolor{gray!12}\red{23.08}{\scriptsize$\pm$1.45} & \cellcolor{gray!12}\red{36.18} & \cellcolor{gray!12}\red{55.30} & \cellcolor{gray!12}\red{45.17} \\
 \midrule
\bottomrule[1.pt]
\end{tabular}
}
\end{sidewaystable*}
        	\begin{figure*}[]
		\centerline{\includegraphics[width=7.3in]{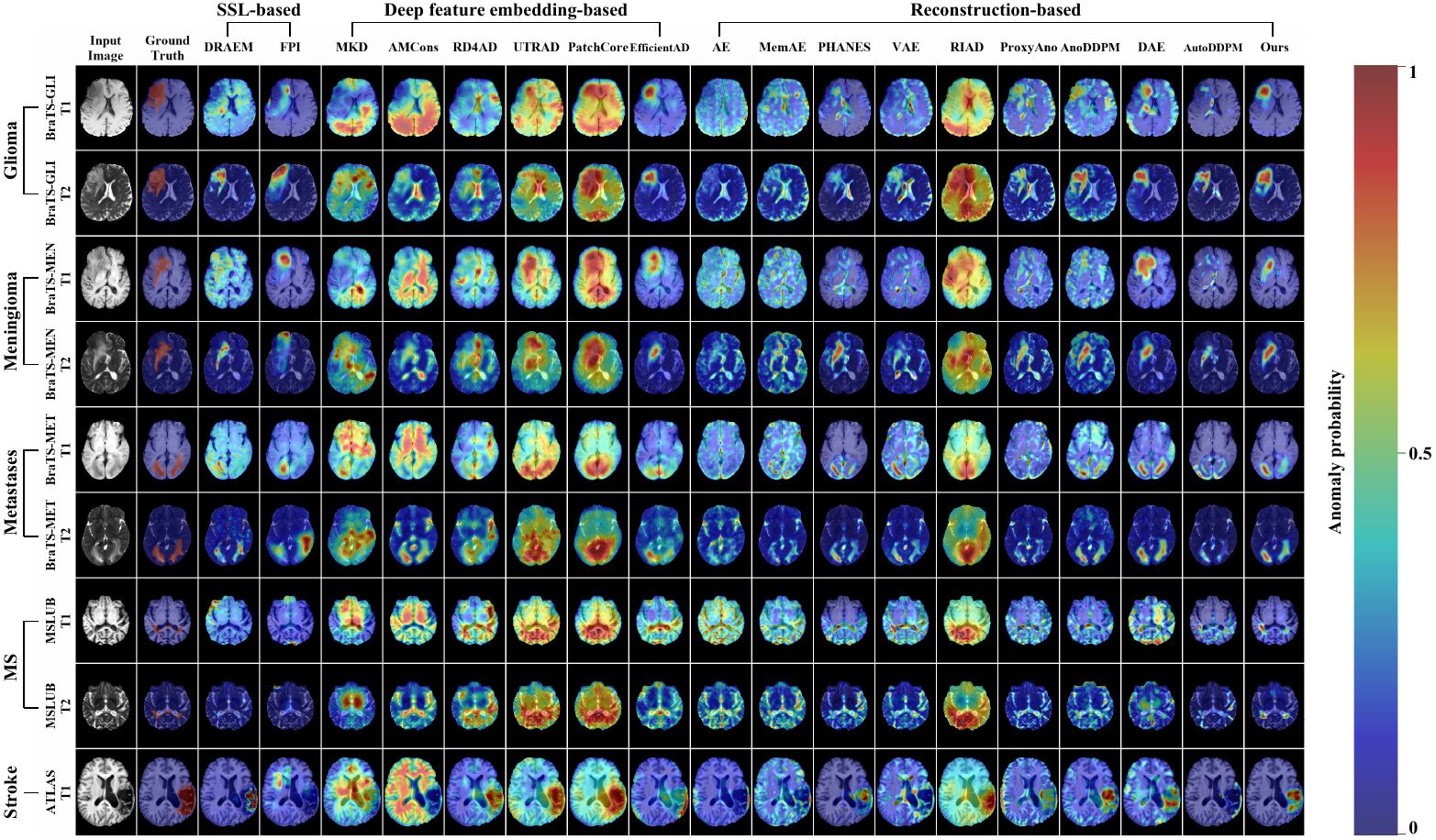}}
		\caption{\textcolor{black}{Comparison of anomaly maps generated by our method (three orthogonal views in \ref{Anomaly map visualization}) and 17 SOTA methods. All anomaly maps are scaled to [0, 1] using Max-Min normalization for clearer display. Our method shows superior detection of various brain anomalies in T1 and T2 scans.}}
		\label{fig2}
	\end{figure*}
	\subsection{Performance, Comparison with SOTA Methods and Generalizability}
	\subsubsection{Performance and Comparison with SOTA Methods}
	Table \ref{tab3} reports the AP and DSC metrics of our method and 17 SOTA methods over T1 and T2 images from nine different validation datasets, respectively. Table \ref{tab3} demonstrates that the proposed method shows considerable superiority in anomaly detection on all datasets compared with 17 SOTA methods. Specifically, based on average AP and DSC metrics of different methods over T1 and T2 images of nine different datasets, we obtained the following results: (1) For the T1 images (4,443 images), our method attained an AP/DSC of 41.99\%/36.18\%, yielding a minimum absolute improvement of 16.83\%/12.55\% (25.16\% vs. 41.99\%, 23.63\% vs. 36.18\%) compared with 17 SOTA methods. (2) For the T2 images (3,788 images), it achieved 69.51\%/55.30\%, representing an absolute improvement of at least 17.78\%/14.39\% (51.73\% vs. 69.51\%, 40.91\% vs. 55.30\%). (3) For the T1 and T2 images (8,231 images), the method yielded 54.94\%/45.17\%, with a minimum absolute improvement of 18.32\%/13.64\% (36.62\% vs. 54.94\%, 31.53\% vs. 45.17\%).
    
    The pathologies in Table \ref{tab3} vary in detection performance, which may be related to lesion size. For example, experimental results indicate that while most comparative methods performed reasonably well on glioma, which typically presents as a large lesion, our method achieved superior detection results. In contrast, for the challenging task of detecting small lesions (e.g., multiple sclerosis), where all comparative methods' performance was severely limited, our method achieved the highest detection performance.
                
    The comparative methods in Table \ref{tab3} fall into three categories. (1) Self-supervised learning-based methods, such as DRAEM \citep{ref8} and FPI \citep{ref9}, rely on synthesized anomalies for anomaly detection. Experimental results show that they have limited generalizability in detecting diverse pathology types, such as stroke (ATLAS \citep{ref54}). In contrast, our method does not rely on anomaly simulation, thereby avoiding this limitation and demonstrating stronger generalizability across different pathology types. (2) Deep feature embedding-based methods, such as PatchCore \citep{ref10} and EfficientAD \citep{ref11}, detect anomalies in a downsampled feature space, which risks the loss of fine-grained information. Experimental results show that they have limited detection performance for small lesions (e.g., multiple sclerosis, MSLUB \citep{ref53}). In contrast, our method addresses this granularity limitation by performing detection in the full-resolution image space, and effectively reconstructs personalized normal details to minimize the loss of fine-grained information. (3) Other reconstruction-based methods, such as AnoDDPM  \citep{wyatt2022anoddpm}, DAE \citep{ref13}, and AutoDDPM \citep{bercea2023mask}, exhibit lower detection performance than our method on all datasets because they often face challenges in effectively suppressing anomalies (e.g., DAE) or accurately reconstructing personalized normal details (e.g., AnoDDPM and AutoDDPM). As demonstrated in Table \ref{tab3}, our method achieved superior detection results compared to other reconstruction-based methods by addressing these challenges: it simultaneously suppresses anomalies and effectively reconstructs normal details.
    \begin{figure*}[t]
		\centerline{\includegraphics[width=7.0in]{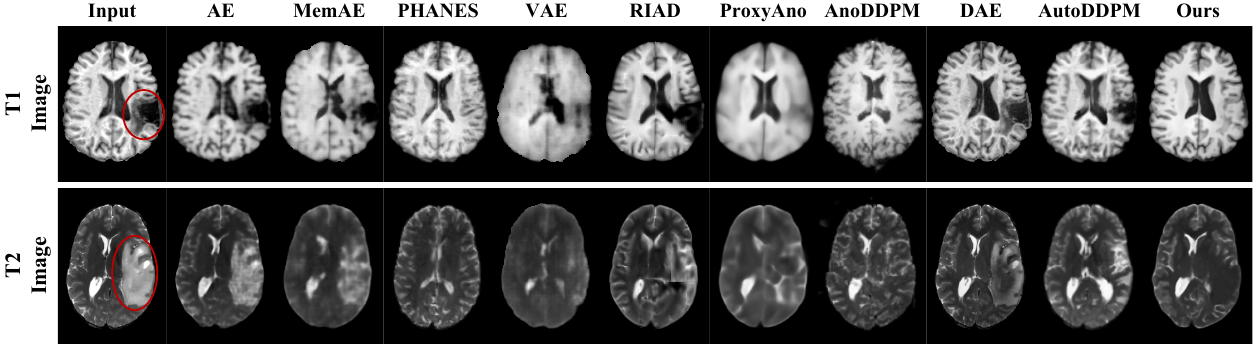}}
		\caption{\textcolor{black}{Comparison of pseudo-healthy images reconstructed by our method and other reconstruction-based methods. Our method repairs anomalies more effectively while preserving normal brain details with high fidelity.}}
		\label{fig3}
	\end{figure*}
	\subsubsection{Generalizability for Imaging Sequences, Multi-Center Datasets and Lesion Types}
	Table \ref{tab3} shows that (1) the single detection model trained by the proposed method can detect anomalies from T1 and T2 images, demonstrating its good generalization capability to different imaging sequences. In contrast, as presented in their method design and implementation, other comparison methods for brain anomaly detection, such as AutoDDPM \citep{bercea2023mask}, AnoDDPM \citep{wyatt2022anoddpm}, PHANES \citep{bercea2023reversing}, FPI \citep{ref9}, and AMCons \citep{ref28} typically train models and perform detection on a single MRI sequence such as T1 or FLAIR. (2) The proposed method exhibits stable detection performance for the same lesion from multi-center datasets. For example, adult glioma from three datasets (BraTS-GLI, UPenn-GBM, UCSF-PDGM) showed similar detection performance, with AP metrics of 83.45\%, 87.34\%, and 85.79\%, and DSC metrics of 72.52\%, 75.40\%, and 73.03\% for T2 images, respectively. In contrast, many comparison methods, such as DRAEM, VAE, and ProxyAno, showed significant variability in detection performance across these three datasets. (3) Compared to other 17 SOTA methods, the proposed method demonstrates considerable anomaly detection superiority over all types of lesions.

	\subsection{Visualization of Anomaly Maps and Reconstructed Pseudo-Healthy Images}
	\subsubsection{Anomaly Maps}
	Fig. \ref{fig2} shows anomaly maps generated by our method and 17 SOTA methods from five pairs of T1/T2 images with five different types of lesions. In the anomaly maps, different colors correspond to different anomaly probabilities, as indicated by the color bar in Fig. \ref{fig2}. Fig. \ref{fig2} indicates that the proposed method provides a more accurate localization of anomalies by assigning high probabilities to diverse abnormal regions and maintaining low probabilities in normal regions. However, SSL-based methods only generated high probabilities in a small part of the abnormal regions, and most deep feature embedding-based methods generated high probabilities in some regions that were much larger than the abnormal regions. Additionally, our method achieved the highest detection performance on both T1 and T2 images. In contrast, some methods (e.g., DRAEM, AMCons, and DAE) exhibited notable differences in both detection performance and anomaly probability distributions on T1 and T2 images.
	\subsubsection{Reconstructed Pseudo-Healthy Images}
	Fig. \ref{fig3} shows pseudo-healthy images reconstructed by different methods from the same abnormal image. Fig. \ref{fig3} demonstrates that the proposed method suppresses nearly the entire lesion region while preserving the personalized normal details of the brain. However, AE, MemAE, DAE, and RIAD are limited in suppressing anomaly regions, while VAE, ProxyAno, MemAE, RIAD, PHANES, AnoDDPM and AutoDDPM tend to generate images exhibiting blurring or a loss of personalized details. Consequently, the proposed method improves anomaly detection, whereas other methods either lead to false negatives or false positives.
            \begin{table}[tbp]\scriptsize
		\caption{\textcolor{black}{Reconstruction quality comparison of our method and other reconstruction-based methods for T1 and T2 sequences. The units for $l_1$-$\text{RE}_{N}$ and $l_1$-$\text{RE}_{A}$ are e-2. The arrows $\uparrow$ and $\downarrow$ indicate that higher and lower values are favorable, respectively. The \red{best}, \bblue{second-best} and \green{third-best} results are highlighted in red, blue and green, respectively.}}
		\label{tab4}
		\centering
        \renewcommand{\arraystretch}{1.08}
		\setlength{\tabcolsep}{1.25mm}{
        \resizebox{\columnwidth}{!}{
			\begin{tabular}{lcccccc}
				\toprule[1.pt]
         \multirow{2}{*}{Method} & \multicolumn{3}{c}{T1 (N=4,443)} & \multicolumn{3}{c}{T2 (N=3,788)} \\ 
         & $l_1$-$\text{RE}_{N}$$\downarrow$ & $l_1$-$\text{RE}_{A}$$\uparrow$ & $l_1$-ratio$\uparrow$ & $l_1$-$\text{RE}_{N}$$\downarrow$ & $l_1$-$\text{RE}_{A}$$\uparrow$ & $l_1$-ratio$\uparrow$ \\ 
        \hline
        AE & 5.59 & 5.60 & 1.00 & 3.91 & 4.20 & 1.07 \\
        MemAE & 9.33 & 9.61 & 1.03 & 7.67 & 10.10 & 1.32 \\
        PHANES & 11.67 & 16.05 & 1.38 & 8.61 & 19.69 & 2.29 \\
        VAE & 13.20 & 15.71 & 1.19 & 9.62 & 20.03 & 2.08 \\
        RIAD & 9.12 & 9.27 & 1.02 & 7.21 & 9.15 & 1.27 \\
        ProxyAno & 10.40 & 13.06 & 1.26 & 8.48 & 15.60 & 1.84 \\
        AnoDDPM & 14.29 & 19.72 & 1.38 & 9.49 & 21.56 & 2.27 \\
        DAE & 6.72 & 9.45 & \green{1.41} & 3.60 & 10.32 & \bblue{2.87} \\
        AutoDDPM & 9.19 & 13.88 & \bblue{1.51} & 7.35 & 18.30 & \green{2.49} \\
        Ours & 8.71 & 16.63 & \red{1.91} & 6.95 & 22.36 & \red{3.22} \\
        \toprule[1.pt]
			\end{tabular}}}
	\end{table}

    To quantitatively evaluate reconstruction quality, we calculated the $l_1$ errors between all abnormal images and reconstructed images, and separately calculated the average reconstruction error for normal ($l_1$-$\text{RE}_{N}$) and abnormal regions ($l_1$-$\text{RE}_{A}$) based on voxel-level ground truth annotations. We also analyzed the $l_1$-ratio \citep{behrendt2025guided}, defined as $l_1$-$\text{RE}_{A}$ divided by $l_1$-$\text{RE}_{N}$, to evaluate the model's ability to reconstruct normal regions while simultaneously suppressing abnormal regions. 
    
    As shown in Table \ref{tab4}, our method achieved the highest $l_1$-ratio on both T1 and T2 sequences, indicating that the proposed method achieves an optimal balance between reconstructing normal regions and suppressing anomalies. In contrast, although AE and DAE achieve higher reconstruction accuracy for normal regions, their limited capability for anomaly suppression (as visually illustrated in Fig. \ref{fig3}) results in lower sensitivity to anomalies, as evidenced by the quantitative detection results in Table \ref{tab3}. 
    \begin{figure}[t!]
		\centerline{\includegraphics[width=3.5in]{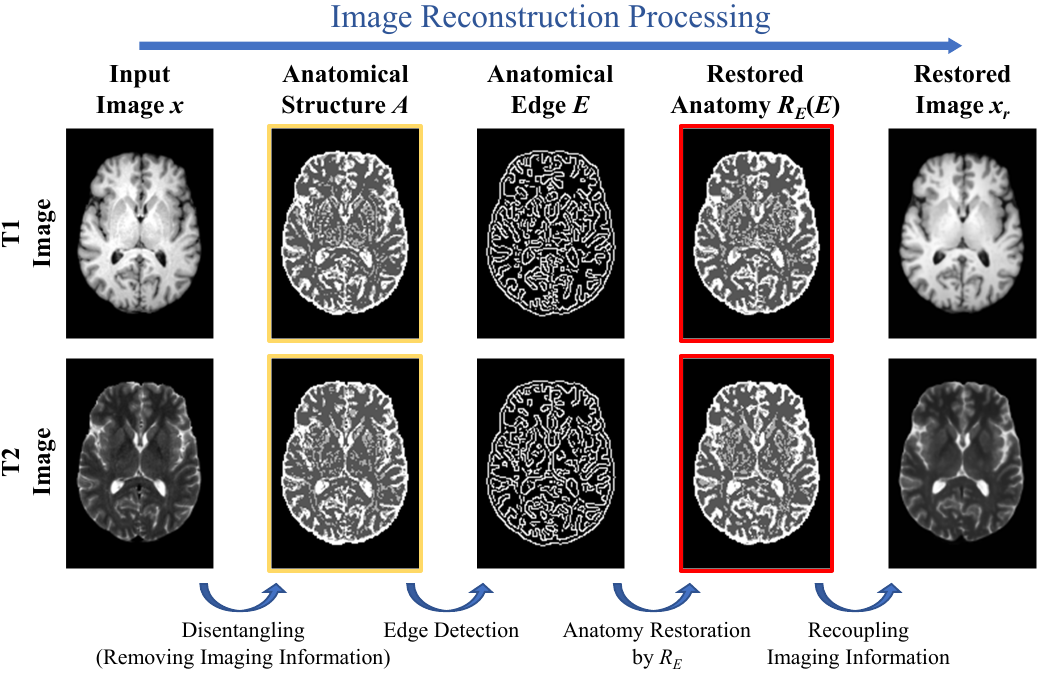}}
		\caption{\textcolor{black}{The pseudo-healthy image reconstruction pipeline and a visual analysis of the restored anatomical structure.}}
		\label{fig4}
	\end{figure}
    \subsection{Analysis for Anatomy Restorer}\label{Analysis for Anatomy Restorer}
    Fig. \ref{fig4} illustrates the reconstruction pipeline of the proposed method for pseudo-healthy images. As shown in Fig. \ref{fig4}, the proposed method firstly removes all imaging information (e.g., sequence type information and continuous grayscale information) from the original MRI image $x$, obtaining an imaging-independent, cross-sequence (e.g., T1 and T2) aligned anatomical image $A$ (yellow boxes), which is represented as a multi-class semantic map. Anatomical edges $E$, which contain only imaging-independent spatial structure information, are then extracted from this anatomical image.
    
    These anatomical edges $E$ are provided as the sole input to the anatomy restorer $R_E$, which fundamentally restricts the influence of imaging information on $R_E$.  Furthermore, we constrain the output $R_E(E)$ via Eq. \ref{eq_accl}, aligning it with the disentangled anatomical code $R_A(A)$ that is also free of imaging information. Therefore, by excluding imaging information from both its input (anatomical spatial structure information) and its output (alignment with the disentangled anatomical code), the $R_E$ is able to focus on imaging-independent anatomical learning and restoration without being biased by imaging information.
    
    Moreover, anatomical edges $E$ contain the critical high-frequency structural details of a brain anatomical image, which are sufficient for effective anatomical restoration. To verify this, we added a decoder to $R_E$ to visualize its edge-based restoration output $R_E(E)$. As shown in Fig. \ref{fig4}, the restored anatomical image (red boxes) from $R_E$ is similar to the original anatomical image (yellow boxes), demonstrating that the $R_E$ can effectively use the personalized structural information within the anatomical edges to perform anatomical restoration.
    
    \subsection{Ablations Studies}
    \begin{table}[t!]\scriptsize
		\caption{\textcolor{black}{Comparison of Different Anomaly Scores. The best two results are marked in bold and underlined.}}
		\label{tab5}
		\centering
        \renewcommand{\arraystretch}{1.1}
		\setlength{\tabcolsep}{2.65mm}{
        \resizebox{\columnwidth}{!}{
			\begin{tabular}{ccccccc}
				\toprule[1.pt]
				\multirow{2}{*}{Anomaly score} & \multicolumn{2}{c}{T1 (N=4,443)} & \multicolumn{2}{c}{T2 (N=3,788)} & \multicolumn{2}{c}{Average} \\
				& AP$\uparrow$ & DSC$\uparrow$ & AP$\uparrow$ & DSC$\uparrow$ & AP$\uparrow$ & DSC$\uparrow$              \\ \hline
				SSIM                             & \underline{36.40}           & \underline{33.80}          & 43.32           & 43.02          & 39.66             & 38.14            \\
				L1                               & 31.61           & 28.16          & \underline{67.97}           & \underline{54.10}          & \underline{48.72}             & \underline{40.36}            \\
				\textbf{Hybrid (Ours)}                    & \textbf{43.61}           & \textbf{36.89}          & \textbf{70.78}           & \textbf{56.18}          & \textbf{56.40}             & \textbf{45.96}            \\ \toprule[1.pt]
			\end{tabular}
		}
        }
	\end{table}
        
	\subsubsection{Ablation Study of Anomaly Scores}
	We compared three different anomaly scores: The L1 anomaly score, the structural similarity (SSIM) anomaly score, and our proposed hybrid anomaly score across all abnormal datasets, as shown in Table \ref{tab5}. The proposed hybrid anomaly score outperformed the L1 and SSIM anomaly scores in terms of detection performance. Using the hybrid anomaly score, anomaly detection performance  was improved absolutely by 7.21\%/3.09\% (36.40\% vs. 43.61\%, 33.80\% vs. 36.89\%) in the AP/DSC metrics for T1 images and by 2.81\%/2.08\% (67.97\% vs. 70.78\%, 54.10\% vs. 56.18\%) for T2 images, demonstrating its effectiveness, especially for the T1 images.
	
	\subsubsection{Ablation Study of Loss Functions}
	We conducted ablation experiments on loss functions to verify their contributions, including Eqs. \ref{eq_a_con}-\ref{eq_m_sim} and Eq. \ref{eq_accl}. The validation was performed on the BraTS-GLI dataset, and the results are presented in Table \ref{tab6}. As shown in Table \ref{tab6}, the consistency losses $L_{\text{con}}^A$ (Eq. \ref{eq_a_con}) and $L_{\text{con}}^M$ (Eq. \ref{eq_m_con}) directly strengthen the disentangled representation performance by introducing prior knowledge of shared anatomical or modality information between paired images or same-modality images. Meanwhile, the similarity losses $L_{\text{sim}}^A$ (Eq. \ref{eq_a_sim}) and $L_{\text{sim}}^M$ (Eq. \ref{eq_m_sim}) prevent information leakage between representations by constraining anatomical differences between MRIs of different subjects and modality differences between MRIs of different sequences, thereby further enhancing the disentangled representation. Additionally, the anatomical code consistency loss $L_{\text{con}}^{AE}$ (Eq. \ref{eq_accl}) guides the model to learn anatomy and personalized information by aligning anatomical codes.
	\begin{table}[t!]\scriptsize
		\caption{\textcolor{black}{The Ablation Study for Loss Functions. The best results are marked in bold.}}
		\label{tab6}
		\centering
        \renewcommand{\arraystretch}{1.0}
		\setlength{\tabcolsep}{1.5mm}{
        \resizebox{\columnwidth}{!}{
			\begin{tabular}{ccccccccccc}
				\toprule[1.pt]
				\multirow{2}{*}{$L_{\text{con}}^A$} & \multirow{2}{*}{$L_{\text{sim}}^A$} & \multirow{2}{*}{$L_{\text{con}}^M$} & \multirow{2}{*}{$L_{\text{sim}}^M$} & \multirow{2}{*}{$L_{\text{con}}^{AE}$} & \multicolumn{2}{c}{T1} & \multicolumn{2}{c}{T2} & \multicolumn{2}{c}{Average} \\
				&  &  &  &  & AP$\uparrow$ & DSC$\uparrow$ & AP$\uparrow$ & DSC$\uparrow$ & AP$\uparrow$ & DSC$\uparrow$ \\
				\hline
				&  &   &   &   & 42.82 & 40.41 & 67.53 & 60.02 & 55.18 & 50.22 \\
				& \checkmark &   &   &   & 46.88 & 42.01 & 78.07 & 67.81 & 62.48 & 54.91 \\
				\checkmark  &   &   &   &   & 48.96 & 43.96 & 79.66 & 67.94 & 64.31 & 55.95 \\
				\checkmark  & \checkmark  &   &   &   & 49.42 & 44.77 & 80.16 & 68.94 & 64.79 & 56.86 \\
				\checkmark  & \checkmark  &   & \checkmark  &   & 51.69 & 44.27 & 80.91 & 70.84 & 66.30 & 57.56 \\
				\checkmark  & \checkmark  & \checkmark  &   &   & 51.82 & 45.42 & 81.23 & 69.84 & 66.53 & 57.63 \\
				\checkmark  & \checkmark  & \checkmark  & \checkmark  &   & 53.03 & 47.56 & 82.59 & 71.55 & 67.81 & 59.56 \\
				\checkmark  & \checkmark  & \checkmark  & \checkmark  & \checkmark  & \textbf{54.97} & \textbf{49.32} & \textbf{85.10} & \textbf{73.65} & \textbf{70.04} & \textbf{61.49}\\
				\toprule[1.pt]
		\end{tabular}
        }
        }
	\end{table}
	\subsubsection{Ablation Study of Major Modules}
		\begin{table}[]\scriptsize
		\caption{\textcolor{black}{The Ablation Study for Major Modules. The best results are marked in bold.}}
		\label{tab7}
		\centering
        \renewcommand{\arraystretch}{1.1}
		\setlength{\tabcolsep}{0.9mm}{
        \resizebox{\columnwidth}{!}{
			\begin{tabular}{ccccccccc}
				\toprule[1.pt]
				\multirow{2}{*}{\begin{tabular}[c]{@{}c@{}}Disentangling\\ representation\end{tabular}} & \multirow{2}{*}{\begin{tabular}[c]{@{}c@{}}Edge-to-image\\ restoration\end{tabular}} & \multirow{2}{*}{\begin{tabular}[c]{@{}c@{}}Modality\\ reuse\end{tabular}} & \multicolumn{2}{c}{T1} & \multicolumn{2}{c}{T2} & \multicolumn{2}{c}{Average} \\
				&  &  & AP$\uparrow$ & DSC$\uparrow$ & AP$\uparrow$ & DSC$\uparrow$ & AP$\uparrow$ & DSC$\uparrow$ \\
				\hline
				\checkmark &  & \checkmark & 32.75 & 32.25 & 38.11 & 47.96 & 35.43 & 40.11 \\
				& \checkmark &  & 52.49 & 44.55 & 72.68 & 61.75 & 62.59 & 53.15 \\
				\checkmark & \checkmark &  & 54.56 & 46.36 & 82.48 & 71.78 & 68.52 & 59.07 \\
				\checkmark & \checkmark & \checkmark & \textbf{54.97} & \textbf{49.32} & \textbf{85.10} & \textbf{73.65} & \textbf{70.04} & \textbf{61.49}\\
				\toprule[1.pt]
		\end{tabular}
        }
        }
	\end{table}
	We also validated the effectiveness of major modules in our method, including the disentangling representation, edge-to-image restoration, and  modality reuse (i.e., imaging recoupling, which is discussed separately as it is the key to imaging alignment), through ablation experiments on the BraTS-GLI dataset, and the results are presented in Table \ref{tab7}. The first, second and third rows of Table \ref{tab7} show, respectively, that the exclusion of edge-to-image restoration resulted in a substantial decline in the average AP/DSC metrics from 70.04\%/61.49\% to 35.43\%/40.11\%. Similarly, omitting the disentangling representation reduced the metrics to 62.59\%/53.15\%, while the absence of modality reuse caused a decrease to 68.52\%/59.07\%. These results highlight the effectiveness of these modules.
    \subsection{Effectiveness Analysis of Major Modules}
\begin{figure}[h]
		\centerline{\includegraphics[width=3.5in]{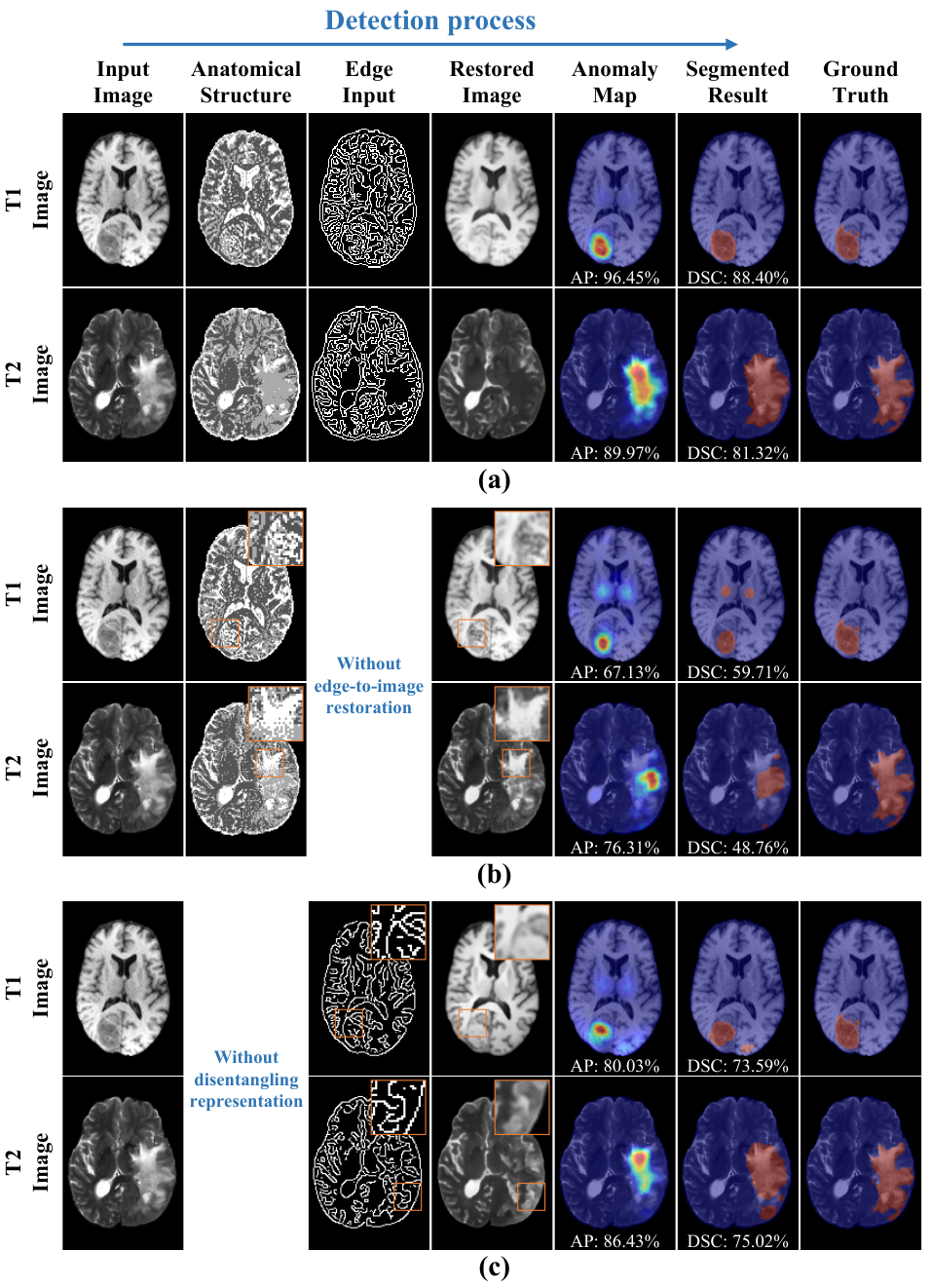}}
		\caption{\textcolor{black}{Visualization of intermediate outputs of our method in different ablation setting. (a) Image restoration based on anatomical edges and modality representation (ours). (b) Image reconstruction without edge-to-image restoration. (c) Image restoration without disentangling representation.}}
		\label{fig5}
	\end{figure}

    We further assess the effectiveness of the major modules of our method by visualizing intuitively the intermediate outputs of our method in different ablation settings, as shown in Fig. \ref{fig5}.
	\subsubsection{Effectiveness of Edge-to-Image Restoration}
	As shown in Fig. \ref{fig5}(b), without this module, abnormal regions in the input images also exhibited abnormal semantics in anatomical images, suggesting that disentangling representation alone cannot effectively suppress anomalies. In contrast, our method employed edge-to-image restoration to ensure that abnormal semantic pixels were invisible to restoration network $R_{ED}$ composed of $R_E$ and $R_D$, better suppressing anomalies.
	\subsubsection{Effectiveness of Disentangling Representation}
	We extracted edges from input images and reconstructed them without disentangling representations. As illustrated in Fig. \ref{fig5}(c), in the restored images, the abnormal regions near the edges were not well repaired, especially in the T2 images. This is primarily due to the overfitting of the restoration network to invariant image edges, causing it to generate inaccurate local anatomy for the residual abnormal edges. In contrast, our method has improved the repair of abnormal regions near abnormal edges through disentangling representation, as shown in Fig. \ref{fig5}(a). Specifically, the unsupervised disentanglement process introduces anatomical noise, as noted in previous works \citep{ref3,ref4}, which generates edge noise. This edge noise enables the restoration network to undergo denoising training, helping it adapt to abnormal edges and improving its ability to handle real abnormal edges, as discussed in previous work \citep{ref13}.
	\subsubsection{Effectiveness of Modality Reuse}
    	\begin{figure*}[]
		\centerline{\includegraphics[width=7.0in]{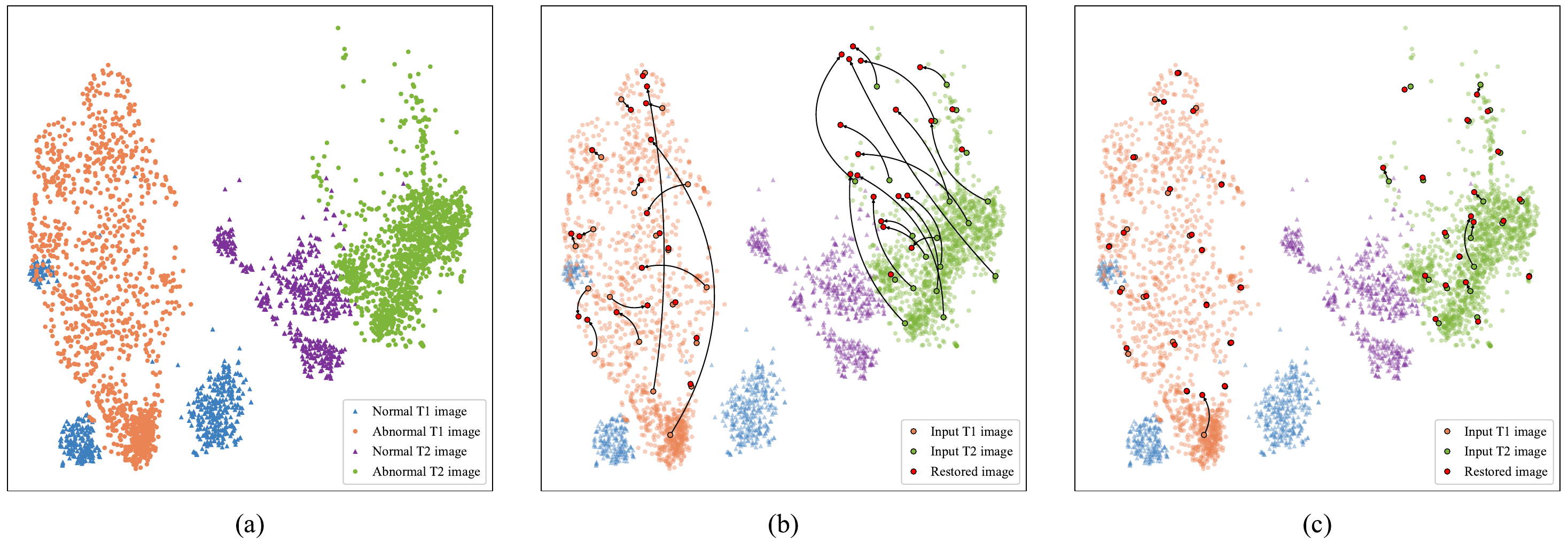}}
		\caption{\textcolor{black}{The t-SNE visualization of the modality representations. (a) Modality representations of original normal and abnormal images. (b) Modality representation shifts of restored images relative to input images without modality reuse. (c) Modality representation shifts of restored images relative to input images with modality reuse (ours).}}
		\label{fig6}
	\end{figure*}
    
    This section visually analyzes the modality reuse mechanism, which is designed to preserve the personalized modality representation (imaging information) of the input image in the restored pseudo-healthy image (PHI), that is, to align the modality representation of the PHI with input image, thereby reducing false positives in normal regions of the difference-based anomaly map. We used a pre-trained modality extractor ($E_M$) in our model to obtain high-dimensional modality representations from different normal (IXI\textsuperscript{\ref{fn:ixi_dataset}}) and abnormal (BraTS-GLI \citep{ref42,ref43}) images and visualized them using t-SNE \citep{Van08} in Fig. \ref{fig6}, where closer proximity between points indicates higher imaging similarity. Fig. \ref{fig6}(a) reveals significant modality representation differences between T1 and T2 images, with normal T1 images clustering into three groups, corresponding to the three different centers from which the IXI dataset originates. This demonstrates that our model effectively captures personalized modality representations and highlights the imaging heterogeneity across centers.
    
    To evaluate the modality reuse mechanism, Fig. \ref{fig6}(b) and (c) visualize the modality representations of images restored without and with this mechanism, respectively. Arrows connect the modality representation of an original image (input image) to its restored counterpart, illustrating the modality representation shift. Without modality reuse (Fig. \ref{fig6}(b)), the modality representations of the restored images often deviate significantly from the originals (long arrows), indicating a loss of personalized imaging information. In contrast, with modality reuse (Fig. \ref{fig6}(c)), the modality representations remain tightly anchored to the originals (short or absent arrows), confirming that the personalized imaging information is better preserved (aligned). Further quantitative analysis showed that modality reuse improved SSIM scores between input and restored images for T1/T2 normal images from 0.9685/0.9546 to 0.9731/0.9587, suggesting that it enhances image similarity by aligning the modality representation of the restored image with the input image, thereby reducing false positives in normal regions and improving generalizability to multi-modality and multi-center MRIs. Furthermore, we visualized the anatomical representation in \ref{Anatomy visualization}.
	\section{Discussions}
	  We first discuss why our method outperforms other comparison methods in detection performance and generalizability. Next, we discuss the scalability of our method to new imaging sequences. We then analyze how training on different normal datasets affects detection performance of the model, with results summarized in Table~\ref{tab8}. Finally, we discuss the limitations of our method.
      \subsection{Performance and Generalizability Analysis}
	\subsubsection{Performance} As shown in Fig. \ref{fig2}, the size, location, and appearance of abnormal regions in different brain images significantly vary, and normal regions have highly personalized details. Existing reconstruction-based methods cannot simultaneously repair all anomalies to normal and accurately reconstruct personalized details of normal regions, as illustrated in Fig. \ref{fig3}. Our method restores pseudo-healthy images based on edges, ensures that abnormal pixels are invisible to the restoration model to fully suppress any anomalies, and leverages the structural information preserved in edges to effectively restore personalized normal details. Therefore, our method can more effectively detect various brain anomalies.
	\subsubsection{Generalizability} As shown in Fig. \ref{fig6}(a), brain MRIs from different sequences or centers exhibit imaging heterogeneity due to varying imaging conditions. Existing methods do not take this into account, limiting their generalizability to multi-modality and multi-center MRIs. To address this, we propose the disentangled representation module to strip away imaging information, ensuring that the detection process is unaffected by imaging heterogeneity. This brings two clinical benefits: (1) The single model can detect anomalies from T1 and T2 MRIs without training different models for specific sequences, reducing clinical deployment burdens; and (2) our model shows consistent performance across MRIs from multiple centers, as shown in Table \ref{tab3}, enabling the training of a unified model that can be deployed across centers.

    \subsection{Scalability to New Imaging Sequences}
    In our experiments, the model was trained on paired T1 and T2 images and demonstrated its capability to detect anomalies in both sequences. Furthermore, the proposed framework is designed to be scalable, allowing for the incorporation of new imaging sequences, such as FLAIR, without requiring alterations to the network architecture. This incorporation would require retraining the whole model by extending the relevant loss functions (e.g., Eqs. \ref{eq_a_con}-\ref{eq_recon} and Eq. \ref{eq_rest}) to include the new imaging sequence. Specifically, the anatomy consistency loss in Eq. \ref{eq_a_con} would be expanded to
    \begin{equation*}
	\label{eq_a_con_ex}
		L_{\text{con}}^{A} = \sum\limits_{{i,j \in \{1,2,3\}, i \ne j}} \left\| \cos(A_i, A_j) - 1 \right\|_2^2,
	\end{equation*}
    where the indices 1, 2, and 3 represent the T1, T2, and FLAIR sequences, respectively. Other relevant loss functions would be expanded in a similar manner.
     \subsection{Impact of Different Normal Training Dataset}
        \begin{table*}[]
        		\caption{Performance of models trained on different normal datasets and evaluated on four glioma datasets. The best results are marked in bold.}
        		\label{tab8}
        		\centering
                \resizebox{\linewidth}{!}{
        			\begin{tabular}{ccccccccccccccccccccc}
                    \toprule[1.pt]
        \multirow{4}{*}{\makecell{  \\ Trained Dataset}} & \multicolumn{16}{c}{Adult glioma} & \multicolumn{4}{c}{Pediatric glioma} \\
        \cmidrule(l){2-17} \cmidrule(l){18-21} 
        & \multicolumn{4}{c}{\makecell{BraTS-GLI \\ (Age not reported)}} & \multicolumn{4}{c}{\makecell{UPenn-GBM \\ (Ages 18.7-88.5)}} & \multicolumn{4}{c}{\makecell{UCSF-PDGM \\ (Ages 17-94)}} & \multicolumn{4}{c}{Average} & \multicolumn{4}{c}{\makecell{BraTS-PED \\ (Age not reported)}} \\
         \cmidrule(l){2-5}\cmidrule(l){6-9} \cmidrule(l){10-13} \cmidrule(l){14-17} \cmidrule(l){18-21} 
         & \multicolumn{2}{c}{T1} & \multicolumn{2}{c}{T2} & \multicolumn{2}{c}{T1} & \multicolumn{2}{c}{T2} & \multicolumn{2}{c}{T1} & \multicolumn{2}{c}{T2} & \multicolumn{2}{c}{T1} & \multicolumn{2}{c}{T2} & \multicolumn{2}{c}{T1} & \multicolumn{2}{c}{T2} \\
         & AP$\uparrow$ & DSC$\uparrow$ & AP$\uparrow$ & DSC$\uparrow$ & AP$\uparrow$ & DSC$\uparrow$ & AP$\uparrow$ & DSC$\uparrow$ & AP$\uparrow$ & DSC$\uparrow$ & AP$\uparrow$ & DSC$\uparrow$& AP$\uparrow$ & DSC$\uparrow$ & AP$\uparrow$ & DSC$\uparrow$ & AP$\uparrow$ & DSC$\uparrow$ & AP$\uparrow$ & DSC$\uparrow$ \\
         \midrule

        HCP (Ages 22-35) & 49.70 & 44.37 & 72.75 & 62.42 & 58.08 & 48.50 & 76.19 & 65.80 & 46.97 & 45.32 & 74.00 & 63.02 & 51.58 & 46.06 & 74.31 & 63.75 & 29.34 & 43.51 & 56.36 & 60.77 \\

        IXI (Ages 20.0-86.3) & \textbf{54.97} & \textbf{49.32} & \textbf{85.10} & \textbf{73.65} & \textbf{66.01} & \textbf{55.02} & \textbf{88.64} & \textbf{76.22} & \textbf{50.93} & \textbf{48.44} & \textbf{86.19} & \textbf{72.92} & \textbf{57.30} & \textbf{50.93} & \textbf{86.64} & \textbf{74.26} & \textbf{41.76} & \textbf{51.71} & \textbf{66.15} & \textbf{63.69}\\
        
        \bottomrule[1.pt]
        \end{tabular}
        }
	\end{table*}
         
        To assess the impact of training data diversity on our method's performance, we compared models trained on two different normal datasets: the heterogeneous IXI\textsuperscript{\ref{fn:ixi_dataset}} dataset (577 subjects, ages 20.0-86.3, multiple scanners) and the homogeneous Human Connectome Project (HCP) young adult dataset \citep{hcpdataset} (1,113 subjects, ages 22-35, single 3T Siemens Connectome scanner). As detailed in Table~\ref{tab8}, the model trained on the more diverse IXI dataset yielded an absolute average increase of 9.03\%(51.58\% vs. 57.30\%, 74.31\% vs. 86.64\%) in AP on adult glioma datasets (ages 17-94), compared to the one trained on the larger but homogeneous HCP dataset. This result confirms that greater diversity in the training data, encompassing varied imaging settings and a wide age range with diverse brain anatomies, substantially enhances the model's generalization performance. It underscores the importance of using heterogeneous datasets for developing robust unsupervised anomaly detection models for real-world clinical scenarios.
    
    \subsection{Limitation Analysis}
    
    \begin{figure}[h]
    \centerline{\includegraphics[width=3.5in]{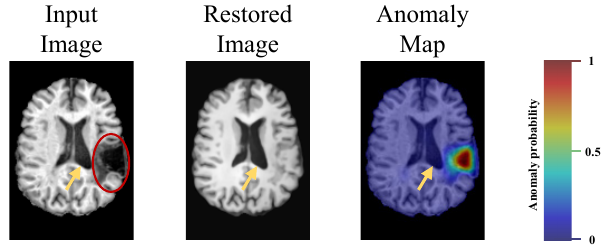}}
    \caption{\textcolor{black}{Limitations of our method for detecting abnormal anatomical deformation. Our model successfully detects the lesion with abnormal signal intensity (red circle) but may face challenges in identifying the abnormal anatomical deformations (yellow arrow).}}
    \label{fig7}
    \end{figure}
    
    As illustrated in Fig. \ref{fig7}, our model successfully corrects abnormal intensities (red circle) that deviate from healthy tissue, allowing for the effective localization of such anomalies. However, a limitation of our method is that our model may face challenges in detecting large-scale anatomical deformations caused by the underlying pathology, such as brain anatomical warping or atrophy (yellow arrow). The fundamental reason is that our pseudo-healthy image generation process is conditioned on the anatomical edges of the input image. When the anatomical structure itself is deformed, these altered anatomical edges are preserved in the generated pseudo-healthy image. To address this challenge, our future work is developing methods that can detect these abnormal anatomical deformations.
    
    In the proposed method, several hyperparameters (e.g., kernel sizes of the post-processing filters and loss weights) were selected based on the full BraTS-GLI \citep{ref42,ref43} and MSLUB \citep{ref53} datasets, which might bias the results towards certain lesion types. In the future, we plan to collect data with more lesion types and set up a dedicated validation set for hyperparameter selection to reduce potential bias. This validation set should be drawn partially from datasets of different lesion types, ensuring a balance in the number of samples for each lesion type.
	\section{Conclusions}
	In this paper, we proposed an unsupervised method for detecting diverse brain anomalies in brain MRI. Our major findings include that our method can (1) detect anomalies more effectively through edge-to-image restoration that adequately repairs anomalies while preserving personalized normal details, and (2) improve model generalizability to multi-modality and multi-center MRIs by disentangling representations. Extensive experiments and comparisons with 17 SOTA methods on nine multi-modality datasets demonstrated its substantial superiority and improvement.
	
\section*{CRediT authorship contribution statement}
\textbf{Tao Yang}: Writing – original draft, Visualization, Validation, Software, Methodology, Formal analysis;
\textbf{Xiuying Wang}: Writing – review \& editing, Visualization, Formal analysis;
\textbf{Hao Liu}: Validation, Methodology, Data curation;
\textbf{Guanzhong Gong}: Investigation, Data curation;
\textbf{Lian-Ming Wu}: Writing – review \& editing, Visualization;
\textbf{Yu-Ping Wang}: Writing – review \& editing, Validation;
\textbf{Lisheng Wang}: Writing – review \& editing, Supervision, Project administration, Validation.

\section*{Declaration of competing interest}
The authors declare that they have no known competing financial interests or personal relationships that could have appeared to influence the work reported in this paper.

\section*{Acknowledgements}
This work was supported by the Major Program of the National Natural Science Foundation of China (Grant No. 12090024) and the Fundamental Research Funds for the Central Universities (Grant No. YG2023QNA47).

\section*{Data Availability}
All datasets used in this study are publicly available at the following links.
\begin{itemize}
    \item The IXI dataset is accessible at: \url{https://brain-development.org/ixi-dataset/}
    \item The HCP dataset is accessible at: \url{https://www.humanconnectome.org/study/hcp-young-adult}
    \item The BraTS-related datasets, including BraTS-GLI, BraTS-SSA, BraTS-PED, BraTS-MEN, and BraTS-MET, can be obtained from the Brain Tumor Segmentation (BraTS) Challenge 2023 at: \url{https://www.synapse.org/#!Synapse:syn51156910/wiki/622341}
    \item The UPenn-GBM dataset is accessible on The Cancer Imaging Archive (TCIA) at: \url{https://www.cancerimagingarchive.net/collection/upenn-gbm/}
    \item The UCSF-PDGM dataset is available at: \url{https://www.cancerimagingarchive.net/collection/ucsf-pdgm/}
    \item The MSLUB dataset is available at: \url{http://lit.fe.uni-lj.si/tools}
    \item The ATLAS dataset can be accessed at: \url{http://fcon_1000.projects.nitrc.org/indi/retro/atlas.html}
\end{itemize}

\section*{Code Availability}
The source code for this work is available at \url{https://github.com/yangtao-hub/DAL}. The implementations for the comparison methods in this study are available at the following links.
\begin{itemize}
    \item The implementations for DRAEM, MKD, RD4AD, UTRAD, PatchCore, and EfficientAD can be obtained from the Anomalib \citep{akcay2022anomalib} (\url{https://github.com/openvinotoolkit/anomalib}) and BMAD \citep{ref40} (\url{https://github.com/DorisBao/BMAD}) libraries.
    \item The implementations for AE and VAE can be obtained from the repository \citep{ref15} (\url{https://github.com/StefanDenn3r/Unsupervised_Anomaly_Detection_Brain_MRI}).
    \item FPI: \url{https://github.com/jemtan/FPI}
    \item AMCons: \url{https://github.com/jusiro/constrained_anomaly_segmentation}
    \item MemAE: \url{https://github.com/donggong1/memae-anomaly-detection}
    \item RIAD: \url{https://github.com/plutoyuxie/Reconstruction-by-inpainting-for-visual-anomaly-detection}
    \item AnoDDPM: \url{https://github.com/Julian-Wyatt/AnoDDPM}
    \item PHANES: \url{https://github.com/ci-ber/PHANES}
    \item DAE: \url{https://github.com/AntanasKascenas/DenoisingAE}
    \item AutoDDPM: \url{https://github.com/ci-ber/autoDDPM}
\end{itemize}

\appendix
\renewcommand{\thefigure}{A.\arabic{figure}}
\renewcommand{\thetable}{A.\arabic{table}}
\renewcommand{\theHfigure}{A.\arabic{figure}}
\renewcommand{\theHtable}{A.\arabic{table}}
\setcounter{figure}{0}
\setcounter{table}{0}
\section{Notation summary}\label{Notation summary}
For clarity and easy reference, we summarize all key notations, modules, and loss functions used throughout our paper in Table \ref{tab:notation}. For each term, we provide its symbol, dimension, and description.
\begin{table}[h]
\caption{\textcolor{black}{Summary of notations, modules and loss functions.}} 
\label{tab:notation}
\begin{center}
\resizebox{0.93\columnwidth}{!}{%
\begin{tabular}{cll}
\toprule[1.pt]
Symbol&Dimension&Description\\
\midrule
\midrule
\multicolumn{3}{l}{Notations}\\ 
\midrule
$D$, $H$, $W$ & $\mathbb{R}^{1}$&  
depth, height, width of an input image \\
 $x$ & $\mathbb{R}^{D \times H \times W}$&  
 input image \\
 $\bar{x}$ & $\mathbb{R}^{D/2 \times H/2 \times W/2}$ &
 downsampled image\\
  $A$ & $\mathbb{R}^{D \times H \times W}$ &
 multi-class anatomical representation\\
   $A_c$ & $\mathbb{R}^{C \times D \times H \times W}$ &
 continuous anatomical representation\\
   $A_{doh}$ & $\mathbb{R}^{C \times D \times H \times W}$ &
 binary anatomical representation\\
$M$& $\mathbb{R}^{m}$& 
modality representation  \\
$M_{brain}$& $\mathbb{R}^{D \times H \times W}$& 
brain mask  \\
$E$& $\mathbb{R}^{D \times H \times W}$& 
anatomical edge  \\
$C_a$& $\mathbb{R}^{c_k \times d_k \times h_k \times w_k}$& 
anatomical code  \\
$\widetilde{C}_a$& $\mathbb{R}^{c_k \times d_k \times h_k \times w_k}$& 
restored anatomical code  \\
${x}_{recon}$& $\mathbb{R}^{D \times H \times W}$ & 
reconstructed image  \\
${x}_{rest}$& $\mathbb{R}^{D \times H \times W}$ & 
restored image  \\
${x}_{a}$& $\mathbb{R}^{D \times H \times W}$ & 
abnormal image  \\
${x}_{r}$& $\mathbb{R}^{D \times H \times W}$ & 
pseudo-healthy image  \\
\midrule
\midrule
\multicolumn{3}{l}{Modules}\\
\midrule
$E_A$ & - & anatomy extractor\\
$E_M$ & - & modality extractor\\
$R_A$ & - & anatomy encoder\\
$R_E$ & - & anatomy restorer\\
$R_D$ & - & representation combiner\\
\midrule
\midrule
\multicolumn{3}{l}{Loss Functions} \\
\midrule
$ L^{con}_{A} $&$\mathbb{R}^{1}$ & anatomy consistency loss\\
$ L^{con}_{M} $&$\mathbb{R}^{1}$ & modality consistency loss\\ 
$ L^{sim}_{A} $&$\mathbb{R}^{1}$ & anatomy similarity loss\\ 
$ L^{sim}_{M} $&$\mathbb{R}^{1}$ & modality similarity loss\\ 
$ L^{recon}_{AD} $&$\mathbb{R}^{1}$ & reconstruction loss\\ 
$ L^{con}_{AE} $&$\mathbb{R}^{1}$ & anatomical code consistency loss\\ 
$ L^{res}_{ED} $&$\mathbb{R}^{1}$ & restoration loss\\
$ \mathcal{L} $&$\mathbb{R}^{1}$ & total loss\\
\midrule
\bottomrule[1.pt]
\end{tabular}
}
\end{center}
\end{table}

\renewcommand{\thefigure}{B.\arabic{figure}}
\renewcommand{\thetable}{B.\arabic{table}}
\renewcommand{\theHfigure}{B.\arabic{figure}}
\renewcommand{\theHtable}{B.\arabic{table}}
\setcounter{figure}{0}
\setcounter{table}{0}
\section{Binarization operation}\label{Binarization operation}
We elaborate on the necessity of binarization for $A_c$ from the following two perspectives.

From an intuitive imaging perspective: A specific brain anatomical structure, such as white matter, can be ideally represented by a binary map (0 for background, 1 for the anatomical structure). A specific sequence of MRI scan represents this anatomy with continuous grayscale values by introducing imaging information (e.g., T1-weighted or T2-weighted), thereby mixing anatomical and imaging information. When our anatomy extractor extracts a continuous probability map, $A_c$, from the MRI, these continuous probability values in $A_c$ not only encode the anatomical information but also implicitly carry imaging information, as they are related to the continuous grayscale values in MRI. Therefore, through binarization, we convert the continuous $A_c$ back into a binary anatomical map, thereby stripping out the implicit imaging information.

From an information theory perspective: To completely disentangle an MRI image into mutually independent anatomical representation ($A$) and modality representation ($M$), their mutual information needs to be minimized, i.e., $I(A;M)\to0$. This means we must ensure that the imaging information, such as continuous grayscale values in MRI, cannot be inferred from the anatomical representation $A_c$. Before binarization, the continuous probability values (distribution) in $A_c$ are inherently and implicitly correlated with the continuous grayscale values (distribution) in MRI. Through this correlation, the continuous grayscale values might be inferred, leading to imaging information leakage and incomplete disentanglement (i.e., $I(A;M)>0$). By binarizing $A_c$, we transform it from a continuous probability map into a binary semantic map containing only values of 0 or 1. This binarization significantly reduces the information capacity of $A_c$, making it infeasible to infer the continuous grayscale values from the binarized $A_c$, which hence prevents $A_c$ from capturing imaging information and achieves a more thorough disentanglement. 
\renewcommand{\thefigure}{C.\arabic{figure}}
\renewcommand{\thetable}{C.\arabic{table}}
\renewcommand{\theHfigure}{C.\arabic{figure}}
\renewcommand{\theHtable}{C.\arabic{table}}
\setcounter{figure}{0}
\setcounter{table}{0}
\section{Margin hyperparameter}\label{Margin hyperparameter}
\begin{figure}[h]
\centerline{\includegraphics[width=3.5in]{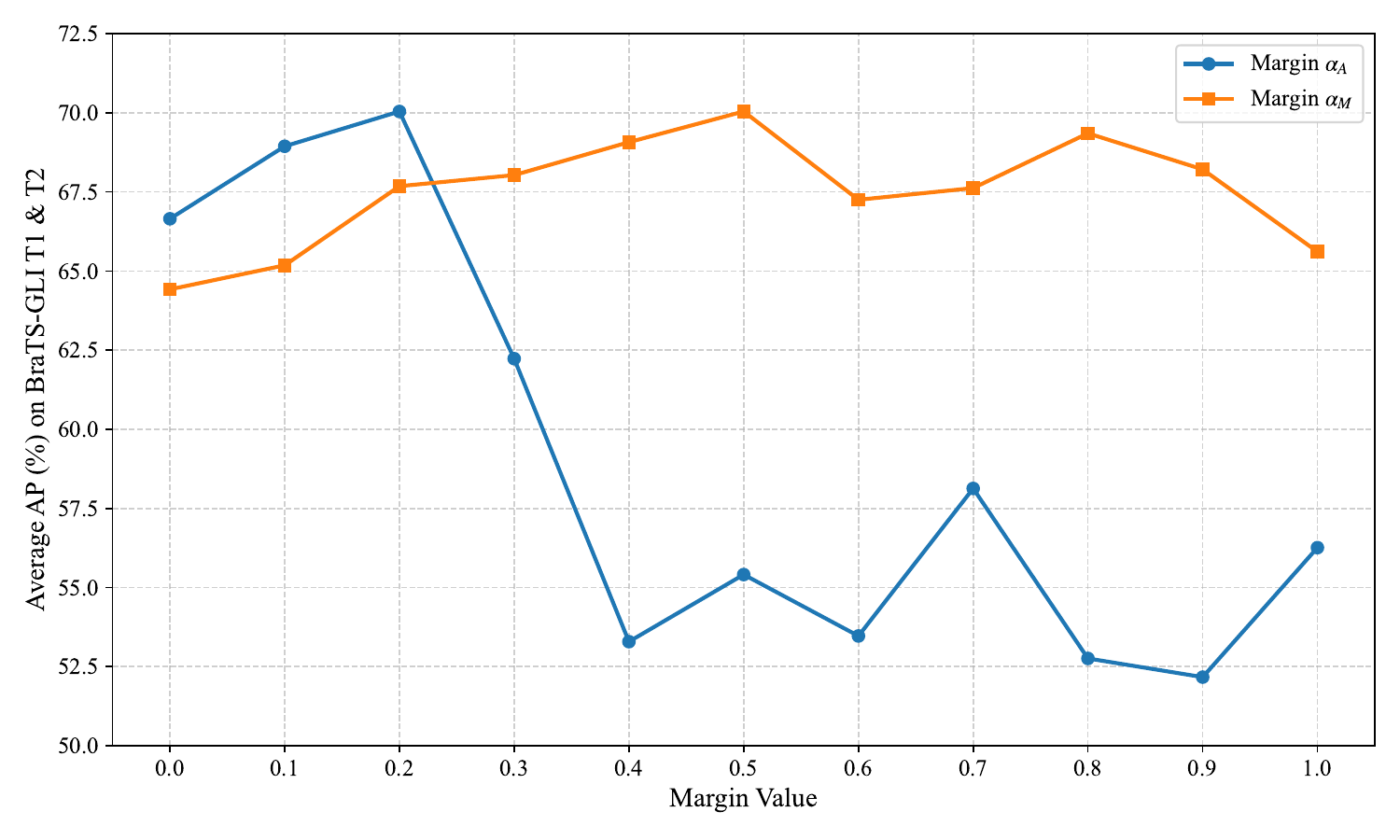}}
\caption{\textcolor{black}{Influence of margin hyperparameters ($\alpha_A$, $\alpha_M$) on anomaly detection performance.}}
\label{figc1}
\end{figure}

We performed the margin hyperparameter search from the set \{0, 0.1, 0.2, \dots, 1.0\}. The optimal margin values ($\alpha_A = 0.2$ and $\alpha_M = 0.5$) were selected to maximize the average AP across the T1-weighted and T2-weighted sequences of the abnormal BraTS-GLI dataset (N=1,251) \citep{ref42,ref43}, as shown in Fig. \ref{figc1}.

\renewcommand{\thefigure}{D.\arabic{figure}}
\renewcommand{\thetable}{D.\arabic{table}}
\renewcommand{\theHfigure}{D.\arabic{figure}}
\renewcommand{\theHtable}{D.\arabic{table}}
\setcounter{figure}{0}
\setcounter{table}{0}
\section{Filter's kernel size}\label{Filter size}
\begin{figure}[h]
\centerline{\includegraphics[width=3.5in]{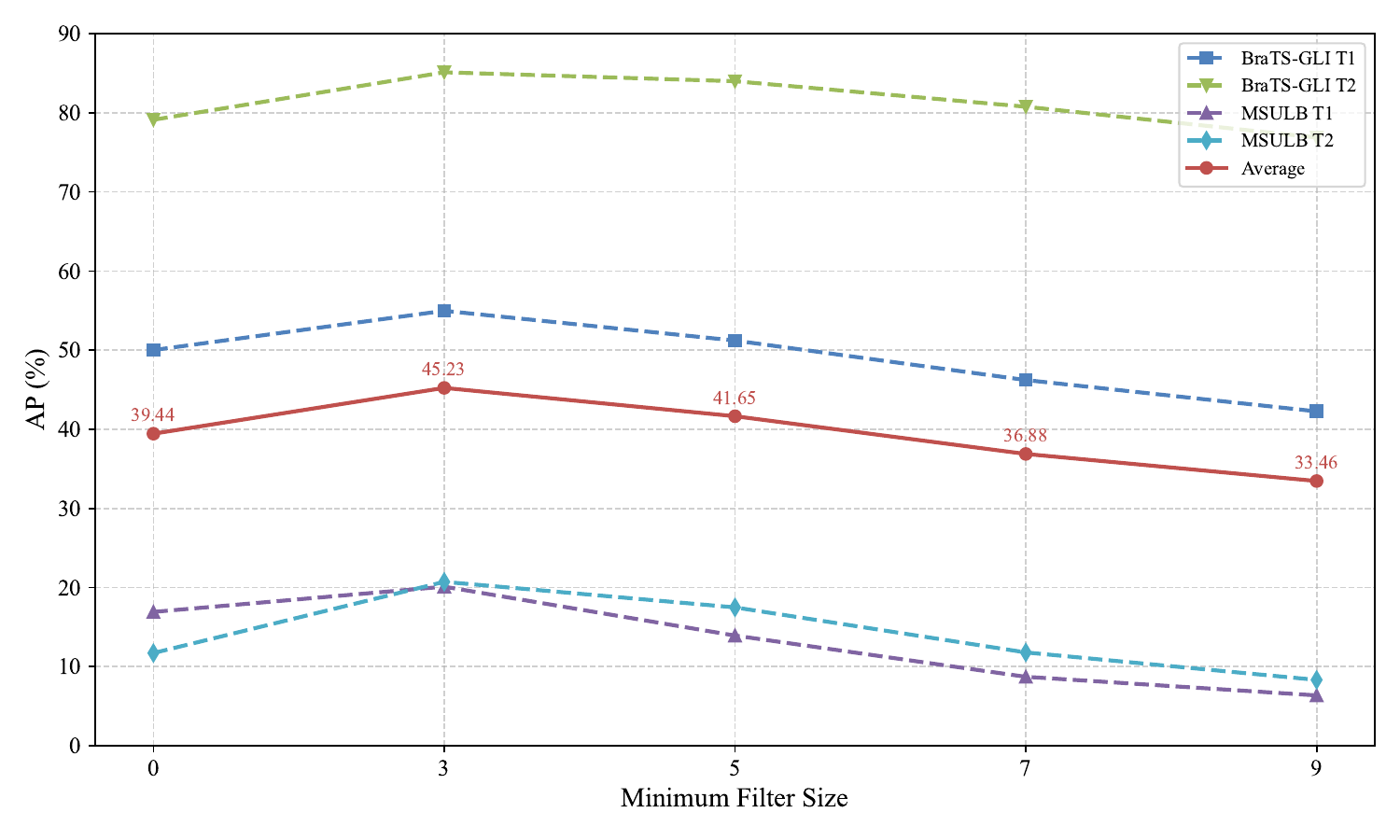}}
\caption{\textcolor{black}{Effect of the minimum filter's kernel size on detection performance, where a size of 0 indicates the filter is not applied.}}
\label{figd1}
\end{figure}

\begin{figure}[h]
\centerline{\includegraphics[width=3.5in]{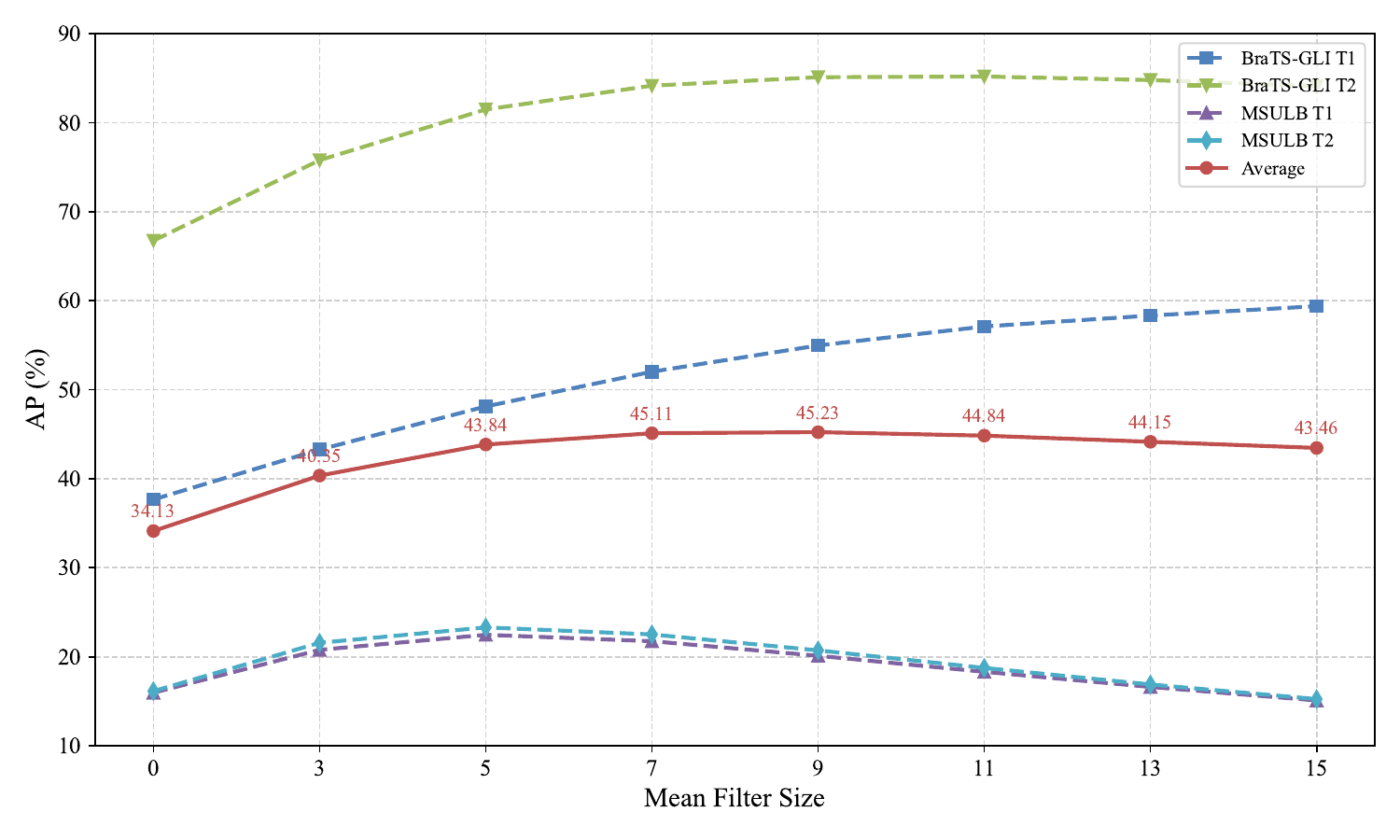}}
\caption{\textcolor{black}{Effect of the mean filter's kernel size on detection performance, where a size of 0 indicates the filter is not applied.}}
\label{figd2}
\end{figure}

Our post-processing pipeline applies a minimum filter to remove small false positives, followed by a mean filter to smooth the anomaly map. The kernel sizes for the minimum and mean filters are set to \(3 \times 3 \times 3\) and \(9 \times 9 \times 9\), respectively. These values were determined through a hyperparameter search within \{0, 3, 5, 7, 9\} for minimum filter and \{0, 3, 5, 7, \dots, 15\} for mean filter, where 0 indicates the filter is not applied, aiming to maximize the average AP on both T1-weighted and T2-weighted sequences across both large-lesion (BraTS-GLI \citep{ref42,ref43}, adult glioma) and small-lesion (MSLUB \citep{ref53}, multiple sclerosis) datasets, as shown in Fig. \ref{figd1} and Fig. \ref{figd2}. 

\renewcommand{\thefigure}{E.\arabic{figure}}
\renewcommand{\thetable}{E.\arabic{table}}
\renewcommand{\theHfigure}{E.\arabic{figure}}
\renewcommand{\theHtable}{E.\arabic{table}}
\setcounter{figure}{0}
\setcounter{table}{0}
\begin{figure*}[p]
\centerline{\includegraphics[width=6.5in]{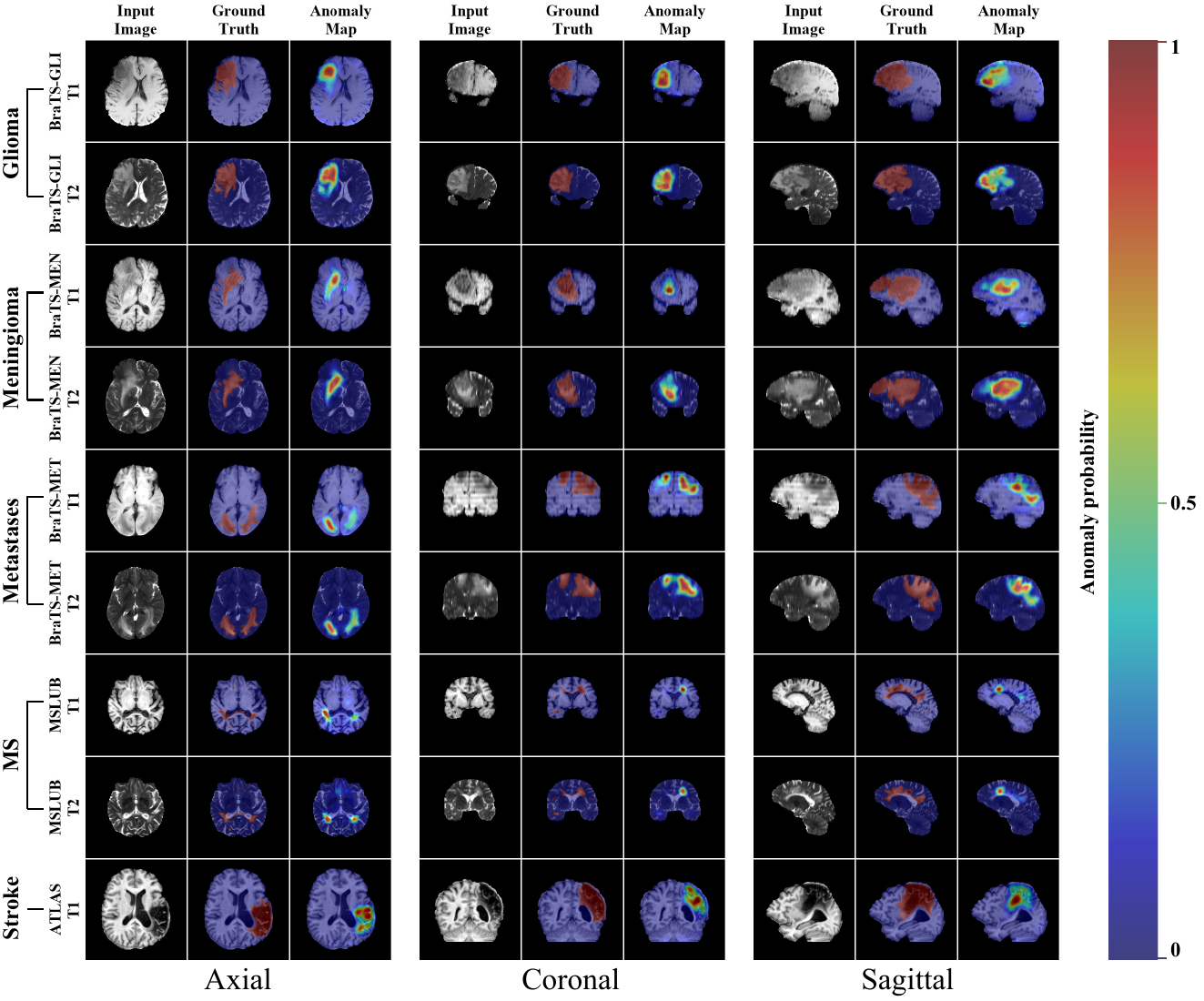}} 
\caption{\textcolor{black}{Anomaly maps generated by the proposed method for 3D brain MRI in three orthogonal planes (axial, coronal, and sagittal).}}
\label{fige1} 
\end{figure*}
\section{Anomaly map visualization}\label{Anomaly map visualization}
We visualized the anomaly maps generated by our method for 3D brain MRI in three orthogonal planes (axial, coronal, and sagittal), as shown in Fig. \ref{fige1}.

\renewcommand{\thefigure}{F.\arabic{figure}}
\renewcommand{\thetable}{F.\arabic{table}}
\renewcommand{\theHfigure}{F.\arabic{figure}}
\renewcommand{\theHtable}{F.\arabic{table}}
\setcounter{figure}{0}
\setcounter{table}{0}
\begin{figure*}[p]
\centerline{\includegraphics[width=3.2in]{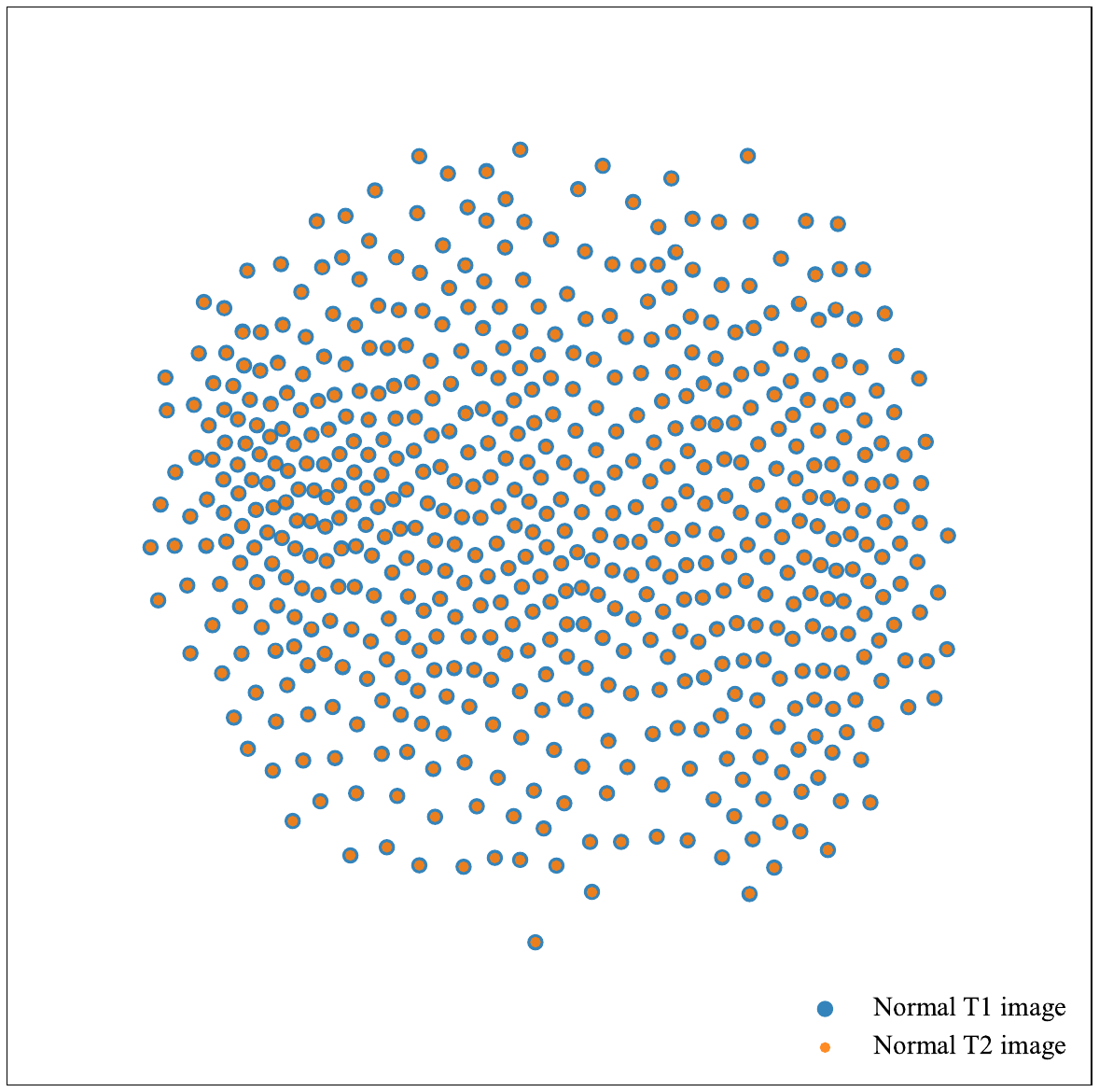}} 
\caption{\textcolor{black}{t-SNE visualization of anatomical structures from healthy brains in the IXI\textsuperscript{\ref{fn:ixi_dataset}} dataset.}}
\label{figf1} 
\end{figure*}
\section{Anatomy visualization}\label{Anatomy visualization}
We visualized the anatomical representations for healthy brains from the IXI\textsuperscript{\ref{fn:ixi_dataset}} dataset using t-SNE \citep{Van08}, as shown in Fig. \ref{figf1}. The result confirms that the anatomical representations from different subjects do not form distinct clusters but are instead scattered. This also demonstrates that our model successfully captures the unique and varied anatomical representations of each individual.

\clearpage
\FloatBarrier

\bibliographystyle{model2-names.bst}\biboptions{authoryear}
\bibliography{refs}

@article{ref1,
	title={Emergency triage of brain computed tomography via anomaly detection with a deep generative model},
	author={Lee, Seungjun and Jeong, Boryeong and Kim, Minjee and Jang, Ryoungwoo and Paik, Wooyul and Kang, Jiseon and Chung, Won Jung and Hong, Gil-Sun and Kim, Namkug},
	journal={Nature Communications},
	volume={13},
	number={1},
	pages={4251},
	year={2022},
	publisher={Nature Publishing Group UK London}
}

@article{ref3,
	title={Unsupervised {MR} harmonization by learning disentangled representations using information bottleneck theory},
	author={Zuo, Lianrui and Dewey, Blake E and Liu, Yihao and He, Yufan and Newsome, Scott D and Mowry, Ellen M and Resnick, Susan M and Prince, Jerry L and Carass, Aaron},
	journal={NeuroImage},
	volume={243},
	pages={118569},
	year={2021},
	publisher={Elsevier}
}

@article{ref4,
	title={Disentangled representation learning in cardiac image analysis},
	author={Chartsias, Agisilaos and Joyce, Thomas and Papanastasiou, Giorgos and Semple, Scott and Williams, Michelle and Newby, David E and Dharmakumar, Rohan and Tsaftaris, Sotirios A},
	journal={Medical Image Analysis},
	volume={58},
	pages={101535},
	year={2019},
	publisher={Elsevier}
}

@article{ref5,
	title={Federated disentangled representation learning for unsupervised brain anomaly detection},
	author={Bercea, Cosmin I and Wiestler, Benedikt and Rueckert, Daniel and Albarqouni, Shadi},
	journal={Nature Machine Intelligence},
	volume={4},
	number={8},
	pages={685--695},
	year={2022},
	publisher={Nature Publishing Group UK London}
}

@inproceedings{ref8,
	title={{DRAEM} -- A discriminatively trained reconstruction embedding for surface anomaly detection},
	author={Zavrtanik, Vitjan and Kristan, Matej and Sko{\v{c}}aj, Danijel},
	booktitle={Proceedings of the IEEE/CVF International Conference on Computer Vision (ICCV)},
	pages={8330--8339},
	year={2021}
}

@article{ref9,
	title={Detecting outliers with foreign patch interpolation},
	author={Tan, Jeremy and Hou, Benjamin and Batten, James and Qiu, Huaqi and Kainz, Bernhard and others},
	journal={Machine Learning for Biomedical Imaging},
	volume={1},
	number={April 2022 issue},
	pages={1--27},
	year={2022}
}

@inproceedings{ref10,
	title={Towards total recall in industrial anomaly detection},
	author={Roth, Karsten and Pemula, Latha and Zepeda, Joaquin and Sch{\"o}lkopf, Bernhard and Brox, Thomas and Gehler, Peter},
	booktitle={Proceedings of the IEEE/CVF Conference on Computer Vision and Pattern Recognition (CVPR)},
	pages={14318--14328},
	year={2022}
}

@inproceedings{ref11,
	title={{EfficientAD}: Accurate visual anomaly detection at millisecond-level latencies},
	author={Batzner, Kilian and Heckler, Lars and K{\"o}nig, Rebecca},
	booktitle={Proceedings of the IEEE/CVF Winter Conference on Applications of Computer Vision (WACV)},
	pages={128--138},
	year={2024}
}

@article{ref12,
	title={Proxy-bridged image reconstruction network for anomaly detection in medical images},
	author={Zhou, Kang and Li, Jing and Luo, Weixin and Li, Zhengxin and Yang, Jianlong and Fu, Huazhu and Cheng, Jun and Liu, Jiang and Gao, Shenghua},
	journal={IEEE Transactions on Medical Imaging},
	year={2022},
	volume={41},
	number={3},
	pages={582--594},
	publisher={IEEE}
}

@article{ref13,
	title={The role of noise in denoising models for anomaly detection in medical images},
	author={Kascenas, Antanas and Sanchez, Pedro and Schrempf, Patrick and Wang, Chaoyang and Clackett, William and Mikhael, Shadia S and Voisey, Jeremy P and Goatman, Keith and Weir, Alexander and Pugeault, Nicolas and others},
	journal={Medical Image Analysis},
	volume={90},
	pages={102963},
	year={2023},
	publisher={Elsevier}
}

@inproceedings{ref14,
	title={Unsupervised anomaly localization using variational auto-encoders},
	author={Zimmerer, David and Isensee, Fabian and Petersen, Jens and Kohl, Simon and Maier-Hein, Klaus},
	booktitle={Medical Image Computing and Computer-Assisted Intervention (MICCAI)},
	pages={289--297},
	year={2019},
	organization={Springer}
}

@article{ref15,
	title={Autoencoders for unsupervised anomaly segmentation in brain {MR} images: a comparative study},
	author={Baur, Christoph and Denner, Stefan and Wiestler, Benedikt and Navab, Nassir and Albarqouni, Shadi},
	journal={Medical Image Analysis},
	volume={69},
	pages={101952},
	year={2021},
	publisher={Elsevier}
}

@inproceedings{ref16,
	title={Deep autoencoding models for unsupervised anomaly segmentation in brain {MR} images},
	author={Baur, Christoph and Wiestler, Benedikt and Albarqouni, Shadi and Navab, Nassir},
	booktitle={Brainlesion: Glioma, Multiple Sclerosis, Stroke and Traumatic Brain Injuries},
	pages={161--169},
	year={2019},
	organization={Springer}
}

@inproceedings{ref17,
	title={Memorizing normality to detect anomaly: Memory-augmented deep autoencoder for unsupervised anomaly detection},
	author={Gong, Dong and Liu, Lingqiao and Le, Vuong and Saha, Budhaditya and Mansour, Moussa Reda and Venkatesh, Svetha and Hengel, Anton van den},
	booktitle={Proceedings of the IEEE/CVF International Conference on Computer Vision (ICCV)},
	pages={1705--1714},
	year={2019}
}

@article{ref18,
	title={Reconstruction by inpainting for visual anomaly detection},
	author={Zavrtanik, Vitjan and Kristan, Matej and Sko{\v{c}}aj, Danijel},
	journal={Pattern Recognition},
	volume={112},
	pages={107706},
	year={2021},
	publisher={Elsevier}
}

@inproceedings{ref25,
	title={Multiresolution knowledge distillation for anomaly detection},
	author={Salehi, Mohammadreza and Sadjadi, Niousha and Baselizadeh, Soroosh and Rohban, Mohammad H and Rabiee, Hamid R},
	booktitle={Proceedings of the IEEE/CVF Conference on Computer Vision and Pattern Recognition (CVPR)},
	pages={14902--14912},
	year={2021}
}

@inproceedings{ref26,
	title={Anomaly detection via reverse distillation from one-class embedding},
	author={Deng, Hanqiu and Li, Xingyu},
	booktitle={Proceedings of the IEEE/CVF Conference on Computer Vision and Pattern Recognition (CVPR)},
	pages={9737--9746},
	year={2022}
}

@article{ref27,
	title={{UTRAD}: Anomaly detection and localization with {U}-transformer},
	author={Chen, Liyang and You, Zhiyuan and Zhang, Nian and Xi, Juntong and Le, Xinyi},
	journal={Neural Networks},
	volume={147},
	pages={53--62},
	year={2022},
	publisher={Elsevier}
}

@article{ref28,
	title={Constrained unsupervised anomaly segmentation},
	author={Silva-Rodr{\'\i}guez, Julio and Naranjo, Valery and Dolz, Jose},
	journal={Medical Image Analysis},
	volume={80},
	pages={102526},
	year={2022},
	publisher={Elsevier}
}

@article{Van08,
	title={Visualizing data using {t-SNE}},
	author={Van der Maaten, Laurens and Hinton, Geoffrey},
	journal={Journal of Machine Learning Research},
	volume={9},
	number={11},
	pages={2579--2605},
	year={2008}
}

@inproceedings{ref35,
	title={Arbitrary style transfer in real-time with adaptive instance normalization},
	author={Huang, Xun and Belongie, Serge},
	booktitle={Proceedings of the IEEE International Conference on Computer Vision (ICCV)},
	pages={1501--1510},
	year={2017}
}

@article{ref38,
	title={Image quality assessment: from error visibility to structural similarity},
	author={Wang, Zhou and Bovik, Alan C and Sheikh, Hamid R and Simoncelli, Eero P},
	journal={IEEE Transactions on Image Processing},
	volume={13},
	number={4},
	pages={600--612},
	year={2004},
	publisher={IEEE}
}

@inproceedings{ref40,
	title={{BMAD}: Benchmarks for medical anomaly detection},
	author={Bao, Jinan and Sun, Hanshi and Deng, Hanqiu and He, Yinsheng and Zhang, Zhaoxiang and Li, Xingyu},
	booktitle={Proceedings of the IEEE/CVF Conference on Computer Vision and Pattern Recognition (CVPR)},
	pages={4042--4053},
	year={2024}
}

@article{ref42,
	title={The {RSNA-ASNR-MICCAI BraTS} 2021 benchmark on brain tumor segmentation and radiogenomic classification},
	author={Baid, Ujjwal and Ghodasara, Satyam and Mohan, Suyash and Bilello, Michel and Calabrese, Evan and Colak, Errol and Farahani, Keyvan and Kalpathy-Cramer, Jayashree and Kitamura, Felipe C and Pati, Sarthak and others},
	journal={arXiv preprint arXiv:2107.02314},
	year={2021}
}

@article{ref43,
	title={The multimodal brain tumor image segmentation benchmark ({BraTS})},
	author={Menze, Bjoern H and Jakab, Andras and Bauer, Stefan and Kalpathy-Cramer, Jayashree and Farahani, Keyvan and Kirby, Justin and Burren, Yuliya and Porz, Nicole and Slotboom, Johannes and Wiest, Roland and others},
	journal={IEEE Transactions on Medical Imaging},
	year={2015},
	volume={34},
	number={10},
	pages={1993--2024},
	publisher={IEEE}
}

@article{ref45,
	title={The {University of Pennsylvania} glioblastoma ({UPenn-GBM}) cohort: advanced {MRI}, clinical, genomics, \& radiomics},
	author={Bakas, Spyridon and Sako, Chiharu and Akbari, Hamed and Bilello, Michel and Sotiras, Aristeidis and Shukla, Gaurav and Rudie, Jeffrey D and Santamar{\'\i}a, Natali Flores and Kazerooni, Anahita Fathi and Pati, Sarthak and others},
	journal={Scientific Data},
	volume={9},
	number={1},
	pages={453},
	year={2022},
	publisher={Nature Publishing Group UK London}
}

@article{ref46,
	title={Multi-parametric magnetic resonance imaging ({mpMRI}) scans for de novo Glioblastoma ({GBM}) patients from the {University of Pennsylvania Health System} ({UPenn-GBM}) (Version 2) [Data set]},
	author={Bakas, Spyridon and Sako, Chiharu and Akbari, Hamed and Bilello, M and Sotiras, A and Shukla, G and others},
	journal={The Cancer Imaging Archive},
	year={2021}
}

@article{ref47,
	title={The {University of California San Francisco} preoperative diffuse glioma {MRI} dataset},
	author={Calabrese, Evan and Villanueva-Meyer, Javier E and Rudie, Jeffrey D and Rauschecker, Andreas M and Baid, Ujjwal and Bakas, Spyridon and Cha, Soonmee and Mongan, John T and Hess, Christopher P},
	journal={Radiology: Artificial Intelligence},
	volume={4},
	number={6},
	pages={e220058},
	year={2022},
	publisher={Radiological Society of North America}
}

@article{ref48,
	title={The {University of California San Francisco} Preoperative Diffuse Glioma {MRI} ({UCSF-PDGM}) (Version 4) [Dataset]},
	author={Calabrese, E and Villanueva-Meyer, J and Rudie, J and Rauschecker, A and Baid, U and Bakas, S and Cha, S and Mongan, J and Hess, C},
	journal={The Cancer Imaging Archive},
	year={2022}
}

@article{ref49,
	title={The brain tumor segmentation ({BraTS}) challenge 2023: glioma segmentation in {Sub-Saharan Africa} patient population ({BraTS-Africa})},
	author={Adewole, Maruf and Rudie, Jeffrey D and Gbadamosi, Anu and Toyobo, Oluyemisi and Raymond, Confidence and Zhang, Dong and Omidiji, Olubukola and Akinola, Rachel and Suwaid, Mohammad Abba and Emegoakor, Adaobi and others},
	journal={arXiv preprint arXiv:2305.19369},
	year={2023}
}

@article{ref50,
	title={The brain tumor segmentation ({BraTS}) challenge 2023: focus on pediatrics ({CBTN-CONNECT-DIPGR-ASNR-MICCAI BraTS-PEDs})},
	author={Kazerooni, Anahita Fathi and Khalili, Nastaran and Liu, Xinyang and Haldar, Debanjan and Jiang, Zhifan and Anwar, Syed Muhammed and Albrecht, Jake and Adewole, Maruf and Anazodo, Udunna and Anderson, Hannah and others},
	journal={arXiv preprint arXiv:2305.17033},
	year={2023}
}

@article{ref51,
	title={The {ASNR-MICCAI} brain tumor segmentation ({BraTS}) challenge 2023: Intracranial meningioma},
	author={LaBella, Dominic and Adewole, Maruf and Alonso-Basanta, Michelle and Altes, Talissa and Anwar, Syed Muhammad and Baid, Ujjwal and Bergquist, Timothy and Bhalerao, Radhika and Chen, Sully and Chung, Verena and others},
	journal={arXiv preprint arXiv:2305.07642},
	year={2023}
}

@article{ref52,
	title={The brain tumor segmentation ({BraTS-METS}) challenge 2023: Brain metastasis segmentation on pre-treatment {MRI}},
	author={Moawad, Ahmed W and Janas, Anastasia and Baid, Ujjwal and Ramakrishnan, Divya and Jekel, Leon and Krantchev, Kiril and Moy, Harrison and Saluja, Rachit and Osenberg, Klara and Wilms, Klara and others},
	journal={arXiv preprint arXiv:2306.00838},
	year={2023}
}

@article{ref53,
	title={A novel public {MR} image dataset of multiple sclerosis patients with lesion segmentations based on multi-rater consensus},
	author={Lesjak, {\v{Z}}iga and Galimzianova, Alfiia and Koren, Ale{\v{s}} and Lukin, Matej and Pernu{\v{s}}, Franjo and Likar, Bo{\v{s}}tjan and {\v{S}}piclin, {\v{Z}}iga},
	journal={Neuroinformatics},
	volume={16},
	pages={51--63},
	year={2018},
	publisher={Springer}
}

@article{ref54,
	title={A large, curated, open-source stroke neuroimaging dataset to improve lesion segmentation algorithms},
	author={Liew, Sook-Lei and Lo, Bethany P and Donnelly, Miranda R and Zavaliangos-Petropulu, Artemis and Jeong, Jessica N and Barisano, Giuseppe and Hutton, Alexandre and Simon, Julia P and Juliano, Julia M and Suri, Anisha and others},
	journal={Scientific Data},
	volume={9},
	number={1},
	pages={320},
	year={2022},
	publisher={Nature Publishing Group UK London}
}

@article{ref55,
	title={{N4ITK}: improved {N3} bias correction},
	author={Tustison, Nicholas J and Avants, Brian B and Cook, Philip A and Zheng, Yuanjie and Egan, Alexander and Yushkevich, Paul A and Gee, James C},
	journal={IEEE Transactions on Medical Imaging},
	volume={29},
	number={6},
	pages={1310--1320},
	year={2010},
	publisher={IEEE}
}

@article{ref58,
	title={The {SRI24} multichannel atlas of normal adult human brain structure},
	author={Rohlfing, Torsten and Zahr, Natalie M and Sullivan, Edith V and Pfefferbaum, Adolf},
	journal={Human Brain Mapping},
	volume={31},
	number={5},
	pages={798--819},
	year={2010},
	publisher={Wiley Online Library}
}

@article{ref59,
	title={On standardizing the {MR} image intensity scale},
	author={Ny{\'u}l, L{\'a}szl{\'o} G and Udupa, Jayaram K},
	journal={Magnetic Resonance in Medicine},
	volume={42},
	number={6},
	pages={1072--1081},
	year={1999},
	publisher={Wiley Online Library}
}

@article{ref60,
	title={Evaluating intensity normalization on {MRIs} of human brain with multiple sclerosis},
	author={Shah, Mohak and Xiao, Yiming and Subbanna, Nagesh and Francis, Simon and Arnold, Douglas L and Collins, D Louis and Arbel, Tal},
	journal={Medical Image Analysis},
	volume={15},
	number={2},
	pages={267--282},
	year={2011},
	publisher={Elsevier}
}

@article{lagogiannis2023unsupervised,
	title={Unsupervised pathology detection: a deep dive into the state of the art},
	author={Lagogiannis, Ioannis and Meissen, Felix and Kaissis, Georgios and Rueckert, Daniel},
	journal={IEEE Transactions on Medical Imaging},
	year={2024},
	volume={43},
	number={1},
	pages={241--252},
	publisher={IEEE}
}

@article{li2023self,
	title={Self-supervised multi-scale cropping and simple masked attentive predicting for lung {CT}-scan anomaly detection},
	author={Li, Wei and Liu, Guang-Hai and Fan, Haoyi and Li, Zuoyong and Zhang, David},
	journal={IEEE Transactions on Medical Imaging},
	year={2024},
	volume={43},
	number={1},
	pages={594--607},
	publisher={IEEE}
}

@article{xu2024facing,
	title={Facing differences of similarity: Intra-and Inter-correlation unsupervised learning for Chest {X-Ray} anomaly detection},
	author={Xu, Shicheng and Li, Wei and Li, Zuoyong and Zhao, Tiesong and Zhang, Bob},
	journal={IEEE Transactions on Medical Imaging},
	year={2025},
	volume={44},
	number={2},
	pages={801--814},
	publisher={IEEE}
}

@article{dong2023swssl,
	title={{SWSSL}: Sliding window-based self-supervised learning for anomaly detection in high-resolution images},
	author={Dong, Haoyu and Zhang, Yifan and Gu, Hanxue and Konz, Nicholas and Zhang, Yixin and Mazurowski, Maciej A},
	journal={IEEE Transactions on Medical Imaging},
	year={2023},
	volume={42},
	number={12},
	pages={3860--3870},
	publisher={IEEE}
}

@article{cai2025medianomaly,
  title={{Medianomaly}: A comparative study of anomaly detection in medical images},
  author={Cai, Yu and Zhang, Weiwen and Chen, Hao and Cheng, Kwang-Ting},
  journal={Medical Image Analysis},
  pages={103500},
  year={2025},
  publisher={Elsevier}
}

@article{zimmerer2022mood,
  title={{MOOD} 2020: A public Benchmark for Out-of-Distribution Detection and Localization on medical Images},
  author={Zimmerer, David and Full, Peter M and Isensee, Fabian and J{\"a}ger, Paul and Adler, Tim and Petersen, Jens and K{\"o}hler, Gregor and Ross, Tobias and Reinke, Annika and Kascenas, Antanas and others},
  journal={IEEE Transactions on Medical Imaging},
  volume={41},
  number={10},
  pages={2728--2738},
  year={2022},
  publisher={IEEE}
}

@article{hcpdataset,
  title={The {WU-Minn Human Connectome Project}: an overview},
  author={Van Essen, David C and Smith, Stephen M and Barch, Deanna M and Behrens, Timothy EJ and Yacoub, Essa and Ugurbil, Kamil and Wu-Minn HCP Consortium and others},
  journal={NeuroImage},
  volume={80},
  pages={62--79},
  year={2013},
  publisher={Elsevier}
}

@article{labella2024multi,
  title={A multi-institutional meningioma {MRI} dataset for automated multi-sequence image segmentation},
  author={LaBella, Dominic and Khanna, Omaditya and McBurney-Lin, Shan and Mclean, Ryan and Nedelec, Pierre and Rashid, Arif S and Tahon, Nourel Hoda and Altes, Talissa and Baid, Ujjwal and Bhalerao, Radhika and others},
  journal={Scientific Data},
  volume={11},
  number={1},
  pages={496},
  year={2024},
  publisher={Nature Publishing Group UK London}
}

@article{schlegl2019f,
  title={{f-AnoGAN}: Fast unsupervised anomaly detection with generative adversarial networks},
  author={Schlegl, Thomas and Seeb{\"o}ck, Philipp and Waldstein, Sebastian M and Langs, Georg and Schmidt-Erfurth, Ursula},
  journal={Medical Image Analysis},
  volume={54},
  pages={30--44},
  year={2019},
  publisher={Elsevier}
}

@inproceedings{bercea2023reversing,
  title={Reversing the abnormal: Pseudo-healthy generative networks for anomaly detection},
  author={Bercea, Cosmin I and Wiestler, Benedikt and Rueckert, Daniel and Schnabel, Julia A},
  booktitle={Medical Image Computing and Computer-Assisted Intervention (MICCAI)},
  pages={293--303},
  year={2023},
  organization={Springer}
}

@inproceedings{wyatt2022anoddpm,
  title={{AnoDDPM}: Anomaly detection with denoising diffusion probabilistic models using simplex noise},
  author={Wyatt, Julian and Leach, Adam and Schmon, Sebastian M and Willcocks, Chris G},
  booktitle={Proceedings of the IEEE/CVF Conference on Computer Vision and Pattern Recognition (CVPR)},
  pages={650--656},
  year={2022}
}

@inproceedings{bercea2023mask,
  title={Mask, stitch, and re-sample: Enhancing robustness and generalizability in anomaly detection through automatic diffusion models},
  author={Bercea, Cosmin I and Neumayr, Michael and Rueckert, Daniel and Schnabel, Julia A},
  booktitle={ICML 3rd Workshop on Interpretable Machine Learning in Healthcare},
  year={2023}
}

@article{raya2021ai,
  title={{AI}-based strategies to reduce workload in breast cancer screening with mammography and tomosynthesis: a retrospective evaluation},
  author={Raya-Povedano, Jos{\'e} Luis and Romero-Mart{\'\i}n, Sara and El{\'\i}as-Cabot, Esperanza and Gubern-M{\'e}rida, Albert and Rodr{\'\i}guez-Ruiz, Alejandro and {\'A}lvarez-Benito, Marina},
  journal={Radiology},
  volume={300},
  number={1},
  pages={57--65},
  year={2021},
  publisher={Radiological Society of North America}
}

@article{ahn2022association,
  title={Association of artificial intelligence--aided chest radiograph interpretation with reader performance and efficiency},
  author={Ahn, Jong Seok and Ebrahimian, Shadi and McDermott, Shaunagh and Lee, Sanghyup and Naccarato, Laura and Di Capua, John F and Wu, Markus Y and Zhang, Eric W and Muse, Victorine and Miller, Benjamin and others},
  journal={JAMA Network Open},
  volume={5},
  number={8},
  pages={e2229289--e2229289},
  year={2022},
  publisher={American Medical Association}
}

@article{villanueva2017current,
  title={Current clinical brain tumor imaging},
  author={Villanueva-Meyer, Javier E and Mabray, Marc C and Cha, Soonmee},
  journal={Neurosurgery},
  volume={81},
  number={3},
  pages={397--415},
  year={2017},
  publisher={LWW}
}

@article{sahraian2010role,
  title={Role of {MRI} in diagnosis and treatment of multiple sclerosis},
  author={Sahraian, Mohammad Ali and Eshaghi, Arman},
  journal={Clinical Neurology and Neurosurgery},
  volume={112},
  number={7},
  pages={609--615},
  year={2010},
  publisher={Elsevier}
}

@article{yan2018deeplesion,
  title={{DeepLesion}: Automated mining of large-scale lesion annotations and universal lesion detection with deep learning},
  author={Yan, Ke and Wang, Xiaosong and Lu, Le and Summers, Ronald M},
  journal={Journal of Medical Imaging},
  volume={5},
  number={3},
  pages={036501--036501},
  year={2018},
  publisher={Society of Photo-Optical Instrumentation Engineers}
}

@article{varoquaux2023evaluating,
  title={Evaluating machine learning models and their diagnostic value},
  author={Varoquaux, Ga{\"e}l and Colliot, Olivier},
  journal={Machine Learning for Brain Disorders},
  pages={601--630},
  year={2023},
  publisher={Springer}
}

@article{kingma2013auto,
  title={Auto-encoding variational {Bayes}},
  author={Kingma, Diederik P and Welling, Max},
  journal={arXiv preprint arXiv:1312.6114},
  year={2013}
}

@article{goodfellow2014generative,
  title={Generative adversarial nets},
  author={Goodfellow, Ian J and Pouget-Abadie, Jean and Mirza, Mehdi and Xu, Bing and Warde-Farley, David and Ozair, Sherjil and Courville, Aaron and Bengio, Yoshua},
  journal={Advances in Neural Information Processing Systems (NeurIPS)},
  volume={27},
  year={2014}
}

@article{ho2020denoising,
  title={Denoising diffusion probabilistic models},
  author={Ho, Jonathan and Jain, Ajay and Abbeel, Pieter},
  journal={Advances in Neural Information Processing Systems (NeurIPS)},
  volume={33},
  pages={6840--6851},
  year={2020}
}

@article{zeng2022aggregated,
  title={Aggregated contextual transformations for high-resolution image inpainting},
  author={Zeng, Yanhong and Fu, Jianlong and Chao, Hongyang and Guo, Baining},
  journal={IEEE Transactions on Visualization and Computer Graphics},
  volume={29},
  number={7},
  pages={3266--3280},
  year={2022},
  publisher={IEEE}
}

@article{thakur2020brain,
  title={Brain extraction on {MRI} scans in presence of diffuse glioma: Multi-institutional performance evaluation of deep learning methods and robust modality-agnostic training},
  author={Thakur, Siddhesh and Doshi, Jimit and Pati, Sarthak and Rathore, Saima and Sako, Chiharu and Bilello, Michel and Ha, Sung Min and Shukla, Gaurav and Flanders, Adam and Kotrotsou, Aikaterini and others},
  journal={NeuroImage},
  volume={220},
  pages={117081},
  year={2020},
  publisher={Elsevier}
}

@article{davatzikos2018cancer,
  title={Cancer imaging phenomics toolkit: quantitative imaging analytics for precision diagnostics and predictive modeling of clinical outcome},
  author={Davatzikos, Christos and Rathore, Saima and Bakas, Spyridon and Pati, Sarthak and Bergman, Mark and Kalarot, Ratheesh and Sridharan, Patmaa and Gastounioti, Aimilia and Jahani, Nariman and Cohen, Eric and others},
  journal={Journal of Medical Imaging},
  volume={5},
  number={1},
  pages={011018--011018},
  year={2018},
  publisher={Society of Photo-Optical Instrumentation Engineers}
}

@inproceedings{akcay2022anomalib,
  title={{Anomalib}: A deep learning library for anomaly detection},
  author={Akcay, Samet and Ameln, Dick and Vaidya, Ashwin and Lakshmanan, Barath and Ahuja, Nilesh and Genc, Utku},
  booktitle={Proceedings of the IEEE International Conference on Image Processing (ICIP)},
  pages={1706--1710},
  year={2022},
  organization={IEEE}
}

@article{behrendt2025guided,
  title={Guided reconstruction with conditioned diffusion models for unsupervised anomaly detection in brain {MRIs}},
  author={Behrendt, Finn and Bhattacharya, Debayan and Mieling, Robin and Maack, Lennart and Kr{\"u}ger, Julia and Opfer, Roland and Schlaefer, Alexander},
  journal={Computers in Biology and Medicine},
  volume={186},
  pages={109660},
  year={2025},
  publisher={Elsevier}
}
\end{document}